%% file: look_alike.tex
\documentclass[sigconf, screen]{acmart}
\AtBeginDocument{%
  \providecommand\BibTeX{{%
    \normalfont B\kern-0.5em{\scshape i\kern-0.25em b}\kern-0.8em\TeX}}}

\setcopyright{acmcopyright}
\copyrightyear{2020}
\acmYear{2020}
\acmDOI{10.1145/nnnnnnn.nnnnnnn}

\acmConference[AdKDD '20]{AdKDD'20}{August 23, 2020}{San Diego, CA, USA}
\acmBooktitle{AdKDD '20, August 23, 2020, San Diego, CA, USA}
\acmPrice{15.00}
\acmISBN{978-x-xxxx-xxxx-x/YY/MM}

\usepackage{graphicx}
\colorlet{gray6}{black!50!}
\colorlet{gray5}{black!60!}
\colorlet{gray4}{black!70!}
\colorlet{gray3}{black!80!}
\colorlet{gray2}{black!90!}
\colorlet{gray1}{black!100!}

\input{preamble.tex}
\newcolumntype{C}[1]{>{\centering\arraybackslash}p{#1}}
\usepackage{appendix}
\usepackage{multibib}
\newcites{New}{The other list}

\begin{document}
\title{Predicting conversions in display advertising  based on URL embeddings}

\author{Yang Qiu}
\affiliation{%
  \institution{\scalebox{1}{\hspace{-2em} \'{E}cole Polytechnique \& Jellyfish}}
%   \city{Paris}
%   \country{France}
}
\email{yanq.qiu@jellyfish.com}

\author{Nikolaos Tziortziotis}
\affiliation{%
  \institution{Jellyfish}
%   \city{Paris}
%   \country{France}
}
\email{ntziorzi@gmail.com}

\author{Martial Hue}
\affiliation{%
  \institution{Jellyfish}
%   \city{Paris}
%   \country{France}
}
\email{martial.hue@jellyfish.com}

\author{Michalis Vazirgiannis}
\affiliation{%
  \institution{\'{E}cole Polytechnique}
%   \city{Paris}
%   \country{France}
}
\email{mvazirg@lix.polytechnique.fr}

\begin{abstract}
Online display advertising is growing rapidly in recent years thanks to the automation of the ad buying process. 
Real-time bidding (RTB) allows the automated trading of ad impressions between advertisers and publishers through real-time auctions.
In order to increase the effectiveness of their campaigns, advertisers should deliver ads to the users who are highly likely to be converted (i.e., purchase, registration, website visit, etc.) in the near future.
In this study, we introduce and examine different models for estimating the probability of a user converting, given their history of visited URLs. 
Inspired by natural language processing, we introduce three URL embedding models to compute semantically meaningful URL representations.
To demonstrate the effectiveness of the different proposed representation and conversion prediction models, we have conducted experiments on real logged events collected from an advertising platform. 
  %\keywords{URL Embeddings, Display Advertising, Conversion Prediction, User Modeling, E-commerce, Deep Learning}
\end{abstract}
\maketitle              % typeset the header of the contribution
\input{introduction}
\input{related}
\input{preliminaries}
\input{method}
\input{experiments}
\input{conclusions}

\bibliographystyle{ACM-Reference-Format}
\bibliography{misc}

\begin{appendices}
\vspace{2em}
\noindent{\huge\bfseries SUPPLEMENTARY MATERIAL\par}
\vspace{1em}
\input{appendix.tex}
\end{appendices}

\end{document}

%% file: preamble.tex
\usepackage{diagbox}
\usepackage{graphicx}
\usepackage{subcaption}
\usepackage{hyperref}
\hypersetup{
    colorlinks=true,
    linkcolor = blue,
    urlcolor=cyan
}
\usepackage{enumitem}% http://ctan.org/pkg/enumitem

\usepackage{tikz}
\usepackage{dsfont}
\usepackage{isomath}
\usepackage{amssymb}
\usepackage{amsmath}
\usepackage{bm}

\usetikzlibrary{shapes,arrows}
\usetikzlibrary{positioning}

\tikzstyle{rec} = [rectangle, minimum width=2cm, minimum height=1cm, fill=white, draw=black]
\tikzstyle{textnode} = [draw=none,fill=none]
\tikzstyle{arrow} = [thick,->,>=stealth]

\newcommand \E {\mathop{\mbox{\ensuremath{\mathbb{E}}}}\nolimits}

\usepackage{array}
\newcolumntype{P}[1]{>{\centering\arraybackslash}p{#1}}

\newcommand \vx {{\bm{x}}}

\newcommand \vz {\bm{z}}

\newcommand \vb {\vectorsym{b}}
\newcommand \ve {\bm{e}}

\newcommand \Reals {{\mathds{R}}}

\newcommand \CD {{\mathcal{D}}}

\newcommand \CF {{\mathcal{F}}}

\newcommand \CL {{\mathcal{L}}}

\newcommand \param {\bm{\theta}}

\newcommand \defn {\mathrel{\triangleq}}

%% file: introduction.tex
%auto-ignore
\section{Introduction}
\label{sec:introduction}

\begin{figure}[t!]	
  \centering
  \scalebox{.25}{\includegraphics{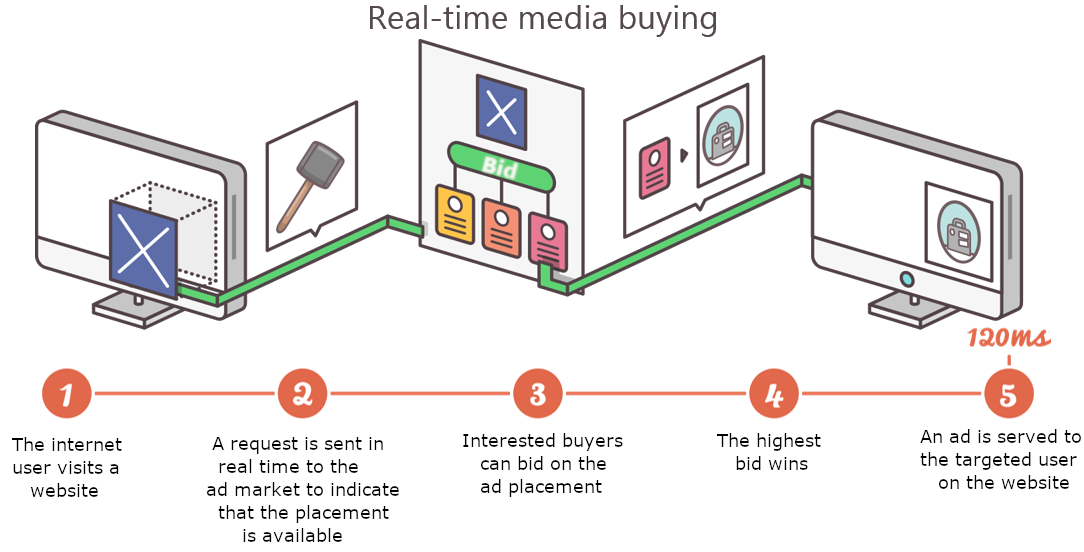}}
    \vspace{-1.em}
  \caption{A high-level overview of RTB procedure.}
  \vspace{-1.em}
  \label{fig:RTB}
\end{figure}

In online display advertising \cite{Wang:2017:DAR}, advertisers promote their products by embedding ads on the publisher's web page.
The majority of all these online display ads are served through Real-Time Bidding (RTB)  \cite{google:2011:ODA}.
RTB allows the publishers to sell their ad placements via the Supply-Side Platform (SSP) and the advertisers to purchase these via the Demand-Side Platform (DSP).
More specifically, each time a user visits a website that contains a banner placement, an auction is triggered.
The publisher sends user's information to the SSP, which forwards this information to the Ad exchange (AdX), and finally the AdX sends a bid request to the DSPs.
Then each DSP decides if it will submit or not a bid response for this impression, based on its information about user, advertisement, urls, etc. 
Once the DSPs send back to the AdX their bids, a public auction takes place with the impression to be sold to the highest bidder. 
Figure~\ref{fig:RTB} briefly illustrates the procedure of online display advertising.

DSPs are agent platforms that help advertisers optimise their advertising strategies.
Roughly speaking, DSPs try to estimate the optimal bid price for an ad impression  in order to maximise the  audience  of the campaigns of their advertisers, given some budget constraints.
The bid price of an ad impression is highly related to the additive value that this impression could have on the advertising campaign (i.e., the number of ad impressions, clicks or conversions, etc.).
In this context, advertisers have at their disposal a number of different pricing models.
In the case where the objective of the advertisers is to maximise the exposure of their advertising message to a targeted audience, paying per impression, referred as \emph{cost-per-mille} (CPM), is probably the best option for them.
Nevertheless, in most of the cases, performance display advertising is more attractive to advertisers that are interested in accomplishing specific goals reducing their risks.
In this case, advertisers are willing to pay for an ad impression if and only if that impression will drive the user to take a predefined action/conversion \cite{Mahdian:2007:PMO}, such as a visit on the advertiser's website, a purchase of a product, etc. 
Two performance payment models have been introduced for this purpose, referred as \emph{cost-per-click} (CPC) and \emph{cost-per-action} (CPA).

In performance-driven display advertising, DSPs submit a bid for a given ad impression based on the CPC or CPA that the advertiser is willing to pay.
To determine the optimal bid price for an ad impression,  DSPs estimate the \emph{expected cost per impression}, called eCPI, which is either equal to the click-through-rate (CTR) for this impression multiplied by the value of CPC, or the conversion rate (CVR) multiplied by the value of CPA \cite{Chen:2011:RBA}.
As a result, accurate CTR/CVR prediction plays an important role in the success of online advertising.
For this purpose, DSPs build CTR/CVR prediction models able to estimate the probability a user converting after their exposure to an advertisement. %, and based on his features.
The accuracy of these models is of high importance for the success of the campaign as if we overestimate click or conversation rates, we will probably submit quite higher bids than we should do, winning possible useless ad impressions.
On the other hand, if these rates are underestimated, we will probably miss ad impressions likely to lead to a user action.

In this work we examine the user conversion problem, where given an advertiser, we want to predict if a user will be converted or not based on their history.
In contrast to previous works that use a number of features related to the user profile, ad information and context information, we consider only the user's browsing history.
More specifically, each user is represented as a sequence of URLs visited by the user in a single day.
Therefore, the problem examined in this paper can be formally described as: given a user's sequence of URLs from a single day, predict the probability this user to take a predefined action on the next day.
In our case, a user is considered as \emph{converted} if they visit the advertiser's website. 
Due to the high cardinality  and diversity of URLs, a compact semantically meaningful representation of URLs is of high importance.
For this purpose, we build and examine three URL embedding models following the idea of word embeddings \cite{Mikolov:2013:w2v_2}. %and have been used with success in natural language processing (NLP). 
The sequential dependency between user's browsing history has also been considered by using a Recurrent Neural Network (RNN) \cite{Graves:2012:RNN}. % \cite{Rumelhart:1988:LRB,Graves:2012:RNN}. 
In total, ten different prediction conversion models have been introduced.
A number of large scale experiments have been executed on a data collected from a real-world advertising platform in order to reveal and compare the prediction abilities of the proposed prediction schemes. Finally, our empirical analysis validates our claims about the effectiveness of our representation models showing that they achieve to group together URLs of the same category. It means that URLs with the same or similar context are also close on the embedding space. 

% The remainder of this paper is organised as follows.
% Section~\ref{sec:related} gives an overview of related work and our contributions, while Section~\ref{sec:preliminaries} presents some preliminaries and formally introduces the architecture of the proposed conversion prediction model.
% Section~\ref{sec:method} presents the details of the four (a one-hot and three embeddings) URL representations used by the prediction model.
% The results of our empirical analysis are presented in Section~\ref{sec:experiments}, and we conclude in Section~\ref{sec:conclusion} with a discussion of future directions. 

%% file: related.tex
%auto-ignore
\section{Related work}
\label{sec:related}

As the performance of a campaign is directly related on how precisely the CVR/CTR is estimated, it has been the objective of considerable research in the past few years.
Typically, the problem of CVR/CTR estimation is formulated as a standard binary classification problem.
Logistic regression has been extensively used to accurately identify conversion events \cite{Richardson:2007:PCE,McMahan:2013:FTRL,Chapelle:2014:SSR}.
\cite{Graepel:2010:WBC} introduced a Bayesian learning model, called Bayesian probit regression, which quantifies the uncertainty over a model's parameters and hence about the CTR of a given ad-impression.  
A precise user representation (a set of features describing user behaviour) constitutes also the foundation for building a linear model able to estimate CVR with high accuracy.
Nevertheless, in most cases it requires a lot of feature engineering effort and the knowledge of the domain.
Moreover, linear models are not capable to reveal the relationship among feature.
To overcome this problem, a number of non-linear models such as factorisation machines \cite{Oentaryo:2014:PRM,Ta:2015:FM} and gradient boosted regression trees \cite{He:2014:PLP} have been also proposed to capture higher order information among features.
A number of different deep learning methods have been also proposed recently for CTR prediction
\cite{Zhang:2014:SCP,Liu:2015:CCP,Zhou:2018:DIN,Chan:2018:CNNCTR}.
%\cite{Zhang:2014:SCP,Liu:2015:CCP,Zhai:2016:DLA,Zhou:2018:DIN,Chan:2018:CNNCTR}. 

Representation learning has been applied with success in several applications and has become a field in itself \cite{rep_review:bengio} in the recent years. 
The URL representation architectures presented in this manuscript have been inspired by those used in  natural language processing (NLP) tasks.
Learning high-quality representations of phrases or documents is a long-standing problem in a wide range of NLP tasks.
% One of the most common document representations is that of bag-of-words (BOW), that represents the occurrence (frequency) of each word within a document.
% Despite its simplicity and effectiveness on some simple tasks, it has not been applied with success on more challenging tasks.
% It mainly comes due to the fact that BOW representation is sparse (curse of dimensionality) and it is not able to capture the distance between individual words \cite{Kusner:2015}.
% % To tackle these problems, different techniques have been proposed up to now that try to learn a latent low-dimensional representation of documents.
% Latent Semantic Indexing (LSI) \cite{Deerwester:1990:LSI} and the Latent Dirichlet Allocation (LDA) \cite{Blei:2003:LDA} are two of the earliest works that have been proposed to tackle these problems.
% % More specifically, LSI maps each document to a semantic space using singular value decomposition (SVD), and LDA represents documents as topic distributions with each one of the topics to be characterised by a distribution over words.
% Despite the fact that both methods produce a coherent document representation compared to BOW, they do not achieve to yield a consistently improvement on the performance of the BOW representation.
% Recently, a number of works \cite{Bengio:2003:NPL,Collobert:2008:UAN,Mnih:2009} have been proposed for learning semantically meaningful representations for words, known as word embeddings.
\emph{Word2Vec} \cite{Mikolov:2013:w2v_2} is one of the most well-known word embeddings algorithms.
The main idea behind \emph{Word2Vec} is that words that appear in similar contexts should be close in the learned embedding space.
For this purpose, a (shallow) neural network language model is applied that consists of an input, a projection, and an output layer.
% Continuous bag-of-words (CBOW) and skip-gram are two different architectures of the \emph{Word2Vec} model for computing continuous vector representations.
% In both cases, a (shallow) neural network language model is applied that consists of an input, a projection, and an output layer.
% In the case of the CBOW architecture the current word is predicted based on its context.
% On the other hand, the objective of the Skip-gram architecture is to find word representations that are useful for predicting the context words.
Its simple architecture makes the training extremely efficient. %, allowing us to train \emph{Word2Vec} architecture on a single machine on billions of words in a single day.
%\cite{Kusner:2015} introduced Word Mover's Distance (WMD), that measures the semantic distance between two documents by calculating the minimum distance that the embedded words of one document should move to reach the embedded words of the other one. 
\cite{search2vec:Grbovic} proposed \emph{search2vec} model that learns user search action representations based on contextual co-occurrence  in user search sessions. 
To the best of our knowledge, URLNet \cite{urlnet} is the only work that learns a URL representation but for the task of  malicious URL detection. In contrast to our unsupervised representation scheme that considers the sequential order of URLs,
%(users' browsing history),
URLNet is an end-to-end (supervised) deep learning framework where its character-level and word-level CNNs are jointly optimized to learn the prediction model.

%% file: preliminaries.tex
%auto-ignore
\section{Proposed conversion prediction architecture}
\label{sec:preliminaries}

% In this section we have to set the general problem that we try to solve

The goal of this paper is to predict the probability a user to be converted one day after, given their browsing history on a single day. More specifically, we consider each user as an ordered sequence of URLs, sorted chronologically.
The notion of conversion corresponds to an action of interest for the advertiser, such as  visit on the landing page, purchase of a product, registration, etc. 
Therefore, we can treat the problem of predicting the user conversion as a binary classification problem \cite{Bishop:2006:PRM}, where given a sequence of URLs visited by a user $U_n = \{url_1^n, \dots, url_{T_n}^n\}$, $n = 1, 2, \dots, N$, we want to predict if $U_n$ will be converted or not, $y_n \in \{0,1\}$. The length of the URL sequence, $T_n$, may be different for each user. 

As an analogy to text classification, we view a sequence of URLs as a document, or a sequence of sentences.
In our case, a URL is itself a sequence of tokens, of length at most three (we ignore the rest tokens as they are quite noisy).  
Each URL\footnote{The \texttt{http(s)://} and \texttt{www} parts of each URL are stripped.} is split with a `/' (slash) character, where the first token corresponds to the domain name. 
For instance, \url{https://en.wikipedia.org/wiki/Main_Page} is mapped to [\texttt{en.wikipedia.org}, \texttt{wiki}, \texttt{Main\_Page}].

In order to apply any supervised classification model, such as logistic regression, etc., a semantically meaningful representation 
%(fixed length feature vector) 
of each URL is needed. 
Therefore, a key intermediate step in our model is the learning of a URL representation. 
More precisely, the proposed conversion prediction scheme is composed of two consecutive training phases. The first one corresponds to the learning of the URL representations, while the second one corresponds to the training of a classifier. It should be mentioned that the training processes of these two models are independent.  
%A high-level overview of the pipelines of these two training phases is illustrated at Fig.\ref{fig:rep_model_learning_process}. 

Due to the high cardinality of URLs, we learn the URL representations implicitly by learning and aggregating their tokens representations. 
In this study, we present and examine four different URL representation models, $f_r{:}~url \rightarrow \vx$, where $\vx \in \Reals^d$ and $d$ is the dimensionality of the embedding space. The first one is the simple \emph{one-hot encoding} that treats tokens as atomic units. The main limitation of this representation is that the representation size grows linearly with the corpus and it doesn't consider the similarity between URLs. To overcome these issues, we propose three different URL embedding models  (see Section~\ref{sec:method}). %that are based on the idea of \emph{word embeddings} \cite{Mikolov:2013:w2v_2} (see Section~\ref{sec:method}). % and have shown excellent performance in various NLP tasks. 
%A detailed description of all the proposed representation models is given later in Section~\ref{sec:method}. 

After having trained a representation model, $f_r$, and given a training set $\CD = \{(U_n, y_n)\}_{n=1}^N$, we produce a new dataset $\CD' = \{(X_n, y_n)\}_{n=1}^N$ where $X_n = \{\vx_1^n, \dots, \vx_{T_n}^n\}$ is a sequence of length $T_n$ with elements $\vx_i^n = f_r(url_i^n)$. 
In a nutshell, $\CD'$ contains sequences of URL embedding vectors along with their labels. 
Then, we apply mapping $f_m{:}~X \rightarrow \vz$ that aggregates the URLs embeddings into an embedding vector $\vz \in \Reals^{m}$, where $m$ can be different from $d$.
It results in a single compact representation $\vz$ for each sequence of URLs. 
Next, our goal is to discover a classification model $f_c{:}~\vz \rightarrow \hat{y}$  from a set $\CF$ of possible models with the minimum empirical risk
\begin{equation}
    \min_{f_c \in \CF} \E_{(X,y) \sim \CD'} [ \ell(f_c(f_m(X)), y)],
\end{equation}
where $\ell \in \Reals$ is a non-negative loss function. % \CY \times \CY \rightarrow \Reals$
% To apply logistic regression, we apply mapping $f_m: X \rightarrow \vz$ that aggregates the URLs embeddings into an embedding vector $\vz \in \Reals^d$.
% It results in a single compact representation $\vz$ for each sequence of URLs. 
% In our case, $f_m$ returns the average of the URLs embedding vectors presented on a sequence: $f_m(X) = \frac{1}{T} \sum_{i=1}^T \vx_i$. 
In fact, classifier $f_c$ is trained on dataset $\CD'' = \{(\vz_n,y_n)\}_{n=1}^N$.
In this work, we use logistic regression where the conversion conditional probability of user $(n)$ given their browsing history $X_n$, is modeled as: 
\begin{equation} 
p(y_n = 1 | X_n) = \sigma(\param^\top f_m(X_n) + b),
\label{eq:sigmoidlayer}
\end{equation}
where $\param$ is a vector with the unknown model parameters, $b$ is the bias term and $\sigma(\cdot)$ is the \emph{logistic sigmoid} function. 
From a geometrical point of view, $\param^\top f_m(X_n) + b$ is a hyperplane that separates the two classes. 
To learn the unknown model parameters, the cross-entropy loss is applied:
\begin{equation}
    \CL = - \E_{(X, y) \sim \CD'} [y \log f_c(f_m(X)) + (1-y) \log (1 - f_c(f_m(X)))].
\end{equation}
%A graphical illustration of the proposed conversion prediction model architecture is presented at Fig.~\ref{fig:model_architecture}.

% \begin{figure}[t!]
%     \centering
%     \begin{subfigure}[t]{1.\columnwidth}
%     \centering
%     \scalebox{0.6}{\input{figures/models/representation_learning.tikz}}
%     \caption{URL representation learning}
%     \label{fig:rep_learning}
%     \end{subfigure} 
    
%     \begin{subfigure}[t]{1.\columnwidth}
%     \centering
%     \scalebox{.6}{\input{figures/models/model_learning.tikz}}
%     \caption{Conversion prediction model learning}
%     \label{fig:model_learning}
%     \end{subfigure}
%     \caption{URL representation and conversion classifier learning pipeline. The binary labels are not needed for training the URL representation model.}
%     \label{fig:rep_model_learning_process}
% \end{figure} 

Next, we are describing the three different mapping functions $f_m$ adopted in our work.
The first one returns the average of the URLs embedding vectors presented on a sequence: $f_m^{(1)}(X) = \frac{1}{T} \sum_{i=1}^T \vx_i$.  
The second one considers the dependencies among the features of the embedding vector returned by the first mapping function.
To be more precise, it returns the output of a dense layer with rectified linear units (ReLU), %\cite{Goodfellow:2016:DL}
 that takes as input the average of the URLs embedding vectors, $f^{(2)}_m(X) = g(\param^{(1)\top} f_m^{(1)}(X) + \vb^{(1)})$.
The ReLU uses the activation function $g(z) = \max\{0,z\}$.
The main limitation of applying one of the two aforementioned mappings is that they do not take into account the chronological order in which the URLs appeared on the sequence.
To overcome this limitation we resort to the well-known Long Short Term Memory network (LSTM) \cite{LSTM:HochreiterS97} that is a special kind of RNNs \cite{Rumelhart:1988:LRB} and is suitable to process variable-length sequences. 
%It has been shown that LSTM is able to learn long-term dependencies more easily compared to the simple recurrent architectures.
To be more precise, the third mapping function $f_m^{(3)}$ is an LSTM network able to map an input sequence of URL embeddings $X$ to a fixed-sized vector $\vz$, that can be considered as the representation of the whole sequence.
In all cases, we are feeding the produced vector $\vz$ to a final dense layer with sigmoid activation function.
In the rest of the paper, we denote as \texttt{LR}, \texttt{DLR} and \texttt{RNN} the prediction conversion models which are using the ``average'', ``dense'' and ``LSTM''  mapping functions, respectively. 
A graphical illustration of the proposed conversion prediction model architecture is presented at Fig.\ref{fig:model_architecture}.

\begin{figure}[t]
    \centering
    \scalebox{.43}{\input{./model_architecture.tikz}}
    \caption{The proposed conversion prediction model architecture. It consists of three parts: i) URL embedding layer ($f_r$), ii) URL sequence embedding layer ($f_m$), and iii) Logistic regression classifier ($f_c$). Only the unknown classifier parameters (Eq.~\ref{eq:sigmoidlayer}) of the dense layer and these of LSTM and ``dense'' mappings are trainable.}
    \label{fig:model_architecture}
    \vspace{-1em}
\end{figure}

%% file: model_architecture.tikz
%auto-ignore
\begin{tikzpicture}[align=center]

\node[rectangle, dashed, minimum width=12cm, minimum height=2cm, draw=black] at (5.2, .4) {};

\node[rectangle, minimum width=1cm, minimum height=0.8cm, draw=blue!50, fill=blue!10, text width=1cm, text centered] (url_1) at (0,0) {$url_1$};
\node[rectangle, minimum width=1cm, minimum height=0.8cm, draw=blue!50, fill=blue!10, text width=1cm, text centered, right=1.5cm of url_1] (url_2) {$url_2$};
\node[circle, draw=blue!70, right=1.5cm of url_2] (d_1) {};
\node[circle, draw=blue!70, right=1.5cm of d_1] (d_2) {};
\node[circle, draw=blue!70, right=1.5cm of d_2] (d_3) {};
\node[rectangle, minimum width=1cm, minimum height=0.8cm, draw=blue!50, fill=blue!10, text width=1cm, text centered, right=1.cm of d_3] (url_T) {$url_T$};

\node[textnode, above right=0.5cm of url_2] (p1) {};
\node[textnode, above=0.6cm of d_2] (p2) {};
\draw[arrow] (p1) --  (p2) node[midway,above] () {\bf Time};

\node[rectangle, minimum width=5cm, minimum height=1cm, draw=orange!50, fill=orange!5, text width=5cm, text centered, below right=2.5cm and -2cm of d_1] (embedding_layer) {\bf URL Embedding layer, $f_r$};

\draw[arrow] (url_1) --  (embedding_layer);
\draw[arrow] (url_2) --  (embedding_layer);
\draw[arrow] (url_T) -- (embedding_layer);

\node[rectangle, minimum width=1cm, minimum height=0.8cm, draw=green!50, fill=green!10, text width=3cm, text centered, below right=5cm and -4cm of url_1] (emb_1) {$url_1$ embedding};
\node[rectangle, minimum width=1cm, minimum height=0.8cm, draw=green!50, fill=green!10, text width=3cm, text centered, right=1.5cm of emb_1] (emb_2) {$url_2$ embedding};
\node[circle, draw=green!70, right=1cm of emb_2] (ed_1) {};
\node[circle, draw=green!70, right=1cm of ed_1] (ed_2) {};
\node[circle, draw=green!70, right=1cm of ed_2] (ed_3) {};
\node[rectangle, minimum width=1cm, minimum height=0.8cm, draw=green!50, fill=green!10, text width=3cm, text centered, right=1.5cm of ed_3] (emb_T) {$url_T$ embedding};

\draw[arrow] (embedding_layer) -- (emb_1) node[midway,right=0.5cm] {$f_r(url_1)$};
\draw[arrow] (embedding_layer) -- (emb_2) node[midway,right] {$f_r(url_2)$};
\draw[arrow] (embedding_layer) -- (emb_T) node[midway,right=0.5cm] {$f_r(url_T)$};

\node[rectangle, minimum width=5cm, minimum height=1cm, draw=red!50, fill=red!10, text width=5cm, text centered, below=5cm of embedding_layer] (seq_representation) {{\bf URL Sequence \\ Embedding Layer, $f_m$} \\ \{\texttt{average}, \texttt{dense}, \texttt{LSTM}\} };

\draw[arrow] (emb_1) -- (seq_representation) node[midway,right=0.5cm] {$\vx_1$};
\draw[arrow] (emb_2) -- (seq_representation) node[midway,right] {$\vx_2$};
\draw[arrow] (emb_T) -- (seq_representation) node[midway,right=0.5cm] {$\vx_T$};

\node[rectangle, minimum width=5cm, minimum height=1cm, draw=red!50, fill=red!10, text width=5cm, text centered, below=2cm of seq_representation] (dense) {{\bf Dense Layer, $f_c$} \\ (Sigmoid)};

\draw[arrow] (seq_representation) -- (dense) node[midway,right] {$\vz = f_m(X=\{\vx_1,\vx_2,\dots,\vx_T\})$};

\node[rectangle, dashed, draw=black, below=1cm of dense] (output) {\bf Loss function \\ $\ell(f_c(\vz),y)$};
\draw[arrow] (dense) --  (output) node[midway,right] (ar_node) {$f_c(\vz) = \sigma(\param^\top \vz + b)$};
\node[rectangle, dashed, draw=black, right=2cm of ar_node] (true_label) {{\bf Label:} $y$};
\draw[arrow] (true_label) |-  (output);

\end{tikzpicture}

%% file: method.tex
\section{URL representation schemes}
\label{sec:method}

This section introduces the four URL representation models proposed in our work. These models can be divided in two categories: i) one-hot encoding, and ii) embedding representation. There is no need for learning in the case of one-hot encoding. 
On the other hand, the embedding representations of URL tokens are learned in advance and then used to form the final URL representation. A representation is also learned for the so-called ``rare'' and ``none'' tokens, respectively. A token is considered ``rare'' if it is present less than a predefined number (we set it equal to $20$) of times in our data. % (a different threshold can be used for each token position).  
We denote the token embedding vectors as $\ve$.
\paragraph{\textbf{One-hot encoding}} First, we introduce a variant of the standard one-hot encoding for representing URLs that is our baseline. As already mentioned, the token representations are used to get the URL representation. Therefore, in our case, the cardinality of the one-hot encoding is equal to the number of all possible tokens appearing in our data.
Given a one-hot encoding $\{\ve_t\}_{t=1}^{n\_tokens \leq 3}$, for each one of the tokens appearing in $url$, we take their average to encode it: $f_r(url) = \frac{1}{n\_tokens}\sum_{t=1}^{n\_tokens} \ve_t$.
\paragraph{\textbf{Embedding learning}}Despite its simplicity, the aforementioned one-hot encoding does not take into account the similarity between URLs, while on the same time the representation size grows linearly with the corpus. 
In fact, one-hot encoding is sparse (curse of dimensionality) and it is not able to capture the distance between individual urls.
To tackle this problem, we propose three representation schemes inspired by the idea of \emph{Word2Vec} \cite{Mikolov:2013:w2v_2}.% used with success on various NLP tasks. 
More specifically, our representation schemes use the \emph{skip-gram} model that given a target word (URL in our case) tries to predict its context words. More formally, the \emph{skip-gram} model tries to find word representations that can be used for predicting the words in its neighborhood. Therefore, given a sequence of words (URLs) ${url_1, \dots, url_T}$, our objective is the maximisation of the average log probability
\begin{equation}
    \frac{1}{T} \sum_{t=1}^{T} \sum_{-c \leq j \leq c, j \neq 0} \log p(url_{t+j} | url_t),
    \label{eq:skip-gram}
\end{equation}
where $c$ specifies the neighborhood of target URL. 
The conditional probability is defined by using the softmax function, as
$
    p(url_c | url_t) \defn \frac{\exp(\vx_c^\top \vx_t)}{\sum_{c' \in C} \exp(\vx_{c'}^\top \vx_t) },
$
where $\vx_c$ and $\vx_t$ are the representations for context and target URLs respectively, and $C$ is the set of unique words. 

As the direct optimisation of Eq.~\ref{eq:skip-gram} is computationally expensive, we are adopting the \emph{negative sampling} approach \cite{Mikolov:2013:w2v_2}. %introduced by \cite{Mikolov:2013:w2v_2} that is a simplified version of the Noise Contrastive Estimation \cite{NCE:GutmannH12}. 
In negative sampling, we treat the word's representation learning as a binary classification task where we try to distinguish %(using logistic regression) 
the target-context pairs of words presented on the training data from those that are not. 
Following the suggestions of \cite{Mikolov:2013:w2v_2}, for each positive pair we are creating $k$ negative (target-context) pairs.
%In our empirical analysis (see Sec.~\ref{sec:experiments}) we consider two different \emph{\{pos:neg\}} ratios ($\{1{:}1\}$ and $\{1{:}4\}$) to examine their impact on the representation learning.

% The main difference between the proposed representation learning architectures with the original one, is that our URLs are composed of one, two, or three tokens.
In contrast to the original \emph{Word2Vec} model, the proposed representation learning architectures try to learn the tokens representations, instead of the URL representations directly. 
A token representation is also learned for the case of a \texttt{Pad} token. For instance, \url{https://en.wikipedia.org/} is mapped to [\texttt{en.wikipedia.org}, \texttt{Pad}, \texttt{Pad}]. Since URLs are padded, the number of tokens of each URL is equal to three. Then, by combining the token representations we form the final URL representation that will be used for the training of the conversion prediction classifier. Actually, the second phase of our model (conversion classifier) can be seen as a way to test the effectiveness of our representation models.  The main difference between the three proposed embedding representation models is how the token representations are combined to form the final URL embedding vector: 
% Fig~\ref{fig:skipgram} presents the general architecture of the representation model based on which the next three URL representation variants are trained.
\begin{itemize}[noitemsep,topsep=0.2pt,]
    \item \texttt{Domain\_only} representation uses only the representation of the first token to represent the URL, ignoring the representations of the other two tokens.
    \item \texttt{Token\_avg} representation takes the average of the token embedding vectors to represent the URL.
    \item \texttt{Token\_concat} representation concatenates the token embedding vectors to represent the URL. In this case, the dimension of the URL embedding vector is three times the dimension of the token embedding vectors.
\end{itemize}
% \emph{Domain only} representation uses only the representation of the first token to represent the URL. Actually, we ignore the representations of the other tokens.

% \emph{Token Average} representation takes the average of the token embedding vectors to represent the URL.

% \emph{Token Concatenation} representation concatenate the token embedding vectors to represent the URL. In this case, the dimension of the URL embedding vector is three times the dimension of token embedding vector.

For instance, let $\{\ve_t\}_{t=1}^3$ be the token embedding vectors of the three tokens presented on $url = [token_1, token_2, token_3]$. Then, the \texttt{Domain\_only} representation $\vx$ is equal to $\ve_1$, the \texttt{Token\_avg} representation is equal to $\frac{1}{3}\sum_{t=1}^3 \ve_t$, and the \texttt{Token\_concat} representation is given as $[\ve_1^\top, \ve_2^\top, \ve_3^\top]^\top$. A graphical illustration of the proposed embedding learning architecture is provided at appendix.

%% file: experiments.tex
%auto-ignore
\section{Experiments}
\label{sec:experiments}

% This section presents the results of our empirical analysis. 
% First, we introduce the datasets used for training and testing our models as well as the experimental settings.
% Then, we visualise the ability of the representation models to automatically organize URLs and learn the relationships between them. 
% Furthermore, we present the performance of the ten proposed prediction conversion models, where comparisons have been conducted on five different advertisers. 
% % In total, we examine the next ten models: i) \texttt{One\_hot/LR}, ii) \texttt{Domain\_only/LR}, iii) \texttt{Domain\_only/DLR} iv) \texttt{Domain\_only/RNN}, v) \texttt{Token\_avg/LR}, vi) \texttt{Token\_avg/DLR}, vii) \texttt{Token\_avg/RNN}, viii) \texttt{Token\_concat/LR}, ix) \texttt{Token\_concat/DLR}, and x) \texttt{Token \_concat/RNN}.
% The first part of the name of each model specifies the type of URL representation used, while the second one indicates the classifier type.
% As aforementioned, the \texttt{One\_hot/LR} is our baseline.
% %Due to the high dimensionality of \texttt{One\_hot} representation, we do not consider the \texttt{One\_hot/RNN}  \texttt{One\_hot/RNN} model.

% \subsection{Datasets}
% \label{sec:datasets}
This section presents the results of our empirical analysis. 
A real-world RTB dataset was used to train and analyse the performance of the ten proposed prediction models. We built our dataset by using the auction system logs from campaigns launched in France. It should be also mentioned that our dataset is anonymised, and only visited URLs are considered. In this way, each record of the dataset corresponds to a chronologically ordered sequence of visited URLs along with a binary label (specific to the advertiser) that indicates whether or not a conversion has happened on the advertiser's website on the next day.
More precisely, the data composed of sequences of URLs along with their labels of three successive dates, $\CD_d$, $\CD_{d+1}$, and $\CD_{d+2}$, where $\CD_d$ is used for learning representations, and $\CD_{d+1}$ and $\CD_{d+2}$ for training and testing the prediction models, respectively. 
% Therefore, our dataset composed by three data, $\CD_d, \CD_{d+1}, \CD_{d+2}$, where $\CD_d$ is used for learning representations, and $\CD_{d+1}$ and $\CD_{d+2}$ for training and testing the prediction models, respectively. 
% The binary labels are not needed for the training of the URL representation model.
Moreover, the maximum length of a URL sequence is set equal to $500$, where only the most recently visited URLs are kept in each sequence, $T_n \leq 500 , \forall n \in [1,N]$. 
In total we examine the performance of the models on five advertisers, belonging to four different categories: banking, e-shop, newspaper, and telecommunications (see Table~\ref{tab:adv_stats}).
%Table~\ref{tab:adv_stats} presents the number of converted and non-converted records on training and testing data for the five advertisers. 
Details about the statistics of the data are provided in the appendix [\href{https://drive.google.com/drive/folders/1SrxT34qTux5WreomD11RArftJZvHS68P?usp=sharing}{url\_link}] (\url{https://bit.ly/3cSNFXU}). 

\begin{table}[t]
\centering
\caption{Number of converted vs. non-converted records for each one of the $5$ advertisers on the training and testing data.}
\label{tab:adv_stats}
\vspace{-1em}
\resizebox{0.45\textwidth}{!}{%
\begin{tabular}{c|c|c}
\hline
\hline
{\bf Advertiser Category} & {\bf Training} ($5,452,577$)  & {\bf Testing} ($7,164,109$) \\
 \hline
 {\bf Banking} & $(3,746 - 5,448,831)$ & $(8,539 - 7,155,570)$ \\
 \hline
 {\bf E-shop} & $(1,463 - 5,451,114)$ & $(1,821 - 7,162,288)$ \\ 
 \hline
 {\bf Newspaper\_1} & $(1,406 - 5,451,171)$ & $(2,923 - 7,161,186)$ \\
 \hline
 {\bf Newspaper\_2} & $(1,261 - 5,451,316)$ & $(1,291 - 7,162,818)$ \\
 \hline
 {\bf Telecom} & $(1,781 - 5,450,796)$ & $(2,201 - 7,161,908)$ \\ 
 \hline
\end{tabular}%
} \vspace{-1em}
\end{table} 

\subsection{Settings}
All our predictors assume the existence of a URL representation model in order to vectorise the sequence of URLs for each one of the dataset records. 
Our baseline, \texttt{One\_hot/LR}, represents the URL sequences using a one-hot encoding vector of size ${193,409}$, where the first two entries correspond to the ``unknown'' and ``rare'' tokens, respectively. 
% It means that we have $193,407$ non-rare tokens, while the first two spots of the vector are reserved for the unknown and rare tokens. 
The other models rely on an already trained embedding matrix. % (representation learning phase).
Each token is embedding into a $100$-dimensional vector. 
The first three rows correspond to the ``unknown'', ``rare'', and \texttt{Pad} tokens, respectively.
The rest rows contain the embedding vector of all non-rare tokens observed in the dataset $\CD_d$. % (used for the training of representation model). 
%A token is considered as rare if it is appeared less than $20$ times in the dataset.
The number of non-rare tokens is $22,098$ for the \texttt{Domain\_only}, and $187,916$ for both the \texttt{Token\_Average} and \texttt{Token\_Concatenation}. % representations. 
% Therefore, the size of the URL embedding vectors of the \emph{Domain only}, \emph{Token Average} and \emph{Token Concatenation} representations is $100$, $100$, and $300$, respectively. 
To train representations, we have considered two different \emph{\{pos:neg\}} ratios ($\{1{:}1\}$ and $\{1{:}4\}$). Due to page limitations, only the results of the $\{1{:}4\}$ ratio are presented in this manuscript.  See supplementary material [\href{https://drive.google.com/drive/folders/1SrxT34qTux5WreomD11RArftJZvHS68P?usp=sharing}{url\_link}] for a full comparison between these two ratios.

The number of units of the hidden dense layer (dimensionality of its output space) of \texttt{DLR} model is set to $30$. 
Furthermore, the number of hidden units of LSTM is set to $10$ on the \texttt{RNN} model. 
A dense layer with a sigmoid activation function is applied at the end of each one of the three mapping functions $f_m$ (``average'', ``dense'', ``LSTM'') in order to form a binary classifier. 
Through our empirical analysis we have observed that the \texttt{DLR} and \texttt{RNN} prediction models are prone to overfitting. 
To enhance the generalization capabilities of these two models, we are using dropout that is set to $0.5$. 
To be more precise, a dropout layer is added right after the $f_m$ layer of the \texttt{DLR} model, while both the dropout and the recurrent\_dropout parameter of LSTM layer are set equal to $0.5$ on the \texttt{RNN} model. %\cite{Dropout:srivastava14a} that is a simple, computational inexpensive but on the same time powerful regularisation technique. 
%The fraction of the units of the dense hidden layer and the LSTM to drop on the \texttt{DLR} and \texttt{RNN} models, respectively, is set to $0.5$. 

For the training of all representation and prediction models, the mini-batch stochastic optimization has been applied by using Adam optimizer %\cite{Adam:KingmaB14} 
with the default \href{https://www.tensorflow.org/guide/effective_tf2}{Tensorflow $2.0$}
%\footnote{\href{https://www.tensorflow.org/guide/effective_tf2}{Tensorflow 2.0} version used for building our models.} 
settings (i.e., lr=$0.001$, $\text{beta}_1=0.9$, $\text{beta}_2=0.999$). 
More precisely, to train the representation models we are doing one full pass over the whole data that is divided to $200$ parquet files.
%In this way, we guarantee that each URLs presented in our data used for training the embedding matrix. 
%Otherwise, a random initialised embedding vector will be assigned to each URL that have not been used through the training phase. 
The total number of epochs is $200$, equal to the number of parquet files.
At each epoch we are producing the positive and negative pairs based on the data contained on a single parquet file and are feeding them to the representation model. 
%As each parquet file contains various length sequences of different users, the number of steps is slightly different at each epoch.
On the other hand, the size of batches for training prediction models is $64$, while the number of epochs and number of steps per epoch is set to $100$ in both cases. 
To tackle the problem of our unbalanced dataset (see Table~\ref{tab:adv_stats}), the ratio of positive and negative records is $\{1{:}1\}$ in the batches used for the training of the classifiers.
%Logistic loss is used as the loss function for the training of all the models at each prediction task. 

\subsection{Results}
\label{sec:results}

\begin{figure}[t]
    \centering
    \includegraphics[width=1\linewidth]{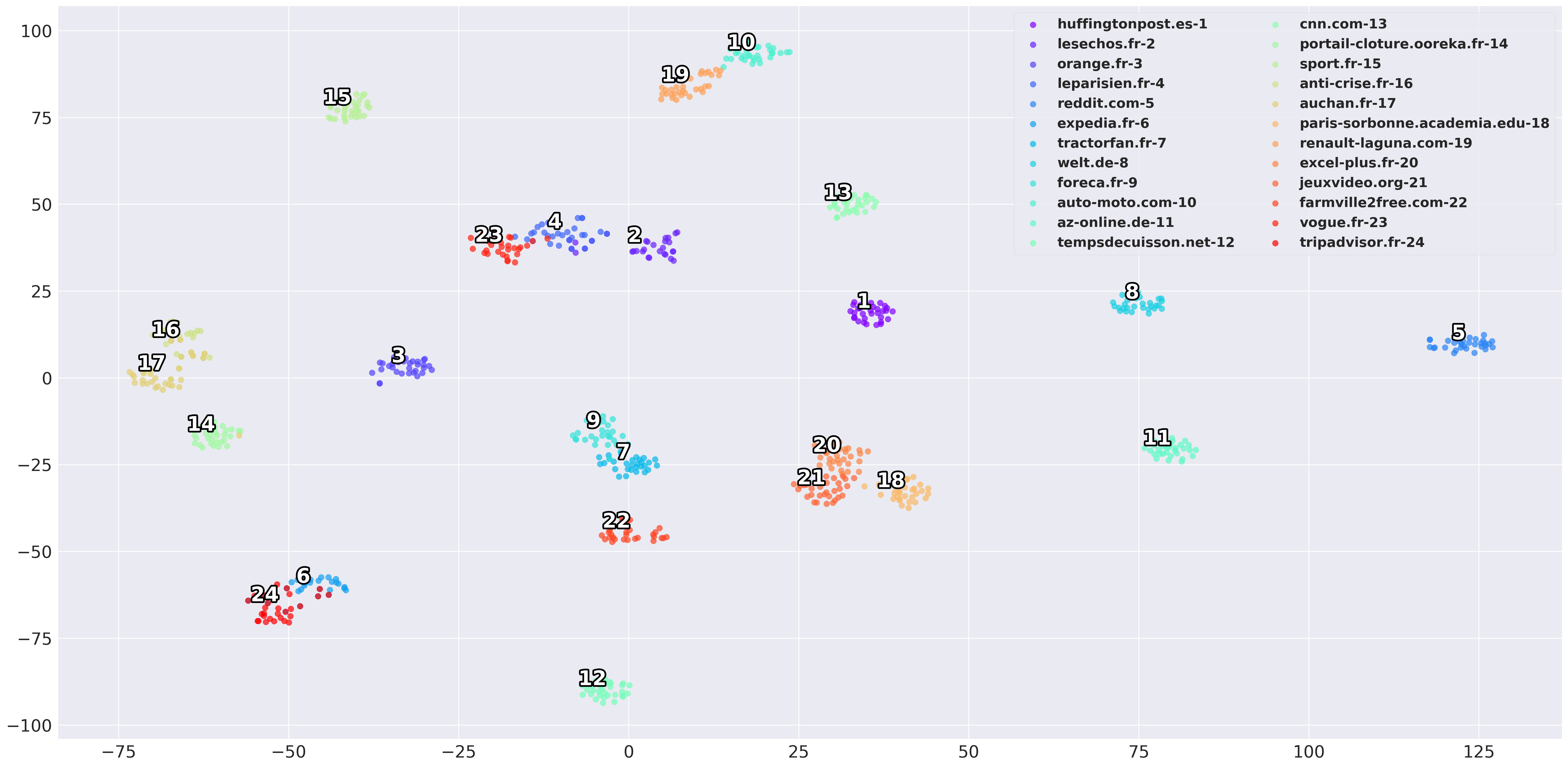}
    \caption{t-SNE visualization of the thirty closest neighbors of $24$ different domains.} %The colors of the points indicate the closest domain of each URL.}
    \label{fig:url_representation}
    \vspace{-1em}
\end{figure}

In this section, we formally present the results of our empirical analysis. 
Firstly, to get an intuition about the ability of the three introduced URL embedding schemes (Sec.~\ref{sec:method}) to group together URLs that belong to the same category (i.e., sports, news, etc.), we will visualise (Fig.~\ref{fig:url_representation}) the embedding vectors of $24$ selected domains along with those of the thirty closest URLs of each one of them. 
To be more precise, we are using the embedding matrix learned by the \texttt{Domain\_only} model.
Cosine similarity has been used to measure the similarity between the embedding vectors of two URLs. 
To project the original high-dimensional embedding vectors on a 2-dimensional space, we apply the Barnes-Hut t-SNE
% \footnote{We have used t-SNE provided by \href{https://scikit-learn.org/stable/modules/generated/sklearn.manifold.TSNE.html}{scikit-learn} library with its perplexity to be equal to $15$ and using cosine similarity metric.}
algorithm \cite{tsne:vandermaaten14a}.% that is able to preserve the pairwise distance between points (i.e. nearby points correspond to similar objects and that distant points correspond to dissimilar objects). 

\begin{table}[t]
\begin{center}
\caption{The $10$-nearest neighbors of $24$ different domains according to our trained \texttt{Domain\_only} representation model.}
\label{tab:nearest_neighbors}
\resizebox{1.\columnwidth}{!}{
\input{./nearest_neighbors_10.tex}
}
\end{center}
\vspace{-1em}
\end{table}

To be guaranteed that the URLs belong to the same category with that of their closest domain, Table~\ref{tab:nearest_neighbors} provides the $10$-nearest URLs for each one of the $24$ domains (see \href{https://drive.google.com/drive/folders/1SrxT34qTux5WreomD11RArftJZvHS68P?usp=sharing}{appendix} for the full list of the $30$-nearest neighbors). 
For instance, all URLs that are on the neighborhood of \href{https://expedia.fr}{expedia.fr} are about \emph{travelling}. Moreover, all the neighbors of \href{https://sport.fr}{sport.fr} ($16$) are URLs about \emph{sports}. 
The visualization of Fig.~\ref{fig:url_representation} illustrates the ability of our model to produce semantically meaningful embeddings. 
Actually, it becomes apparent that we are getting $24$ clearly distinguished clusters. % as the number of selected domains.  
Moreover, we can see that the clusters of `similar' domains are also close on the embedding space. For instance, the URLs embeddings of clusters ($10$) [\href{https://auto-moto.com}{auto-moto.com}] and ($19$) [\href{https://renault-laguna.com}{renault-laguna.com}] are close as they are related to the \emph{automobile} category. The same also holds for the  URLs embeddings of clusters ($2$) [\href{https://lesechos.fr}{lesechos.fr}], ($4$) [\href{https://leparisien.fr}{leparisien.fr}], and ($23$) [\href{https://vogue.fr}{vogue.fr}] that belong to the \emph{news} category.

\begin{figure*}[t]
    \centering
    \begin{subfigure}[t]{.163\textwidth}
    \includegraphics[width=1.\linewidth]{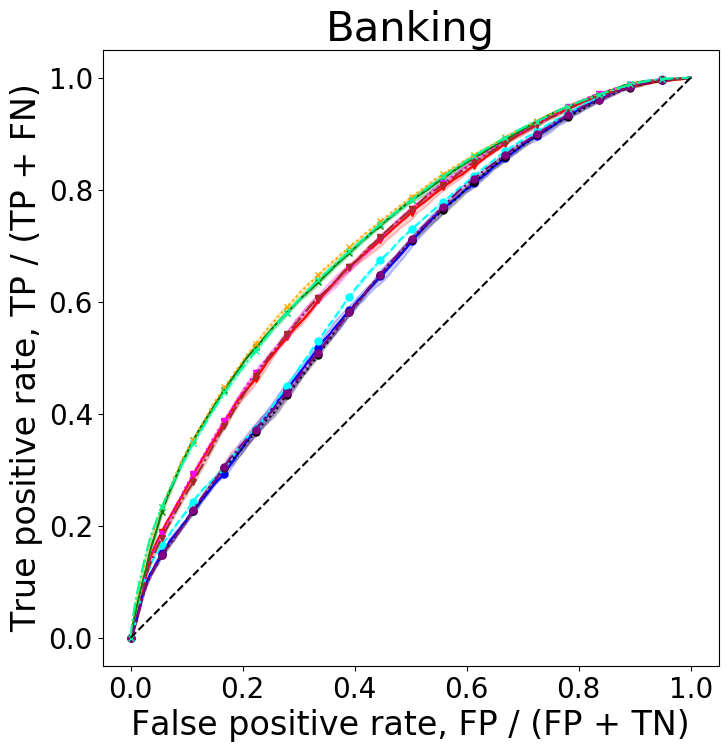}
    \end{subfigure} 
    \begin{subfigure}[t]{.163\textwidth}
    \includegraphics[width=1.\linewidth]{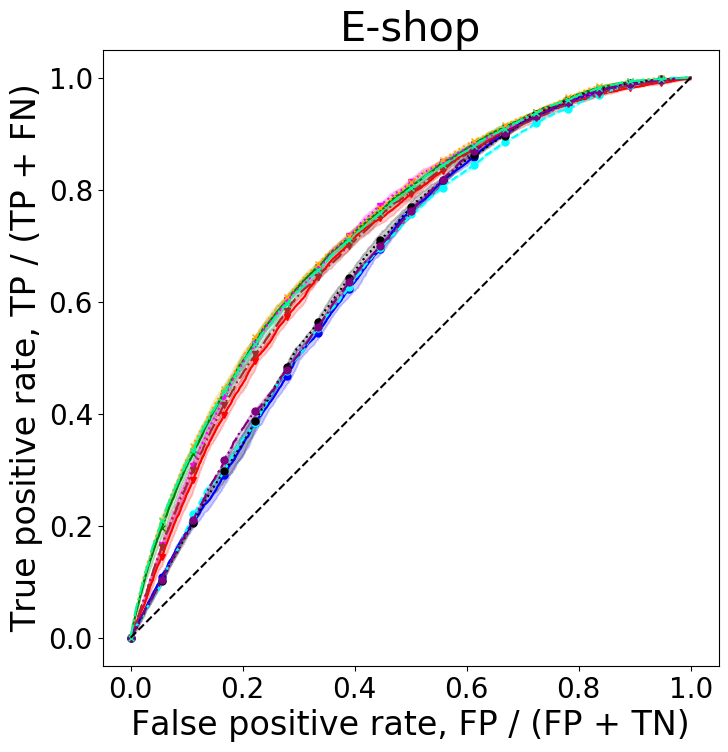}
    \end{subfigure} 
    \begin{subfigure}[t]{.163\textwidth}
    \includegraphics[width=1.\linewidth]{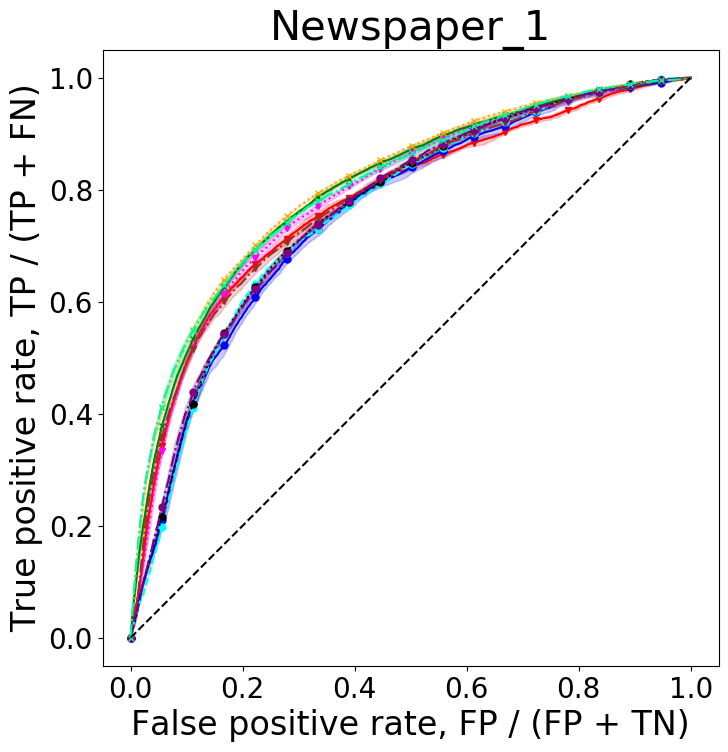}
    \end{subfigure} 
    \begin{subfigure}[t]{.163\textwidth}
    \includegraphics[width=1.\linewidth]{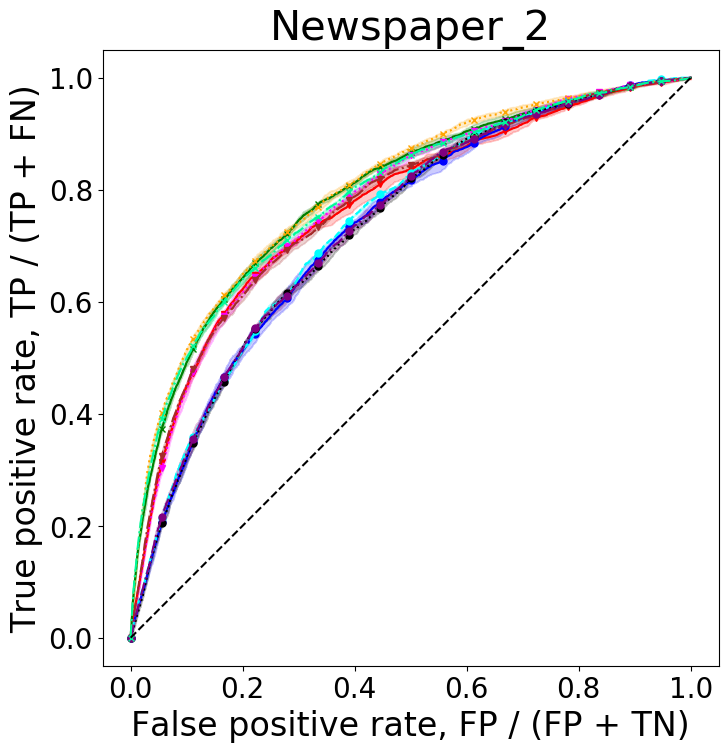}
    \end{subfigure} 
    \begin{subfigure}[t]{.163\textwidth}
    \includegraphics[width=1.\linewidth]{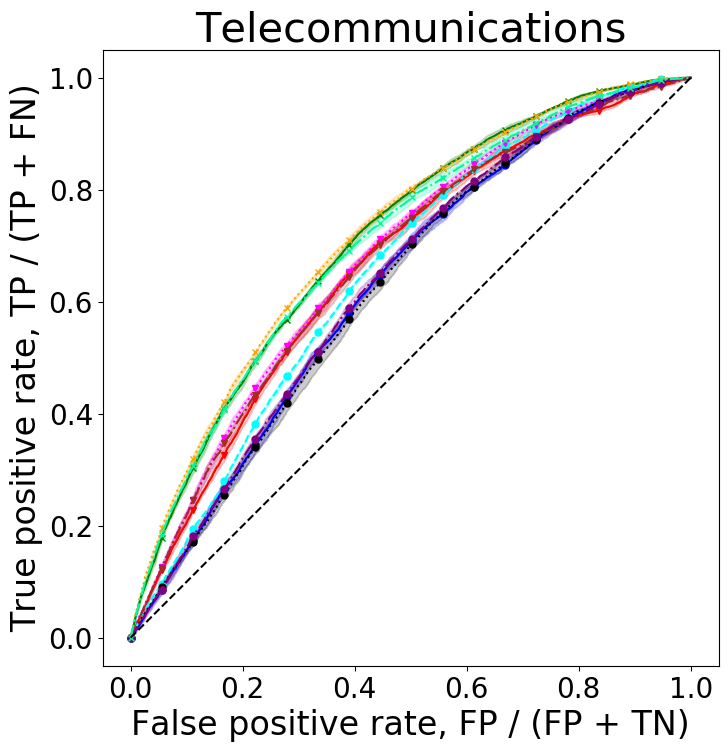}
    \end{subfigure} 
    \begin{subfigure}[t]{.163\textwidth}
    \includegraphics[width=1.\linewidth]{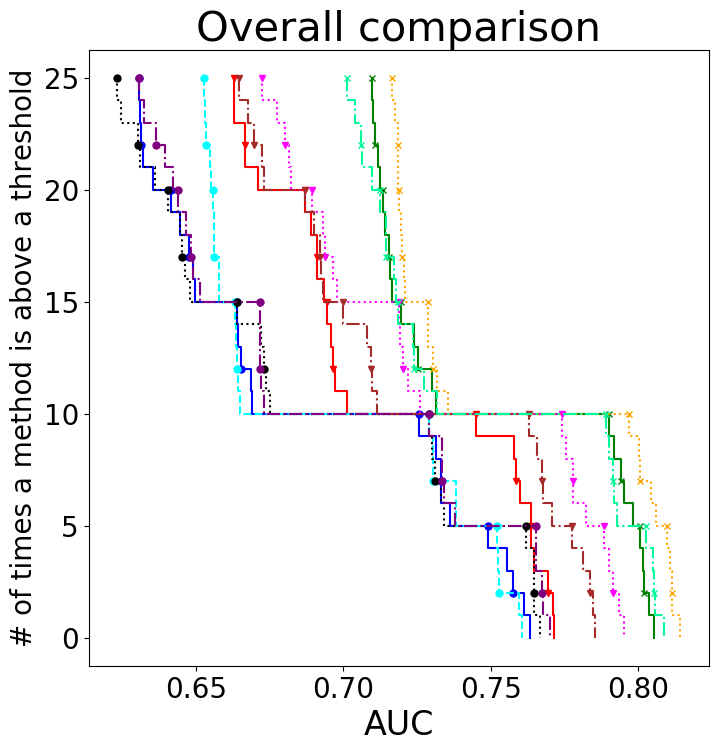}
    \end{subfigure}
    
    \begin{subfigure}[t]{.6\textwidth}
    \includegraphics[width=1.\linewidth]{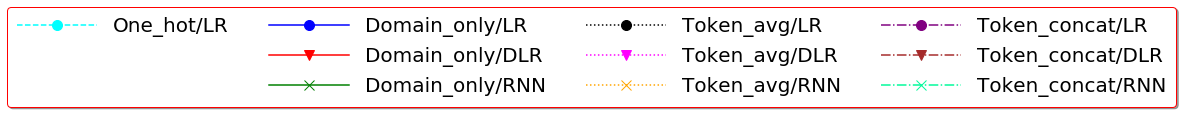}
    \end{subfigure} 
    \vspace{-0.5em}
    \caption{Average ROC curves of the ten conversion prediction models on the five advertisers. Shaded regions represent the std over $5$ independent runs.  The bottom right plot presents the AUC for each one of the $25$ runs ($5$ advertisers $\times$ $5$ independent runs for each advertiser) of each model. The $\bullet$, $\blacktriangledown$ and $\times$ marks indicate the \texttt{LR}, \texttt{DLR} and \texttt{RNN} classification models, respectively.}
    \label{fig:empirical_results}
    \vspace{-.5em}
\end{figure*}

Next, we formally present the numerical results of the ten prediction models on five different advertisers.
The first part of the name of each model specifies the type of URL representation used, while the second one indicates the classifier type.
To evaluate and compare the effectiveness of the models we are using the area under ROC curve (AUC) metric. % that has been used extensively for evaluating user response prediction models in RTB.
More specifically,  we consider the average ($\%$) AUC across five independent runs (see Table~\ref{tab:AUC}), where each run corresponds to a specific seed used for the models initialisation.
Moreover, Figure~\ref{fig:empirical_results} illustrates the average  ROC curves of the prediction models for each advertiser.
%We also plot the standard deviation (shaded regions) of the runs. 
%Apart from the five plots that correspond to the five advertisers, 
The right plot of Fig.~\ref{fig:empirical_results} illustrates the AUC of the models on the $25$ independent runs ($5$ advertisers $\times$ $5$ runs for each advertiser)). 
% To be more precise, we are testing the performance of $25$ prediction models for each type of model in total, one for each advertiser and each independent run.

\begin{table}[t!]
\begin{center}
\caption{Avg ($\%$) and std of the area under ROC curves ($5$ independent runs) of the $10$ prediction models on $5$ advertisers.}
\vspace{-1em}
\label{tab:AUC}
\resizebox{.49\textwidth}{!}{
\huge
\input{./pos_neg_ratio_4_average_auc.tex}
}\end{center}
\vspace{-1.6em}
\end{table}

Based on the results presented in Table~\ref{tab:AUC}, the \texttt{Token\_avg/RNN} model is more effective in predicting user's conversions compared to the rest models. 
Precisely, the \texttt{Token\_avg/RNN} model has the highest average AUC in all advertisers.
All models achieve their highest performance on the newspaper advertisers. 
It is also worth noticing that the performances of \texttt{Domain\_only/LR}, \texttt{Token\_avg/LR}, and \texttt{Token\_concat/LR} are highly competitive, compared to our baseline, \texttt{One\_hot/LR}.
More specifically, \texttt{Token\_concat/LR} clearly outperforms \texttt{One\_hot/LR} in $2$ out of $5$ advertisers (E-shop, Newspaper\_1), and has slightly better performance in one of them (Newspaper\_2).
That remark validates our claims that the proposed representation models produce meaningful embeddings, by distinguishing URLs of the same category and placing them close to the embedding space. 
Taking a closer look at the standard deviations, we can see that the performance of \texttt{One\_hot/LR} is quite stable performance in the three advertisers. 
This was expected as the \texttt{One\_hot} representation is identical over all runs, and therefore the only variation of the performance of \texttt{One\_hot/LR} comes only from the training of the \texttt{LR} classifier.
The performance of \texttt{LR} model is almost the same for each representation model. 
The same also holds for \texttt{DLR} and \texttt{RNN} model where its performance is more or less the same in the cases of  \texttt{Domain\_only} and  \texttt{Token\_concat}, and slightly better in the case of \texttt{Token\_avg}.
On the other hand, \texttt{DLR} performs significantly better when it is combined with \texttt{Token\_avg} and \texttt{Token\_concat}, with \texttt{Token\_avg} to be more preferable (around $1\%$ gain). % compared to \texttt{Token\_concat}).
% It means that the representations of the last two URL tokens does not convey a lot of additional information and that the \textt{Domain\_token} representation adequately describes the proximity between two URLs in the embedding space. 
% Another reason could be that most of the tokens (especially those appeared on the last two places) presented on datasets $\CD_{d+1}$ and $\CD_{d+2}$ are not presented on $\CD_{d}$ that has been used for training the representation models. 
% In that case, they are treated as ``unknown'' tokens and a random initialized embedding vector assigned to them. This vector in not trained as each token is considered as ``known'' in the training phase.

Let us now compare the impact of the type of classifier on the performance of the prediction model. 
%Recall that \texttt{DLR} considers the dependencies among the feature of the embedding vector, while \texttt{RNN} is sensitive to the order of the sequence of URLs in contrast to the other models.  
The overall comparison presented at Fig.~\ref{fig:empirical_results} demonstrates that both \texttt{DLR} and \texttt{RNN} performs significantly better compared to \texttt{LR}, with the \texttt{RNN} to be the best one.
% More specifically, the \texttt{RNN} classifier achieves to predict the conversion probabilities with higher accuracy compared to the rest two models. 
More precisely, the AUC of  \texttt{RNN} is around ${\sim}7\%$ and ${\sim}3\%$ higher compared to those of \texttt{LR} and \texttt{DLR}, respectively.
This means that the consideration of the chronological order in which the URLs appeared on the sequence is of high importance.
On the other hand, choosing \texttt{DLR} over \texttt{LR} improves around ${\sim}4\%$ the performance of the prediction models independent to the representation model. 

To sum up, the main conclusions of our empirical analysis are: i) all three proposed URL embedding models are able to learn high-quality vector representations that precisely capture the URL relationships, ii) the performance of the \texttt{LR} model is relatively invariant to the selection of the representation model, iii) among the three representation models, \texttt{Token\_avg} is more adequate to capture the relationship between URLs, with the \texttt{Token\_concat} second best, iv) the consideration of the chronological order of the visited URLs (\texttt{RNN}) and the learning of dependencies among the embedding features (\texttt{DLR}) are also of high importance as both improve significantly the performance of the prediction model. %, with the \texttt{RNN} model performing better. 
\vspace{-0.05em}

%% file: nearest_neighbors_10.tex
%auto-ignore
\begin{tabular}{C{3.5cm}|p{16.4cm}}
\hline
\hline
\Large
{\bf Domain} & \begin{minipage}{1.4\columnwidth} \center \bf $10$-nearest neighbors \end{minipage} \\
\hline
{\noindent \bf \color{blue} huffingtonpost.es} & cope.es; m.eldiario.es; okdiario.com; verne.elpais.com; blogs.elconfidencial.com; vozpopuli.com; elespanol.com; smoda.elpais.com; libertaddigital.com; cadenaser.com \\
\hline
{ \noindent \bf \color{blue} lesechos.fr} & latribune.fr; afrique.latribune.fr; business.lesechos.fr; bfmbusiness.bfmtv.com; financedemarche.fr; challenges.fr; investopedia.com; actufinance.fr; lopinion.fr; contrepoints.org \\
\hline
{\noindent \bf \color{blue} orange.fr} & actu.orange.fr; lemoteur.orange.fr; messagerie.orange.fr; login.orange.fr; finance.orange.fr; sports.orange.fr; meteo.orange.fr; tendances.orange.fr; programme-tv.orange.fr; news.orange.fr \\
\hline
{\noindent \bf \color{blue} leparisien.fr} & cnews.fr; atlasinfo.fr; lefigaro.fr; lejdd.fr; jforum.fr; marianne.net; video.lefigaro.fr; tendanceouest.com; bladi.net; observalgerie.com \\
\hline
{\noindent \bf \color{blue} reddit.com} &  imgur.com; old.reddit.com; askreddit.reddit.com; pcgamer.com; anime.reddit.com; france.reddit.com; gamefaqs.gamespot.com; totalwar.reddit.com; nintendoswitch.reddit.com; gaming.reddit.com \\
\hline
{\noindent \bf \color{blue} expedia.fr} & momondo.fr; skyscanner.fr; kayak.fr; fr.lastminute.com; fr.hotels.com; flights-results.liligo.fr; ebookers.fr; esky.fr; opodo.com; secure.lastminute.com \\
\hline
{\noindent \bf \color{blue} tractorfan.fr} &  discountfarmer.com; forum.farm-connexion.com; angleterre.meteosun.com; songs-tube.net; materieltp.fr; assovttroc.clicforum.fr; opel-mokka.forumpro.fr; spa-du-dauphine.fr; vanvesactualite.blog4ever.com; calcul-frais-de-notaire.fr \\
\hline
{\noindent \bf \color{blue} welt.de} & zeit.de; sueddeutsche.de; faz.net; tagesspiegel.de; sport1.de; kicker.de; saarbruecker-zeitung.de; tz.de; bild.de; sportbild.bild.de \\
\hline
{\noindent \bf \color{blue} foreca.fr} & my-meteo.com; fr.meteovista.be; fr.tutiempo.net; meteopassion.com; de.sat24.com; nosvolieres.com; meteo-sud-aveyron.over-blog.com; xn--mto-bmab.fr; palombe.com; calculerdistance.fr \\
\hline
{\noindent \bf \color{blue} auto-moto.com} & caradisiac.com; largus.fr; news.autojournal.fr; test-auto.auto-moto.com; auto-mag.info; feline.cc; motorlegend.com; essais.autojournal.fr; automobile-magazine.fr; turbo.fr \\
\hline
{\noindent \bf \color{blue} az-online.de} & abountifulkitchen.com; thesurvivalgardener.com; leinetal24.de; brittanyherself.com; symbols.com; ourpaleolife.com; msl24.de; milliondollarjourney.com; arthritis-health.com; thehollywoodunlocked.com \\
\hline
{\noindent \bf \color{blue} tempsdecuisson.net} & cuisine-facile.com; temps-de-cuisson.info; yummix.fr; aux-fourneaux.fr; cuisinenligne.com; audreycuisine.fr; mamina.fr; uneplumedanslacuisine.com; cnz.to; ricardocuisine.com \\
\hline
{\noindent \bf \color{blue} cnn.com} & us.cnn.com; stadiumtalk.com; thedailybeast.com; itpro.co.uk; uk.reuters.com; euronews.com; theargus.co.uk; theatlantic.com; thedailymash.co.uk; trendscatchers.co.uk \\
\hline
{\noindent \bf \color{blue} portail-cloture.ooreka.fr} & bricolage-facile.net; mur.ooreka.fr; pierreetsol.com; bricolage.jg-laurent.com; abri-de-jardin.ooreka.fr; aac-mo.com; fr.rec.bricolage.narkive.com; decoration.ooreka.fr; piscineinfoservice.com; bricoleurpro.com \\
\hline
{\noindent \bf \color{blue} sport.fr} & infomercato.fr; parisfans.fr; topmercato.com; vipsg.fr; footradio.com; mercatofootanglais.com; le10sport.com; buzzsport.fr; footparisien.com; foot-sur7.fr \\
\hline
{\noindent \bf \color{blue} anti-crise.fr} & cfid.fr; forum.anti-crise.fr; gesti-odr.com; echantillonsclub.com; plusdebonsplans.com; cataloguemate.fr; promoalert.com; argentdubeurre.com; madstef.com; forum.madstef.com \\
\hline
{\noindent \bf \color{blue} auchan.fr} & but.fr; conforama.fr; vente-unique.com; rueducommerce.fr; fr.shopping.com; cdiscount.com; touslesprix.com; promobutler.be; webmarchand.com; mistergooddeal.com \\
\hline
{\noindent \bf \color{blue} paris-sorbonne.academia.edu} & flux-info.fr; elmostrador.cl; makaan.com; univ-montp3.academia.edu; e-lawresources.co.uk; babycenter.com; newocr.com; insight.co.kr; grandes-inventions.com; police-scientifique.com \\
\hline
{\noindent \bf \color{blue} renault-laguna.com} & megane3.fr; gps-carminat.com; megane2.superforum.fr; lesamisdudiag.com; car-actu.com; diagnostic-auto.com; r25-safrane.net; lesamisdelaprog.com; forum.autocadre.com; renault-clio-4.forumpro.fr \\
\hline
{\noindent \bf \color{blue} excel-plus.fr} & tech-connect.info; thehackernews.com; lecompagnon.info; panoptinet.com; slice42.com; aliasdmc.fr; astuces.jeanviet.info; nalaweb.com; patatos.over-blog.com; jiho.com \\
\hline
{\noindent \bf \color{blue} jeuxvideo.org} & alsumaria.tv; minecraft-zh.gamepedia.com; infovisual.info; everyonepiano.com; footstream.live; memedroid.com; darkandlight.gamepedia.com; mbti.forumactif.fr; gachagames.net; honga.net \\
\hline
{\noindent \bf \color{blue} farmville2free.com} & goldenlifegroup.com; fv2freegifts.org; juegossocial.com; fv-zprod-tc-0.farmville.com; fb1.farm2.zynga.com; zy2.farm2.zynga.com; gameskip.com; fv-zprod.farmville.com; megazebra-facebook-trails.mega-zebra.com; farmvilledirt.com \\
\hline
{\noindent \bf \color{blue} vogue.fr} & vanityfair.fr; vogue.com; vivreparis.fr; fr.metrotime.be; brain-magazine.fr; o.nouvelobs.com; parismatch.be; pariszigzag.fr; admagazine.fr; unilad.co.uk \\
\hline
{\noindent \bf \color{blue} tripadvisor.fr} & fr.hotels.com; cityzeum.com; voyages.michelin.fr; lonelyplanet.fr; monnuage.fr; voyageforum.com; rome2rio.com; toocamp.com; virail.fr; partir.com \\
\hline
\end{tabular}

%% file: pos_neg_ratio_4_average_auc.tex
%auto-ignore
\begin{tabular}{c|P{2.5cm}|P{2.5cm}|P{2.9cm}|P{2.9cm}|P{2.5cm}}
    \hline
    \hline
    \diagbox{\bf Method}{\bf Adv} & {\bf Banking} & {\bf E-shop}	& {\bf Newspaper\_1} &	{\bf Newspaper\_2} & {\bf Telecom} \\
    \hline
    \texttt{One\_hot/LR} &  $65.7 \pm 0.093$ &  $66.4 \pm 0.053$ &  $75.5 \pm 0.379$ &  $73.3 \pm 0.400$ &   $65.4 \pm 0.085$ \\
    \hline
    \texttt{Domain\_only/LR} &  $64.6 \pm 0.300$ &  $66.6 \pm 0.217$ &  $75.7 \pm 0.507$ &  $73.2 \pm 0.345$ &   $63.2 \pm 0.168$ \\
    \hline
    \texttt{Domain\_only/DLR} &  $69.0 \pm 0.214$ &  $69.7 \pm 0.234$ &  $76.8 \pm 0.342$ &  $75.7 \pm 0.658$ &   $66.6 \pm 0.303$ \\
    \hline
    \texttt{Domain\_only/RNN} &  $71.4 \pm 0.144$ &  $72.6 \pm 0.422$ &  $80.3 \pm 0.168$ &  $79.4 \pm 0.281$ &   $71.2 \pm 0.250$ \\
    \hline
    \texttt{Token\_avg/LR} &  $64.5 \pm 0.241$ &  $67.2 \pm 0.390$ &  $76.4 \pm 0.152$ &  $73.1 \pm 0.184$ &   $62.9 \pm 0.468$ \\
    \hline
    \texttt{Token\_avg/DLR} &  $69.4 \pm 0.294$ &  $72.1 \pm 0.263$ &  $79.2 \pm 0.242$ &  $77.7 \pm 0.274$ &   $67.9 \pm 0.348$ \\
    \hline
    \texttt{Token\_avg/RNN} &  ${\bf71.9} \pm 0.082$ &  ${\bf73.1} \pm 0.246$ &  $\bf{81.2} \pm 0.153$ &  $\bf{80.2} \pm 0.322$ &   $\bf{71.8} \pm 0.153$ \\
    \hline
    \texttt{Token\_concat/LR} &  $64.8 \pm 0.241$ &  $67.2 \pm 0.060$ &  $76.7 \pm 0.179$ &  $73.4 \pm 0.273$ &   $63.6 \pm 0.425$ \\
    \hline
    \texttt{Token\_concat/DLR} &  $69.1 \pm 0.222$ &  $70.8 \pm 0.400$ &  $78.2 \pm 0.285$ &  $76.7 \pm 0.255$ &   $66.9 \pm 0.310$ \\
    \hline
    \texttt{Token\_concat/RNN} &  $71.5 \pm 0.224$ &  $72.5 \pm 0.460$ &  $80.5 \pm 0.192$ &  $79.1 \pm 0.130$ &   $70.5 \pm 0.278$ \\
    \hline
\end{tabular}

%% file: conclusions.tex
%auto-ignore
\section{Conclusions and Future directions}
\label{sec:conclusion}

In this paper, we considered the problem of user response prediction in display advertising. 
Ten conversion prediction models were proposed to predict user response based on their browsing history. 
To represent the sequence of visited URLs, four different compact URL representations were examined. 
The effectiveness of the proposed models has been experimentally demonstrated offline in a real-world RTB dataset for five different advertisers. 
The impact of the sequential dependency between users' visited URLs on the performance of the predictors has been also examined.
The main conclusions of our empirical analysis were that all three proposed representation models produce a meaningful URL representation, and considering the chronological order of the visited URLs by using \texttt{RNN} significantly improves the model's performance. 
In the future, we intend to propose an online version of our framework 
%(i.e., the representation model will be updated at the end of each day by using the data collected during the day) 
and to extend our empirical analysis to a real-world online scenario. 
% Finally, another interesting direction could be the usage of RepSet  \cite{RepSet:Skianis} that is able to generate representations of unordered variable-sized feature sets (sets of URL representations in our case), and can be considered as an alternative to the \texttt{LR}, \texttt{DLR} and \texttt{RNN} models.

%% file: appendix.tex
%auto-ignore
The purpose of this Supplementary Material is to provide more details about: i) the proposed representation and prediction conversion model architectures, ii) the statistics of data, and iii) present additional experimental results. 
First in Section~\ref{sec:architectures}, we illustrate graphically and explain the architecture of the prediction model. 
Next, Section~\ref{sec:data_stats} presents some statistics for the three data used for the representation training and for the training and testing of the representation and prediction models. Finally, Section~\ref{sec:sup_results} reports some additional experiments where we examine the impact of the \emph{\{pos:neg\}} ratio (used for representation training) on the learning of meaningful URL representations and on the performance of the final prediction model.

\section{Prediction conversion model architecture}
\label{sec:architectures}
\begin{figure}[h!]
    \centering
    \begin{subfigure}[t]{1.\columnwidth}
    \centering
    \scalebox{0.6}{\input{./representation_learning.tikz}}
    \caption{URL representation learning}
    \label{fig:rep_learning}
    \end{subfigure} 
    
    \begin{subfigure}[t]{1.\columnwidth}
    \centering
    \scalebox{.6}{\input{./model_learning.tikz}}
    \caption{Conversion prediction model learning}
    \label{fig:model_learning}
    \end{subfigure}
    \caption{URL representation and conversion classifier learning pipeline. The binary labels are not needed for training the URL representation model.}
    \label{fig:rep_model_learning_process}
\end{figure}
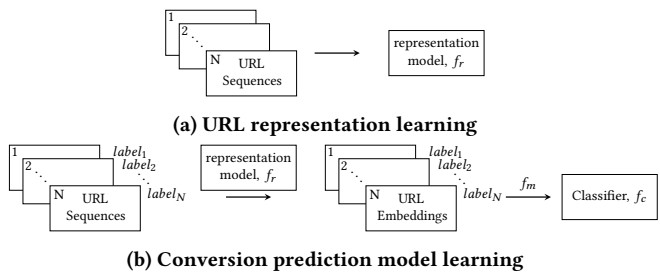

A graphical illustration of the proposed prediction conversion model architecture is presented in this section. 
The objective of the proposed model is to predict with high accuracy if a user will be converted or not given his browsing history.
Actually, the proposed conversion prediction scheme is composed of two consecutive training phases. The first one corresponds to the learning of the URL representations, while the second one corresponds to the training of a classifier. It should be mentioned that the training processes of these two models are independent.  
A high-level overview of the pipelines of these two training phases is illustrated at Fig.\ref{fig:rep_model_learning_process}. 

Figure~\ref{fig:model_architecture} shows the three basic `blocks' of our prediction model. 
First, the URL representation layer, $f_r{:}~url \rightarrow \vx$, maps a single $url$ into a vector on the embedding space. 
For this purpose, a URL representation model is trained in advance to learn meaningful $url$ representations.
Second, the mapping $f_m{:}~X \rightarrow \vz$ aggregates the URLs embeddings and produces a single compact representation $\vz$ for the sequence of the URLs. Three different mapping functions $f_m$ examined in our work: the first one ($f_m^{(1)}$) just returns the average of the URLs embedding vector, the second one ($f_m^{(2)}$) considers the dependencies among the features of the embedding vector returned by the first mapping function, and the third one ($f_m^{(3)}$) uses an LSTM network \cite{LSTM:HochreiterS97,Goodfellow:2016:DL}. Finally, standard logistic regression (dense layer with a sigmoid activation function) is applied to estimate the user conversion probability.

\begin{figure*}
    \centering
    \scalebox{.72}{\input{./skipgram.tikz}}
    \caption{The Skip-gram model architecture used for learning token embeddings. Only the (unknown) parameters of the red blocks are trainable. The dimensionality of the embedding matrices is equal to the number of tokens $\times$ the preferable size of the embedding space. }
    \label{fig:skipgram}
\end{figure*}
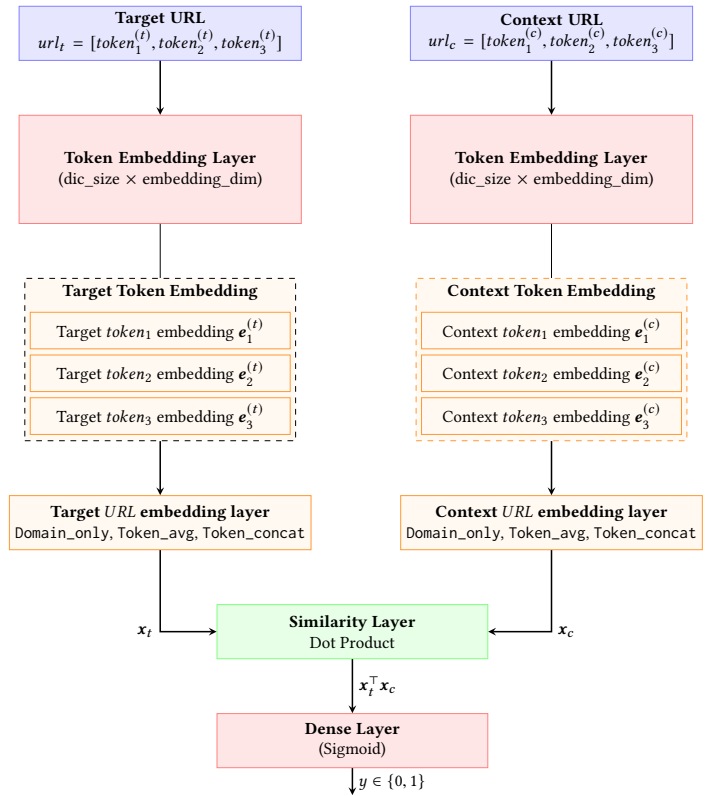

Figure~\ref{fig:skipgram} illustrates a variant of the negative sampling \emph{skip-gram} model \cite{Mikolov:2013:w2v_1,Mikolov:2013:w2v_2} used for learning $url$ semantic meaningful representations. In general, \emph{skip-gram} produces embedding vectors such that the embedding vector of $url_t$ is close to the embedding vectors of the URLs of its neighborhood.
Therefore, the url's representation learning is treated as a binary classification task where we try to distinguish the pairs of URLs presented close on the sequences. 
The main difference between our architecture with the standard one is the \emph{Target URL embedding layer} (and the \emph{Context URL embedding layer}). This layer combines the token representations and forms the final URL representation. 
Roughly speaking, the proposed representation model learns an embedding matrix where each row corresponds to a token appearing in the URLs of our data.
To form the URL representation, we are combining the embedding vectors of each token appearing in the URL.
In this work, we examine three different ways for combining token representations to form the final URL embedding vector (for more details, check Sec. 4 of the paper).

\begin{figure*}[t]
    \centering
    \scalebox{.675}{\input{./model_architecture.tikz}}
    \caption{Proposed conversion prediction model architecture. It consists of three main parts: i) URL embedding layer ($f_r$), ii) URL sequence embedding layer ($f_m$), and iii) Logistic regression classifier ($f_c$). Only the unknown classifier parameters, $\param$ and b, (see Eq.~2 of the paper) of the dense layer and these of LSTM and ``dense'' mappings are trainable.}
    \label{fig:model_architecture}
\end{figure*}

\section{Dataset Statistics}
\label{sec:data_stats}

\begin{figure*}
    \centering
    \begin{subfigure}[t]{.345\textwidth}
    \includegraphics[width=1.\linewidth]{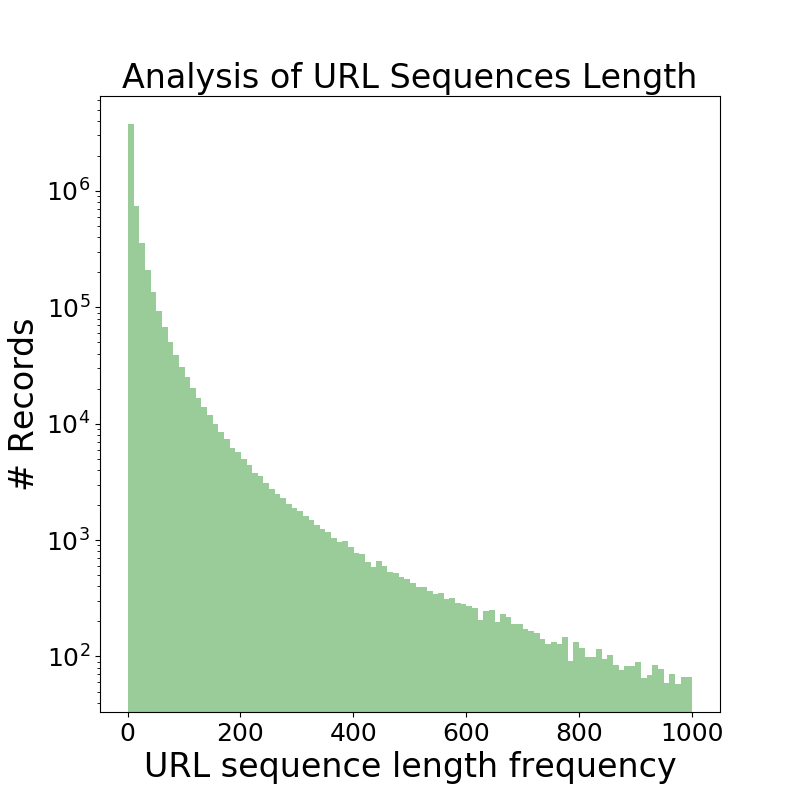} 
    \caption{Dataset $\CD_d$}
    \end{subfigure} \hspace{-1.5em}
    \begin{subfigure}[t]{.345\textwidth}
    \includegraphics[width=1.\linewidth]{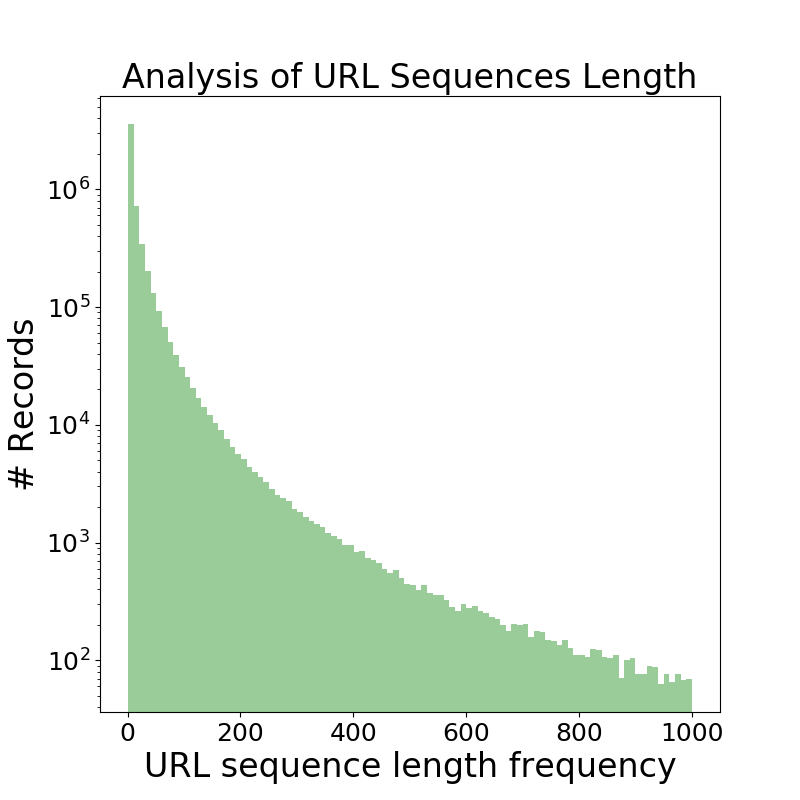}
    \caption{Dataset $\CD_{d+1}$}
    \end{subfigure} \hspace{-1.5em}
    \begin{subfigure}[t]{.345\textwidth}
    \includegraphics[width=1.\linewidth]{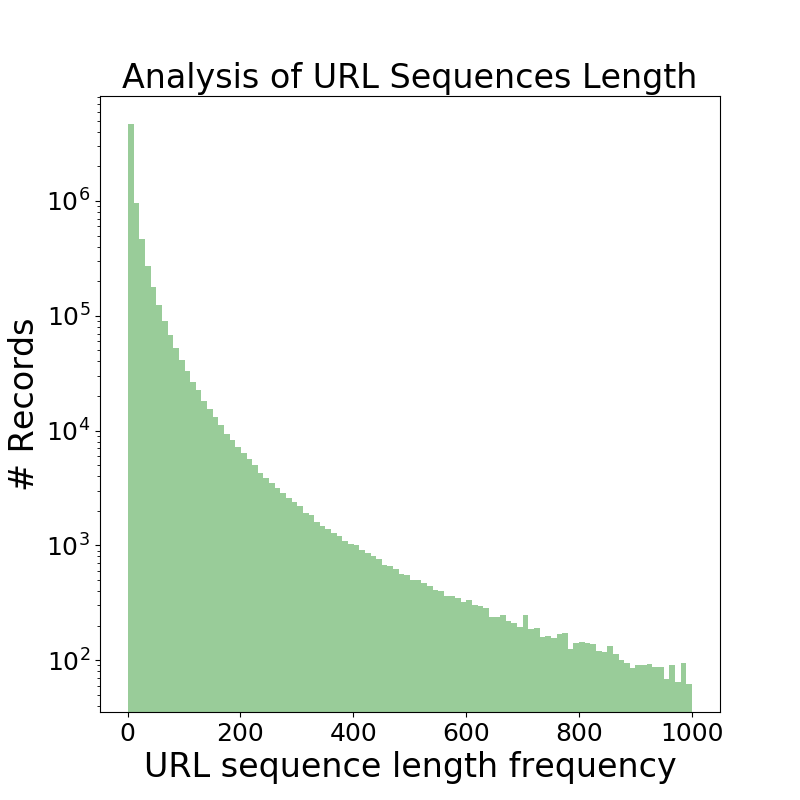}
    \caption{Dataset $\CD_{d+2}$}
    \end{subfigure}
    \caption{Analysis of the URL sequences lengths for the data, $\CD_d$, $\CD_{d+1}$, and $\CD_{d+2}$.}
    \label{fig:seq_analysis}
\end{figure*}

Figure~\ref{fig:seq_analysis} presents an analysis of the URL sequence lengths for $\CD_d$, $\CD_{d+1}$, and $\CD_{d+2}$. 
To be more precise, the distribution of the URL sequence lengths is given.
As it can be easily observed, the length of the URL sequences for most of the records is less than $500$.  Due to this fact, for the training and testing of our prediction models we are using the $500$ most recent visited URLs of each sequence. On the other hand, for the training of the representation the whole sequences are used. 

\begin{figure*}
    \centering
    \begin{subfigure}[t]{1.\textwidth}
    \includegraphics[width=.35\linewidth]{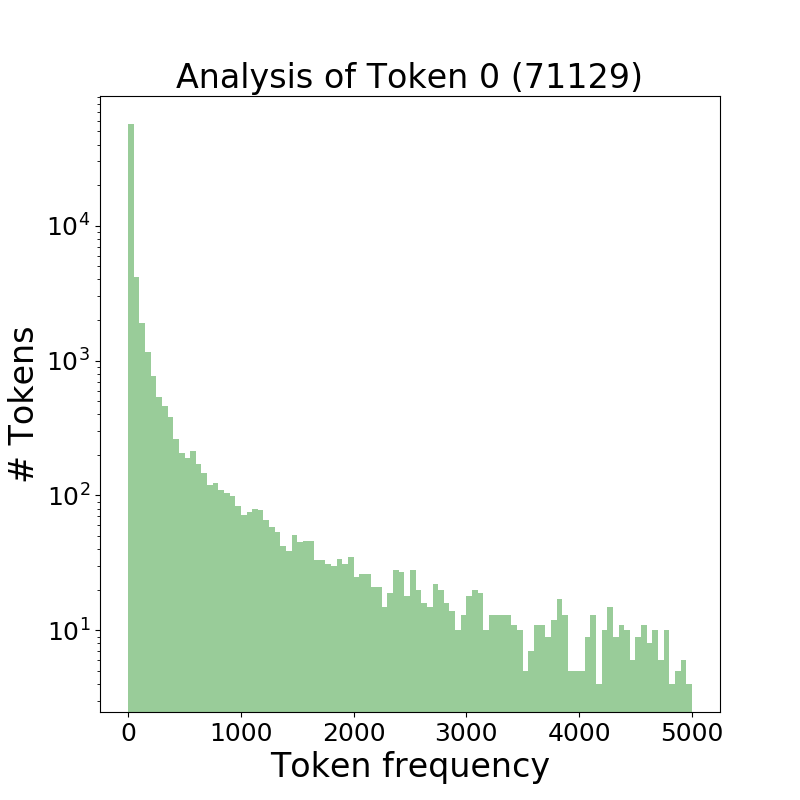} \hspace{-1.5em}
    \includegraphics[width=.35\linewidth]{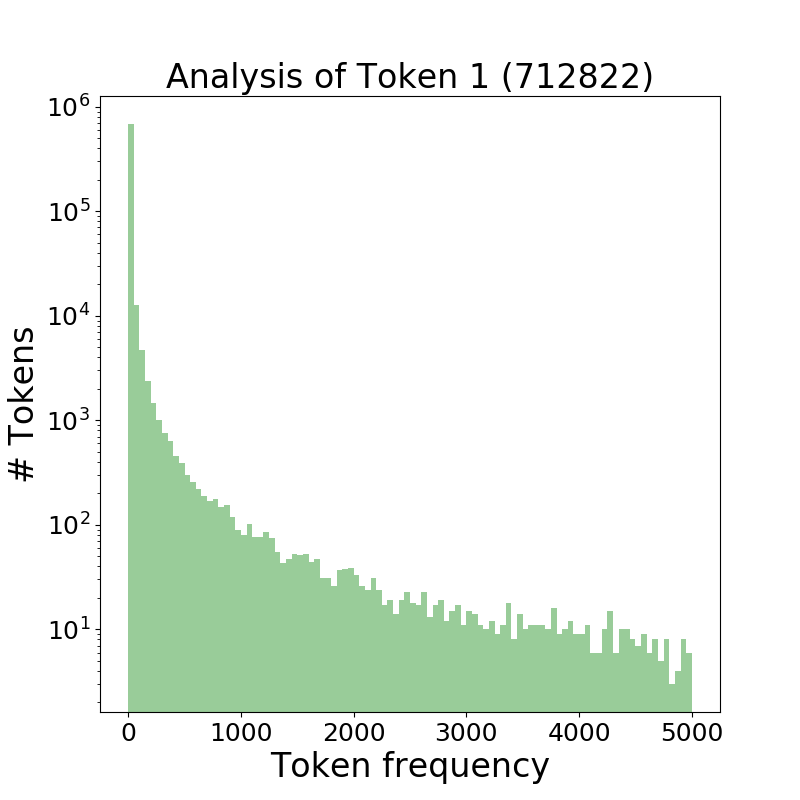} \hspace{-1.5em}
    \includegraphics[width=.35\linewidth]{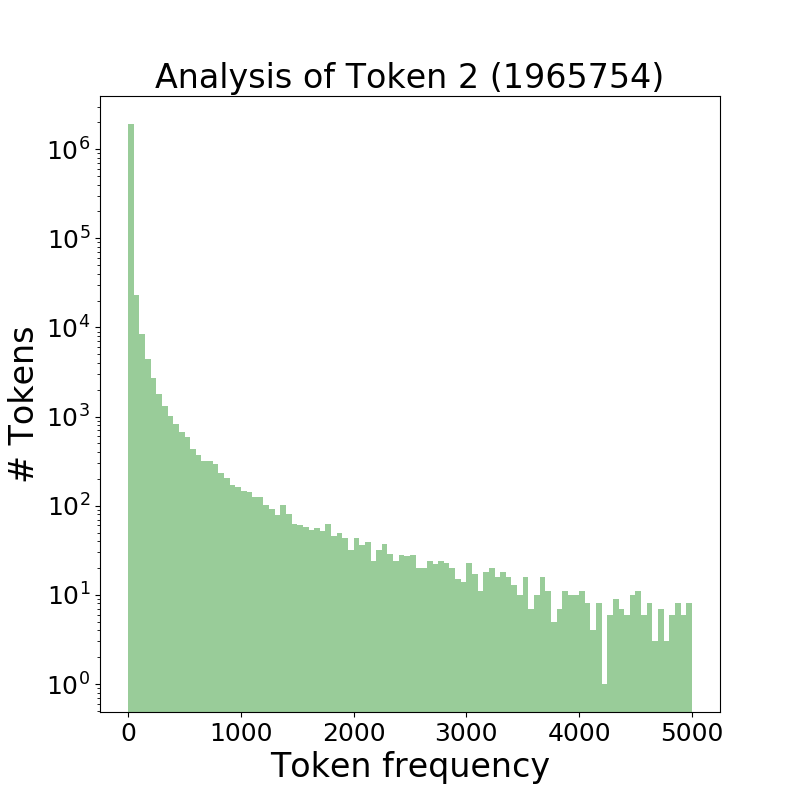}
    \caption{Dataset $\CD_d$}
    \end{subfigure}
    
    \begin{subfigure}[t]{1.\textwidth}
    \includegraphics[width=.35\linewidth]{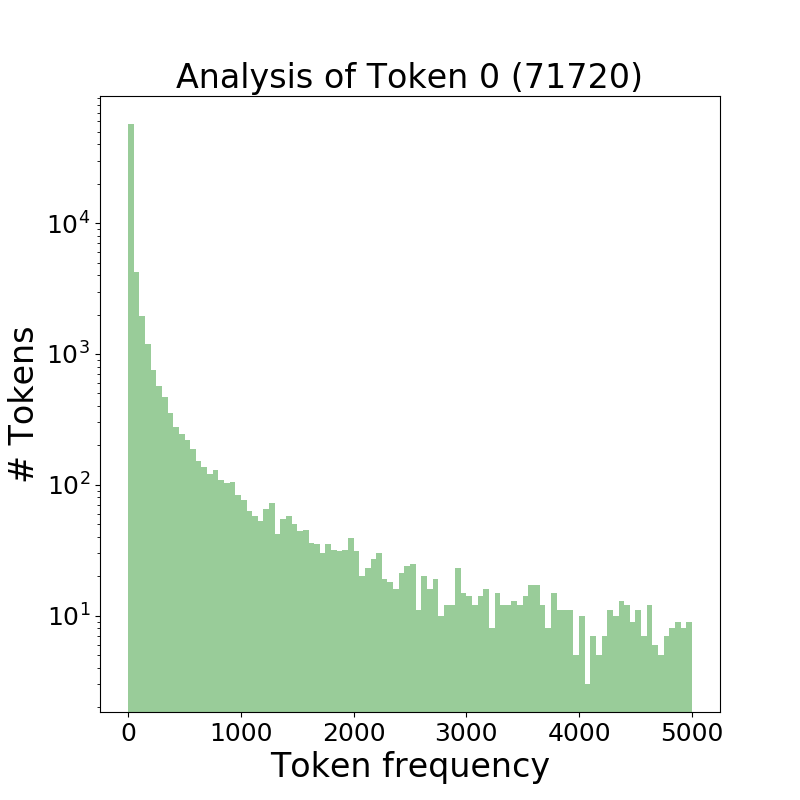} \hspace{-1.5em}
    \includegraphics[width=.35\linewidth]{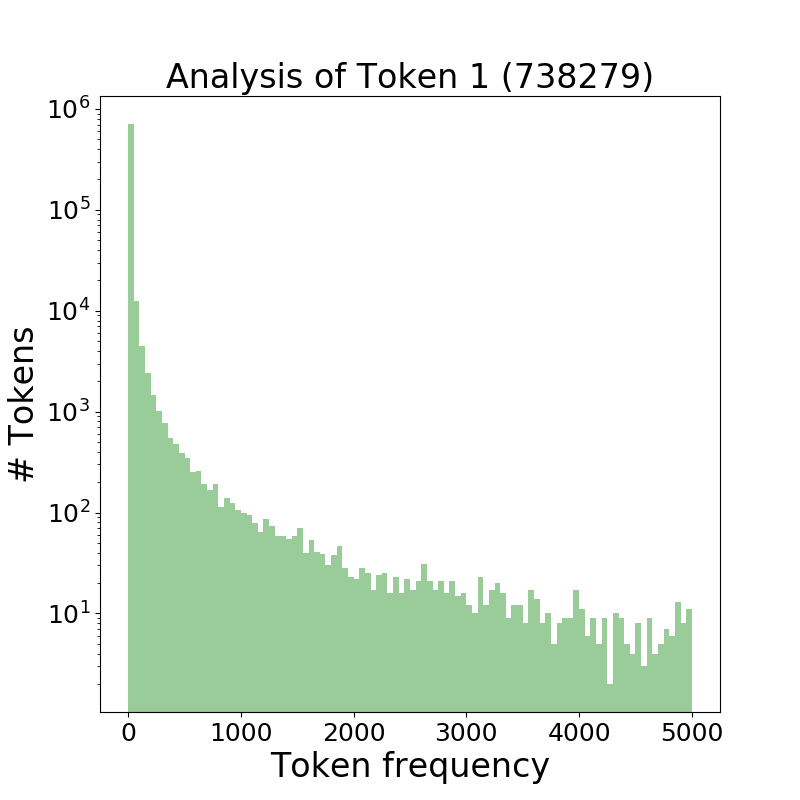} \hspace{-1.5em}
    \includegraphics[width=.35\linewidth]{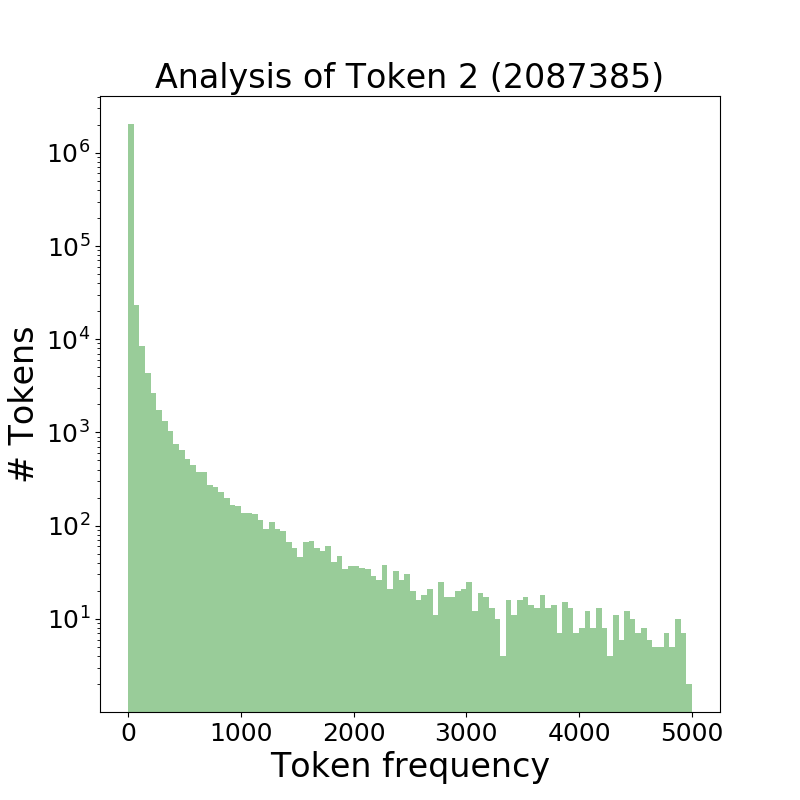}
    \caption{Dataset $\CD_{d+1}$}
    \end{subfigure}
    
    \begin{subfigure}[t]{1.\textwidth}
    \includegraphics[width=.35\linewidth]{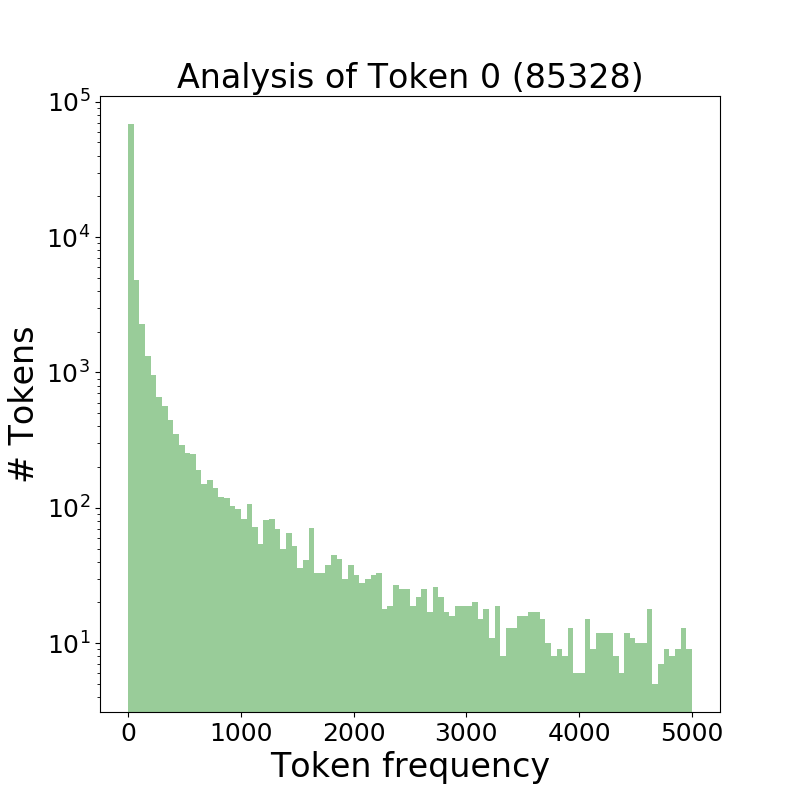} \hspace{-1.5em}
    \includegraphics[width=.35\linewidth]{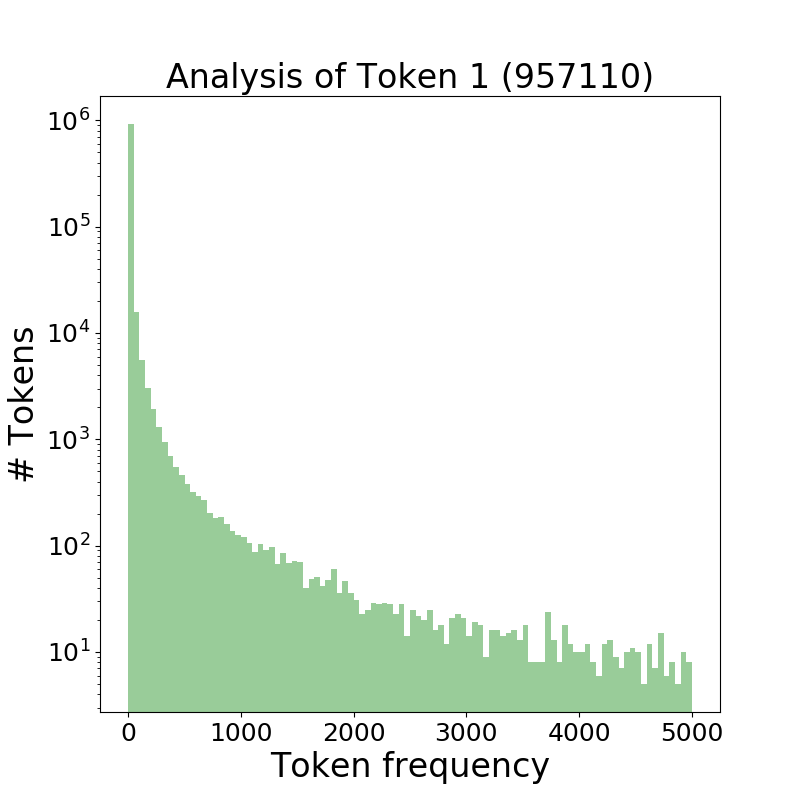} \hspace{-1.5em}
    \includegraphics[width=.35\linewidth]{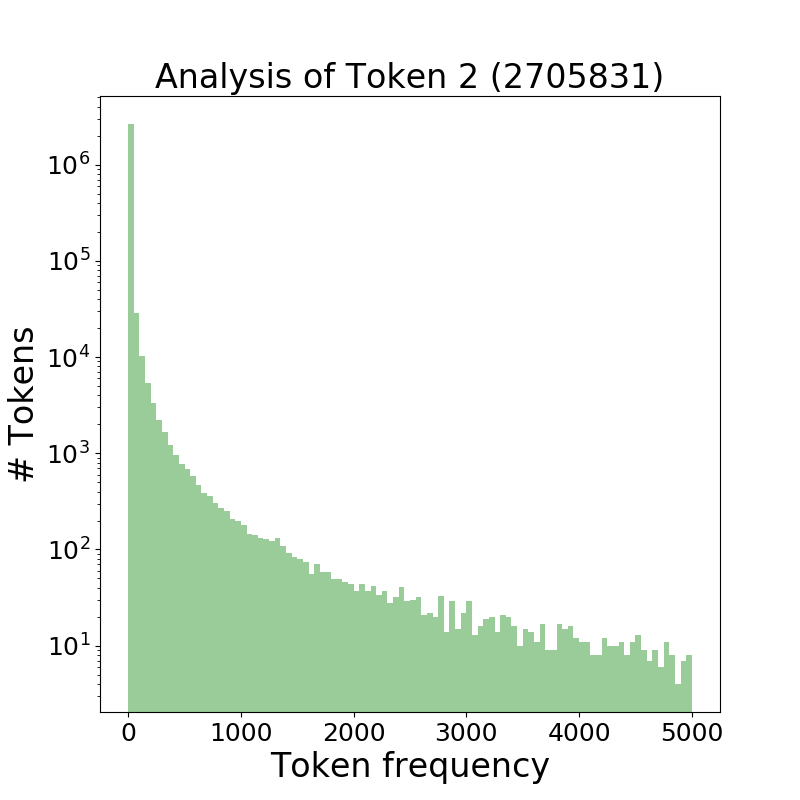}
    \caption{Dataset $\CD_{d+2}$}
    \end{subfigure}
    \caption{Analysis of URL tokens frequencies. X-axis represents the number of times a token is present in the dataset and y-axis shows the number of tokens. In parentheses we give the number of unique tokens.}
    \label{fig:Token_analysis}
\end{figure*}

In our setting, a URL is split with a `/' (slash) character and it  is supposed to consist of at most three tokens. 
Figure~\ref{fig:Token_analysis} shows the token frequency distribution of the first three URL tokens for each dataset.
As it was expected, the tokens at the last two spots are more \emph{rare} in the data. Also, the number of unique tokens in the first place (\emph{domain token}) is significantly smaller than the number of unique tokens on the other two spots. In our analysis, we consider a token ``rare'' if it appears less than $20$ times.

\section{Additional results}
\label{sec:sup_results}

Next, we report the results of our complete empirical analysis, where we examine the impact of the selection of the \emph{\{pos:neg\}} ratio (representation training) on the performance of the prediction model. 
More specifically, we present the performance of nineteen prediction models on the five advertisers described in the manuscript. 
In order to distinguish the nineteen different models, their names consist of three parts separated by a slash (`/') character apart from the \texttt{One\_hot/LR} model. The first part indicates the type of representation (\texttt{Domain\_only}, \texttt{Token\_avg}, \texttt{Token\_concat}), the second defines the kind of prediction model (\texttt{LR}, \texttt{DLR}, \texttt{RNN}), and the last one shows the \emph{\{pos:neg\}} ratio (\texttt{\{1{:}1\}}, \texttt{\{1{:}4\}}) used for the training of the representation model. 

\begin{figure*}[p]
    \centering
    \begin{subfigure}[t]{.33\textwidth}
    \includegraphics[width=1.\linewidth]{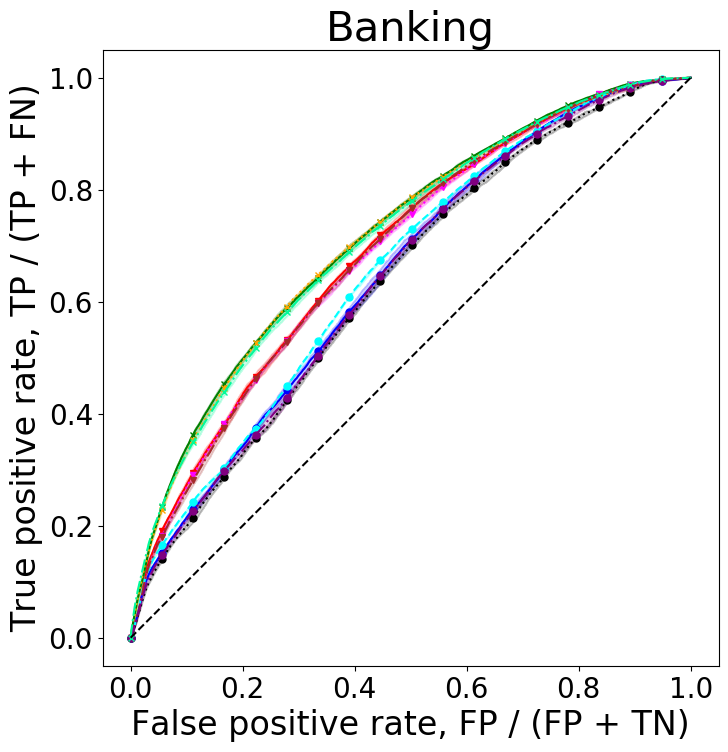}
    \end{subfigure} 
    \begin{subfigure}[t]{.33\textwidth}
    \includegraphics[width=1.\linewidth]{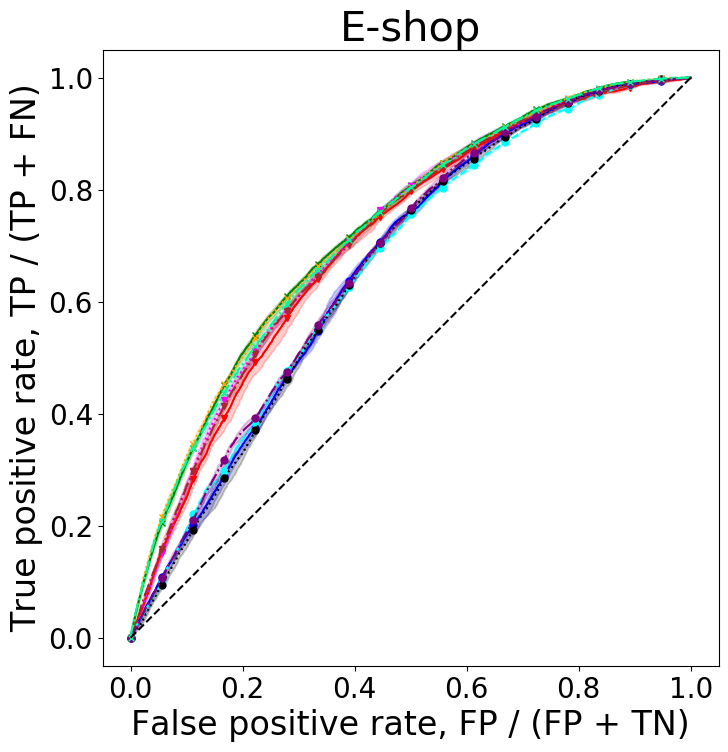}
    \end{subfigure} 
    \begin{subfigure}[t]{.33\textwidth}
    \includegraphics[width=1.\linewidth]{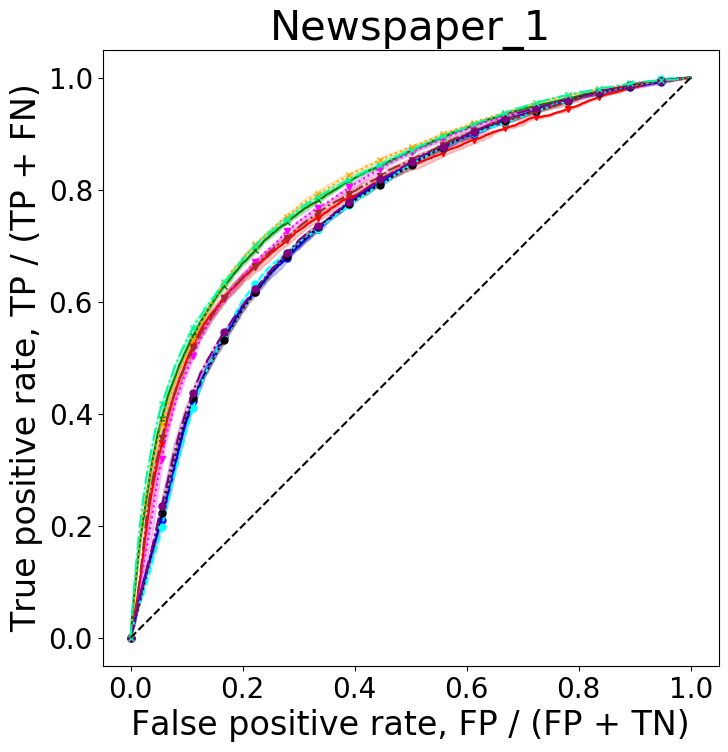}
    \end{subfigure} 
    
    \begin{subfigure}[t]{.33\textwidth}
    \includegraphics[width=1.\linewidth]{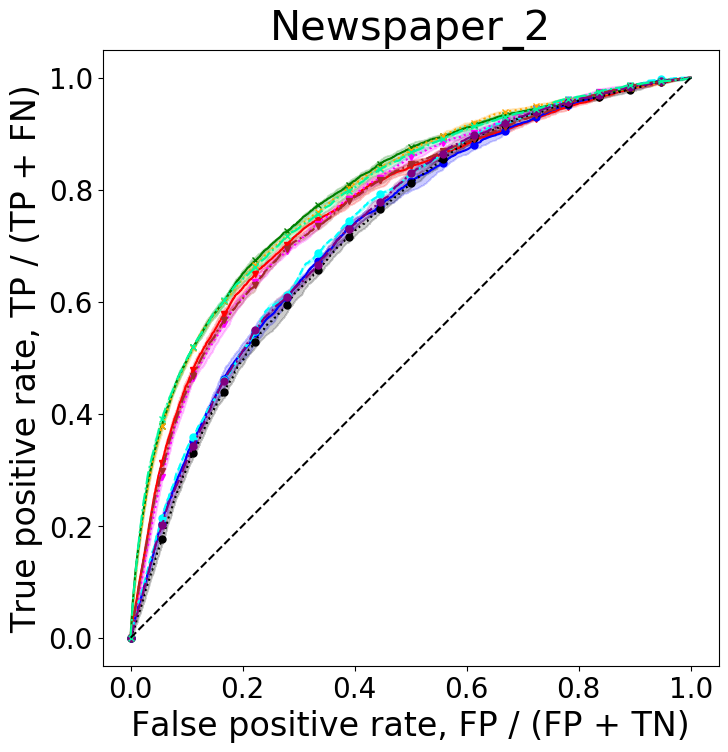}
    \end{subfigure} 
    \begin{subfigure}[t]{.33\textwidth}
    \includegraphics[width=1.\linewidth]{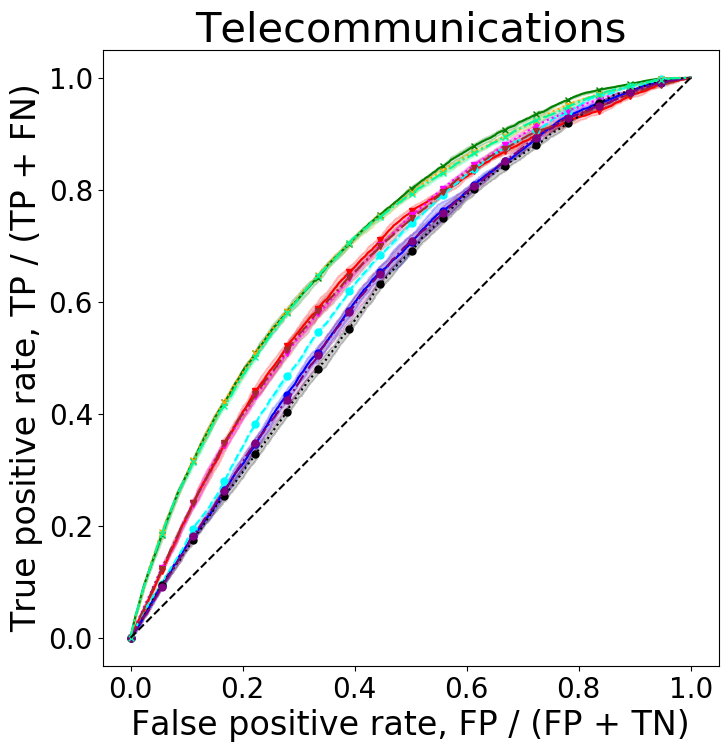}
    \end{subfigure} 
    \begin{subfigure}[t]{.33\textwidth}
    \includegraphics[width=1.\linewidth]{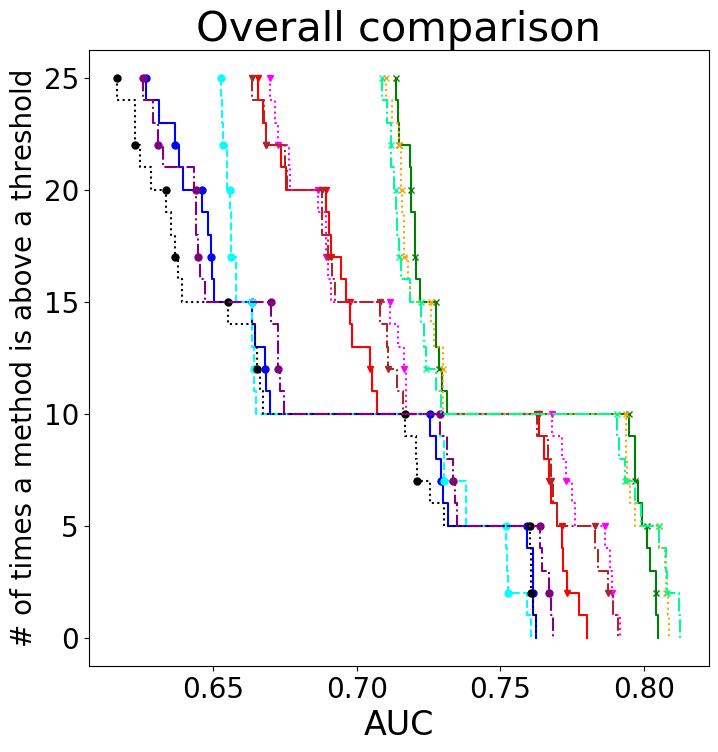}
    \end{subfigure}
    \begin{subfigure}[t]{1.\textwidth}
    \includegraphics[width=1.\linewidth]{./legends.png}
    \end{subfigure} 
    \caption{Average ROC curves of the ten conversion prediction ($\{1{:}1\}$ \emph{pos-neg} ratio) models on the five advertisers. Shaded regions represent the standard deviations over $5$ independent runs.  The bottom right plot presents the AUC for each one of the $25$ independent runs ($5$ advertisers $\times$ $5$ independent runs for each advertiser) of each model. The $\bullet$, $\blacktriangledown$ and $\times$ marks indicate the \texttt{LR}, \texttt{DLR} and \texttt{RNN} classification models, respectively.}
    \label{fig:empirical_results_1}
\end{figure*}

\begin{figure*}[p]
    \centering
    \begin{subfigure}[t]{.32\textwidth}
    \includegraphics[width=1.\linewidth]{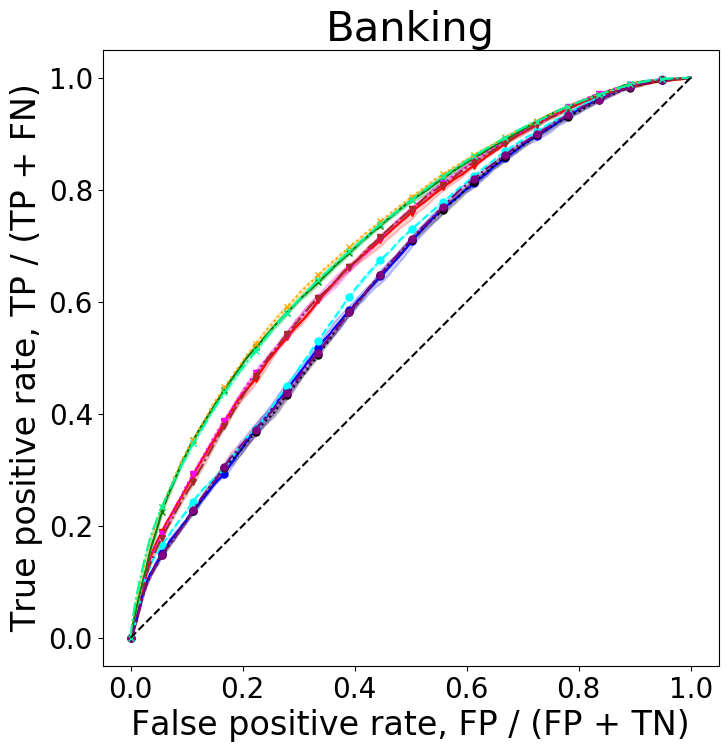}
    \end{subfigure} 
    \begin{subfigure}[t]{.32\textwidth}
    \includegraphics[width=1.\linewidth]{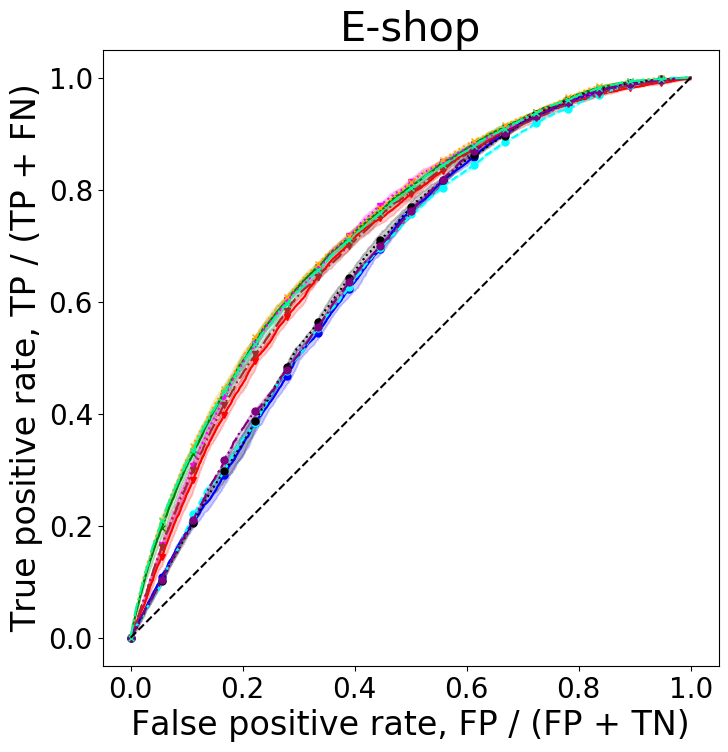}
    \end{subfigure} 
    \begin{subfigure}[t]{.32\textwidth}
    \includegraphics[width=1.\linewidth]{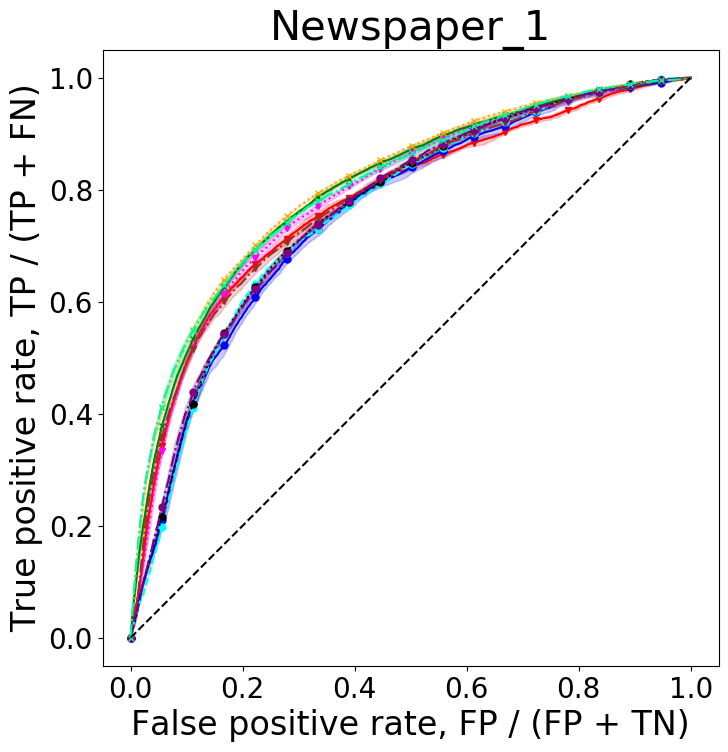}
    \end{subfigure} 
    
    \begin{subfigure}[t]{.32\textwidth}
    \includegraphics[width=1.\linewidth]{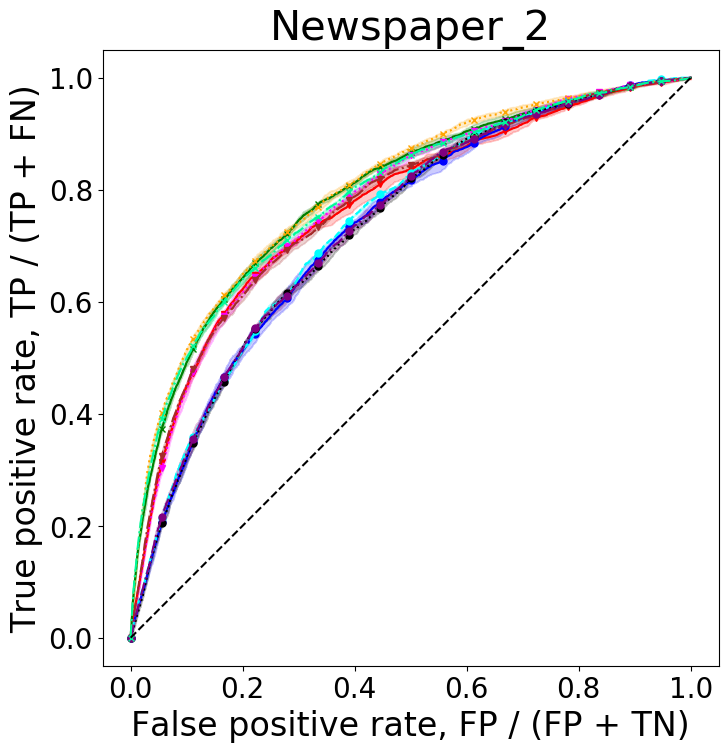}
    \end{subfigure} 
    \begin{subfigure}[t]{.32\textwidth}
    \includegraphics[width=1.\linewidth]{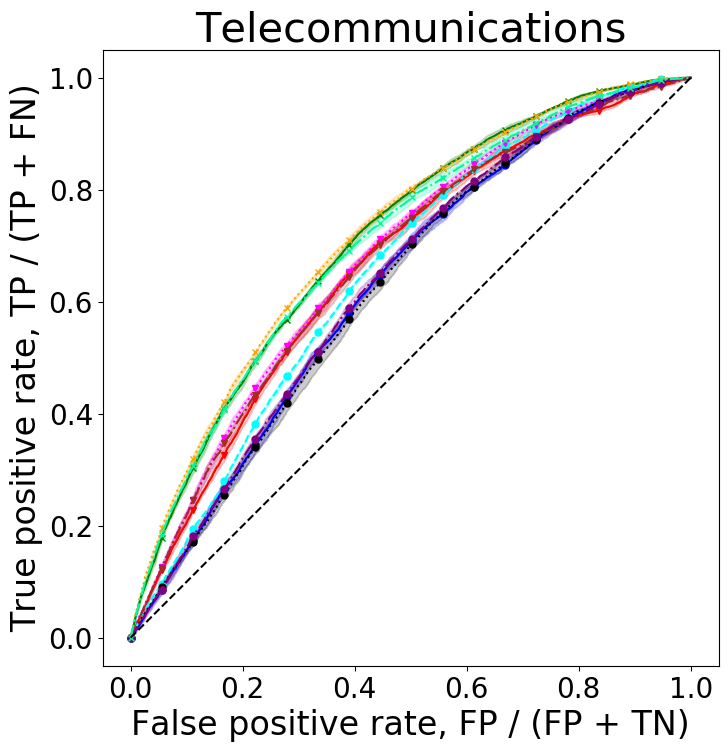}
    \end{subfigure} 
    \begin{subfigure}[t]{.32\textwidth}
    \includegraphics[width=1.\linewidth]{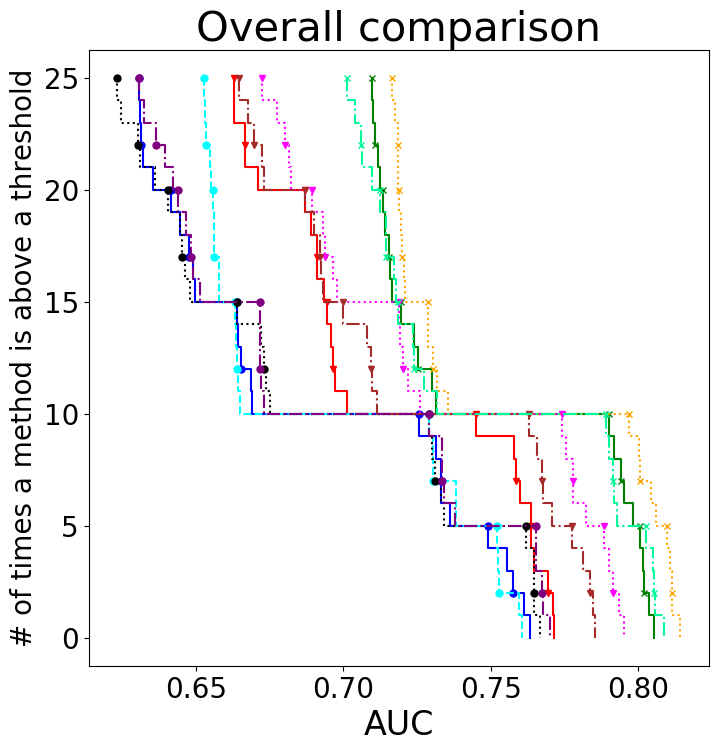}
    \end{subfigure}
    
    \begin{subfigure}[t]{1.\textwidth}
    \includegraphics[width=1.\linewidth]{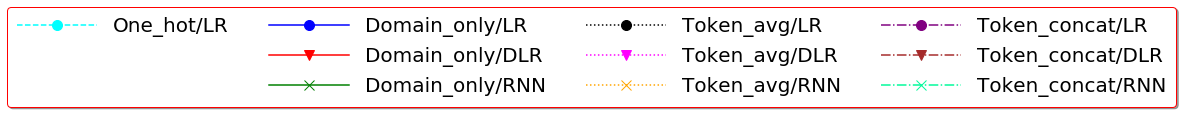}
    \end{subfigure} 
    \caption{Average ROC curves of the ten conversion prediction ($\{1{:}4\}$ \emph{pos-neg} ratio) models on the five advertisers. Shaded regions represent the standard deviations over $5$ independent runs.  The bottom right plot presents the AUC for each one of the $25$ independent runs ($5$ advertisers $\times$ $5$ independent runs for each advertiser) of each model. The $\bullet$, $\blacktriangledown$ and $\times$ marks indicate the \texttt{LR}, \texttt{DLR} and \texttt{RNN} classification models, respectively.}
    \label{fig:empirical_results_4}
\end{figure*}

As already highlighted at the paper, the area under ROC curve (AUC) is the metric that used to evaluate the prediction capabilities of each model. 
Table~\ref{tab:AUC} presents the average AUC of each model across five independent runs (five randomly selected seeds). Moreover, Figures~\ref{fig:empirical_results_1} and \ref{fig:empirical_results_4} illustrate the average ROC curves of the ten prediction models for each advertiser, where \texttt{\{1{:}1\}} and \texttt{\{1{:}4\}} \emph{\{pos:neg\}} ratios used for training representation models, respectively. Next, we present the main conclusions of our ``full'' empirical analysis:
\begin{itemize}
    \item All three proposed URL embedding models are able to learn high-quality vector representations that capture precisely the URL relationships. Even the \texttt{Domain\_only/LR}, \texttt{Token\_avg/LR}, and \texttt{Token\_concat/LR} models are able to predict with high accuracy the probability a user to be converted, and their performance is quite close or better compared to that of our baseline \texttt{One\_hot/LR}.
    
    \item It can be easily verified by the reported results that the prediction model performance is not so sensitive on the \emph{\{pos:neg\}} ratio used for training representation models. 
    
    \item \texttt{\{1{:}1\}} performs better on the \texttt{Domain\_only} representation compared to \texttt{\{1{:}4\}} that performs better on \texttt{Token\_avg}. On the other hand, the performance of both \emph{\{pos:neg\}} ratios are almost equivalent for the \texttt{Token\_concat} model.
    
    \item Among the three representation models, \texttt{Token\_avg} seems to be more adequate to capture the relationships between URLs, with the \texttt{Token\_concat} to be the second best. Moreover, the performance of \texttt{Domain\_only} representation is quite close to that of \texttt{Token\_concat}. 
    
    \item The consideration of the chronological order of the visited URLs (\texttt{RNN}) and the learning of dependencies among the embedding features (\texttt{DLR}) are also of high importance as both improve significantly the performance of the conversion prediction model. To be more precise, it is clear (see bottom right plot of Fig.~\ref{fig:empirical_results_1} and Fig.~\ref{fig:empirical_results_4}) that the \texttt{RNN} model surpass the performance of the rest two classifiers, while \texttt{DLR} performs better compared to \texttt{LR}.
    
    \item The best  prediction conversion models are \texttt{Domain\_only/RNN} and \texttt{Token\_avg/RNN} for \texttt{\{1{:}1\}} and \texttt{\{1{:}4\}} \emph{\{pos{:}neg\}} ratios, respectively (see bottom right plot of Figs.~\ref{fig:empirical_results_1} and Fig.~\ref{fig:empirical_results_4}).
\end{itemize}

\begin{figure*}[p]
    \centering
    \begin{subfigure}[t]{1.\textwidth}
    \includegraphics[width=1\linewidth]{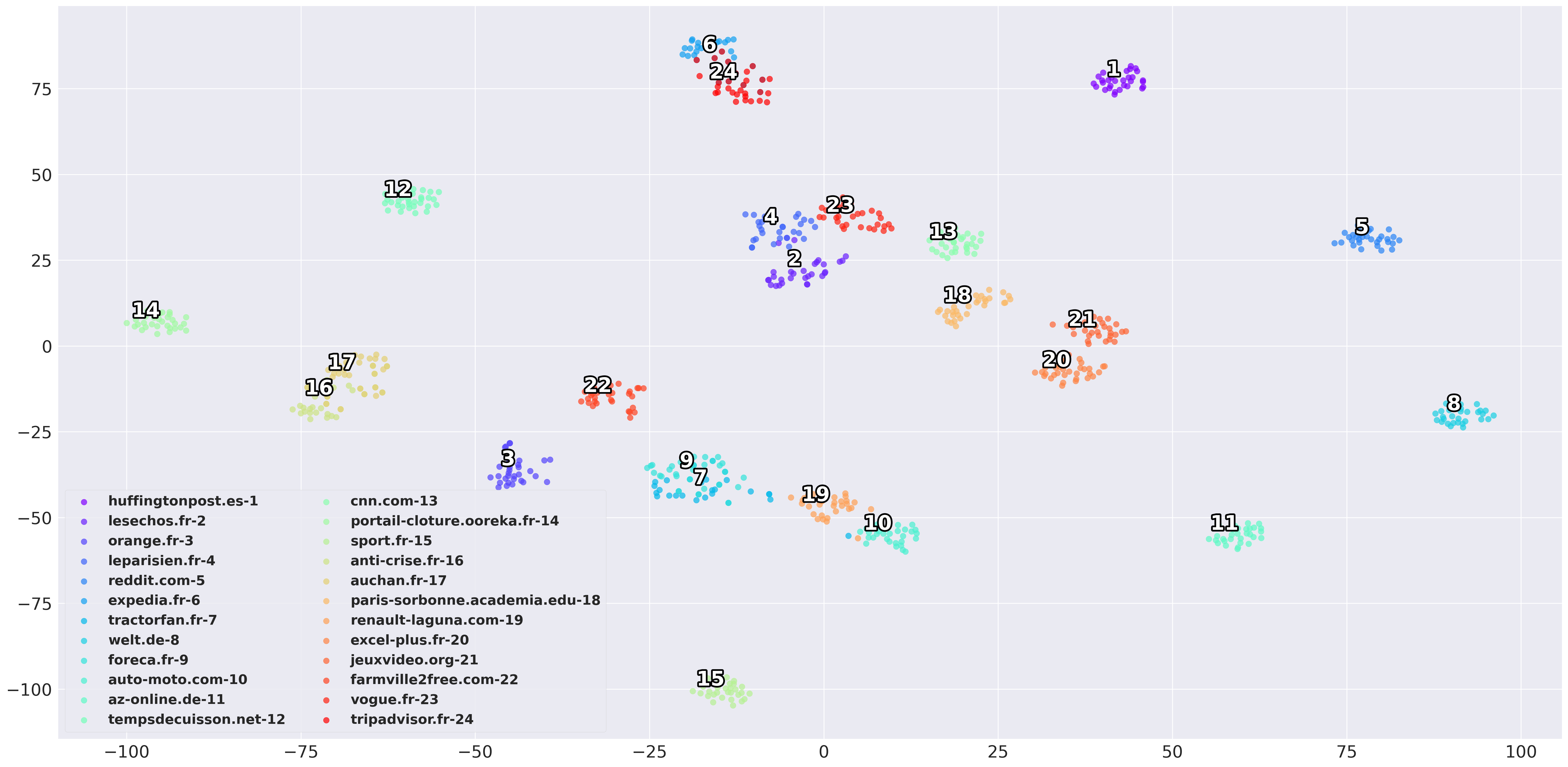}
    \caption{\texttt{Domain\_only/1:1} URL representation model}
    \label{fig:url_representation/1:1}
    \end{subfigure}
    \begin{subfigure}[t]{1.\textwidth}
    \includegraphics[width=1\linewidth]{./similar_words_seaborn.png}
    \caption{\texttt{Domain\_only/1:4} URL representation model}
    \label{fig:url_representation/1:4}
    \end{subfigure}
    \caption{t-SNE visualization of the thirty closest neighbors of $24$ different domains. The colors of the points indicate the closest domain of each URL.}
    \label{fig:url_representation}
\end{figure*}

\begin{figure*}[p]
    \centering
    \includegraphics[width=.193\linewidth]{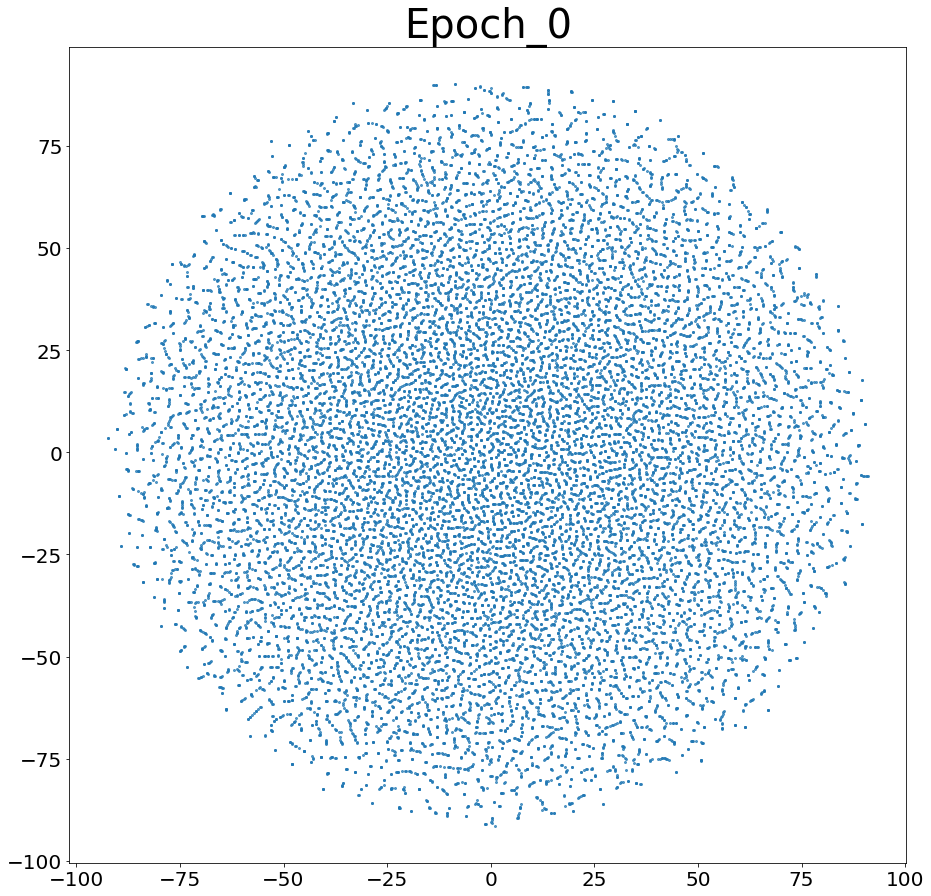}
    \includegraphics[width=.193\linewidth]{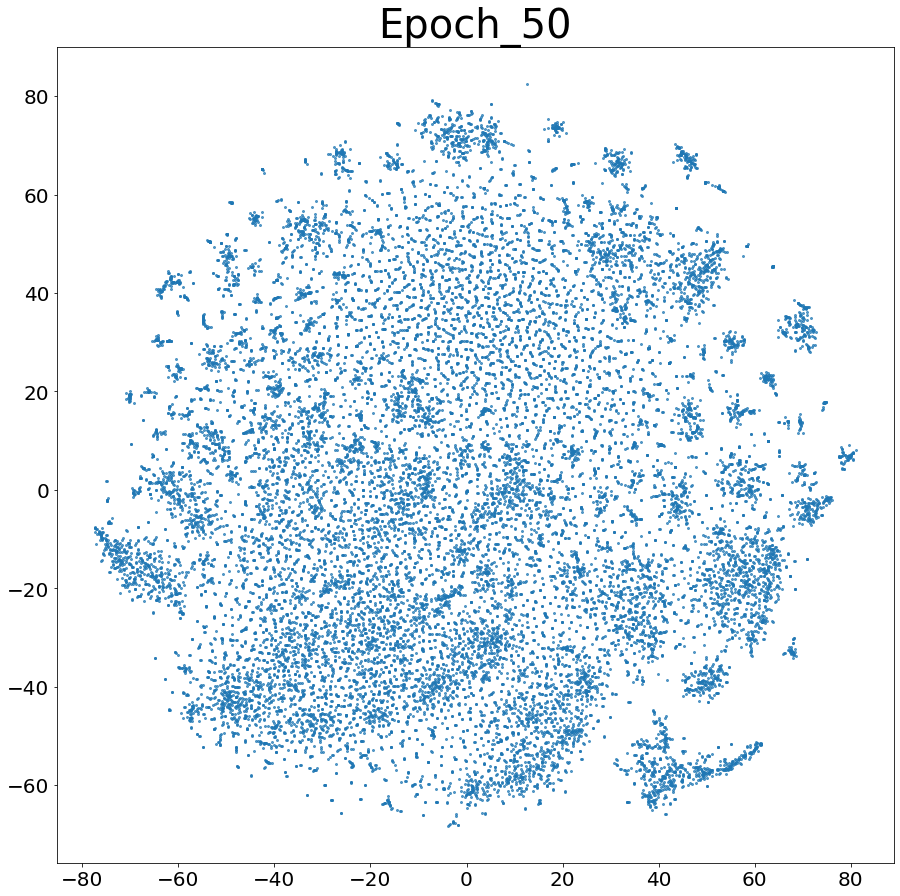}
    \includegraphics[width=.193\linewidth]{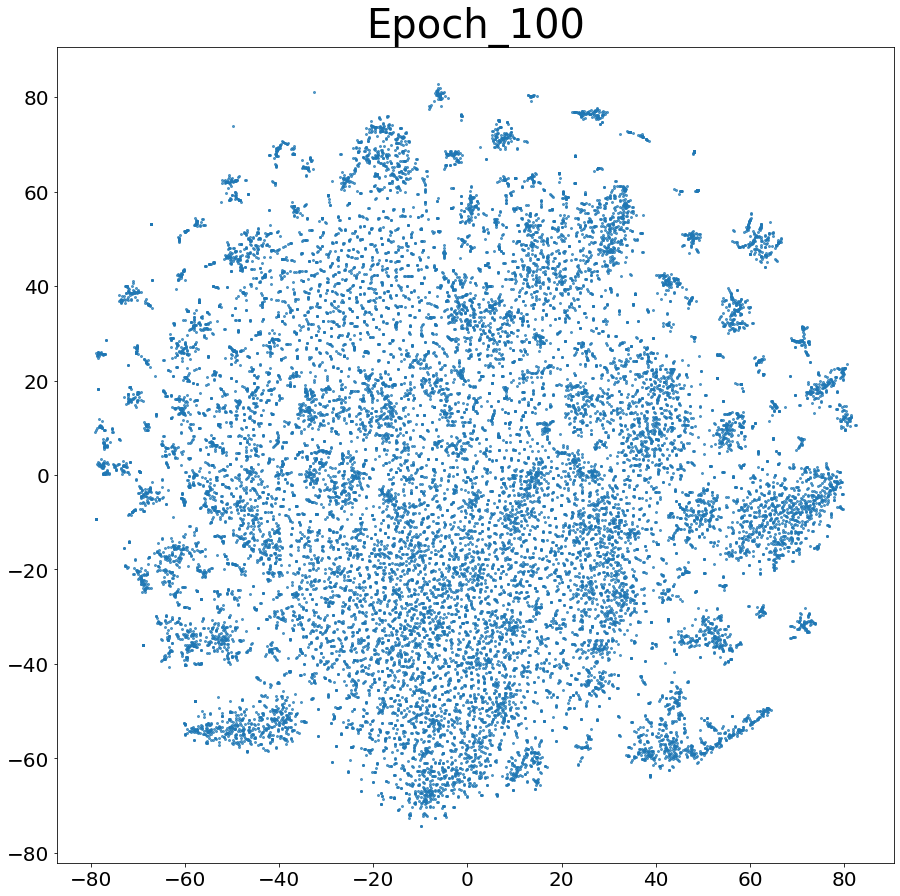}
    \includegraphics[width=.193\linewidth]{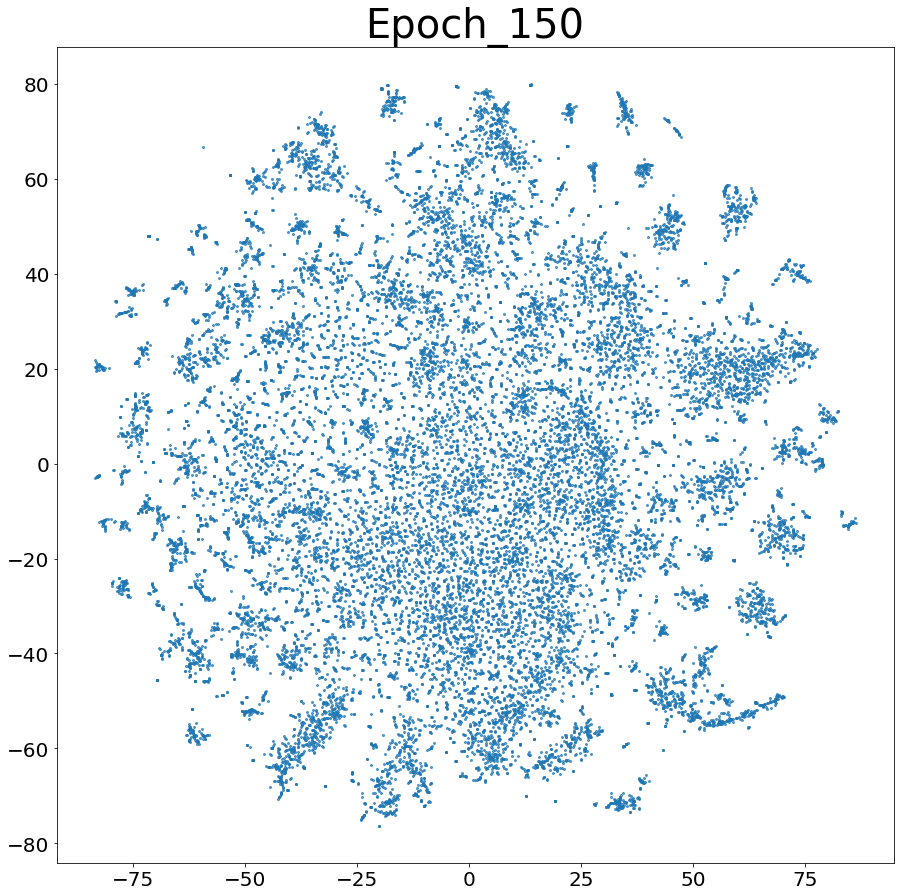}
    \includegraphics[width=.193\linewidth]{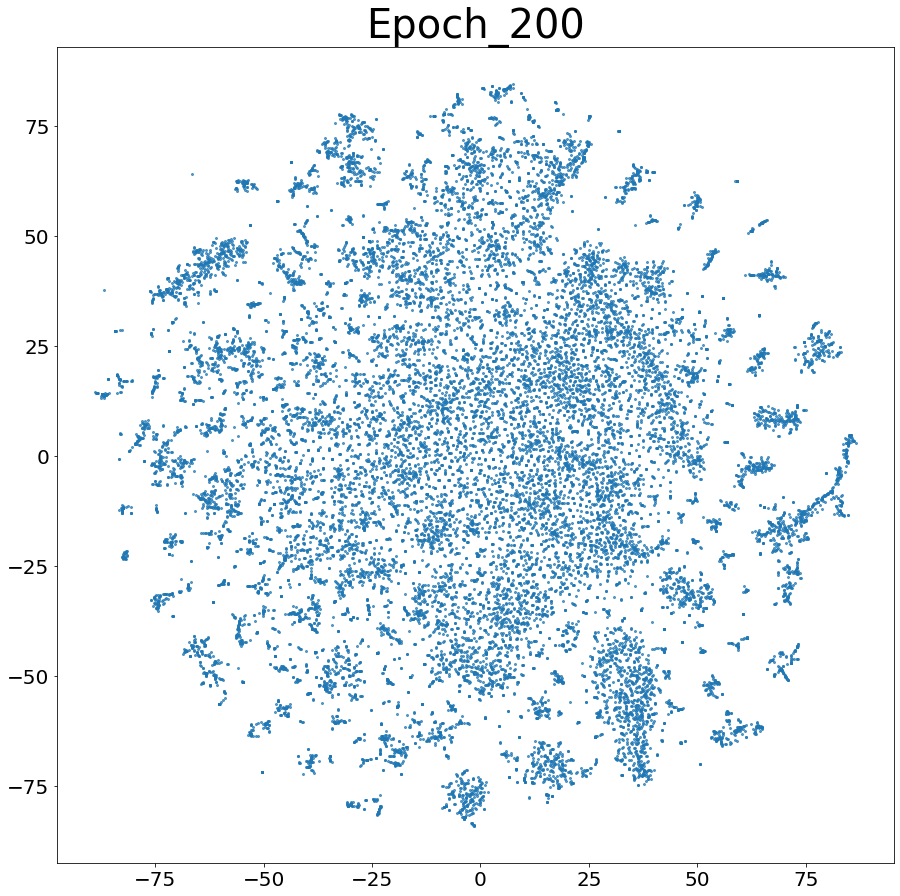}
    \caption{t-SNE visualization of the embedding matrix trained by \texttt{Domain\_only/1{:}4} representation model after $0$, $50$, $100$, $150$, and $200$ epochs, respectively.}
    \label{fig:embeddings_vis}
\end{figure*}

\begin{table*}[t]
    \caption{Average ($\%$) and standard deviation of the area under ROC curves ($5$ independent runs) of the $10$ prediction models on $5$ different campaigns.}
    \label{tab:AUC}
    \Large
    \resizebox{1.\textwidth}{!}{
    \centering
    \begin{tabular}{c|P{2.5cm}|P{2.5cm}|P{2.5cm}|P{2.5cm}|P{2.5cm}}
        \hline
        \hline
        \diagbox{\bf Method}{\bf Advertiser} & {\bf Banking} & {\bf E-shop}	& {\bf Newspaper\_1} &	{\bf Newspaper\_2} & {\bf Telecom} \\
        \hline
        \texttt{One\_hot/LR} &  $65.7 \pm 0.093$ &  $66.4 \pm 0.053$ &  $75.5 \pm 0.379$ &  $73.3 \pm 0.400$ &   $65.4 \pm 0.085$ \\
        \hline
        \texttt{Domain\_only/LR/1:1} &  $64.9 \pm 0.138$ &  $66.7 \pm 0.237$ &  $76.1 \pm 0.109$ &  $72.9 \pm 0.219$ &   $63.4 \pm 0.479$ \\
        \hline
        \texttt{Domain\_only/LR/1:4} &  $64.6 \pm 0.300$ &  $66.6 \pm 0.217$ &  $75.7 \pm 0.507$ &  $73.2 \pm 0.345$ &   $63.2 \pm 0.168$ \\
        \hline
        \texttt{Domain\_only/DLR/1:1} &  $69.2 \pm 0.268$ &  $70.3 \pm 0.383$ &  $77.4 \pm 0.397$ &  $76.7 \pm 0.333$ &   $67.0 \pm 0.370$ \\
        \hline
        \texttt{Domain\_only/DLR/1:4} &  $69.0 \pm 0.214$ &  $69.7 \pm 0.234$ &  $76.8 \pm 0.342$ &  $75.7 \pm 0.658$ &   $66.6 \pm 0.303$ \\
        \hline
        \texttt{Domain\_only/RNN/1:1} &  ${\bf71.9} \pm 0.258$ &  $72.9 \pm 0.149$ &  $80.3 \pm 0.149$ &  $79.7 \pm 0.146$ &   $71.7 \pm 0.252$ \\
        \hline
        \texttt{Domain\_only/RNN/1:4} &  $71.4 \pm 0.144$ &  $72.6 \pm 0.422$ &  $80.3 \pm 0.168$ &  $79.4 \pm 0.281$ &   $71.2 \pm 0.250$ \\
        \hline
        \texttt{Token\_avg/LR/1:1} &  $63.6 \pm 0.188$ &  $66.4 \pm 0.438$ &  $76.1 \pm 0.076$ &  $72.3 \pm 0.466$ &   $62.3 \pm 0.380$ \\
        \hline
        \texttt{Token\_avg/LR/1:4} &  $64.5 \pm 0.241$ &  $67.2 \pm 0.390$ &  $76.4 \pm 0.152$ &  $73.1 \pm 0.184$ &   $62.9 \pm 0.468$ \\
        \hline
        \texttt{Token\_avg/DLR/1:1} &  $68.9 \pm 0.148$ &  $71.5 \pm 0.219$ &  $78.9 \pm 0.175$ &  $77.3 \pm 0.279$ &   $67.3 \pm 0.263$ \\
        \hline
        \texttt{Token\_avg/DLR/1:4} &  $69.4 \pm 0.294$ &  $72.1 \pm 0.263$ &  $79.2 \pm 0.242$ &  $77.7 \pm 0.274$ &   $67.9 \pm 0.348$ \\
        \hline
        \texttt{Token\_avg/RNN/1:1} &  $71.7 \pm 0.116$ &  $72.9 \pm 0.184$ &  $80.8 \pm 0.121$ &  $79.5 \pm 0.119$ &   $71.4 \pm 0.209$ \\
        \hline
        \texttt{Token\_avg/RNN/1:4} &  ${\bf71.9} \pm 0.082$ &  ${\bf73.1} \pm 0.246$ &  $\bf{81.2} \pm 0.153$ &  $\bf{80.2} \pm 0.322$ &   $\bf{71.8} \pm 0.153$ \\
        \hline
        \texttt{Token\_concat/LR/1:1} &  $64.5 \pm 0.142$ &  $67.3 \pm 0.149$ &  $76.6 \pm 0.174$ &  $73.3 \pm 0.216$ &   $63.3 \pm 0.672$ \\
        \hline
        \texttt{Token\_concat/LR/1:4} &  $64.8 \pm 0.241$ &  $67.2 \pm 0.060$ &  $76.7 \pm 0.179$ &  $73.4 \pm 0.273$ &   $63.6 \pm 0.425$ \\
        \hline
        \texttt{Token\_concat/DLR/1:1} &  $69.0 \pm 0.187$ &  $71.2 \pm 0.274$ &  $78.7 \pm 0.304$ &  $76.7 \pm 0.232$ &   $67.0 \pm 0.469$ \\
        \hline
        \texttt{Token\_concat/DLR/1:4} &  $69.1 \pm 0.222$ &  $70.8 \pm 0.400$ &  $78.2 \pm 0.285$ &  $76.7 \pm 0.255$ &   $66.9 \pm 0.310$ \\
        \hline
        \texttt{Token\_concat/RNN/1:1} &  $71.5 \pm 0.214$ &  $72.5 \pm 0.264$ &  $80.9 \pm 0.286$ &  $79.4 \pm 0.318$ &   $71.2 \pm 0.180$ \\
        \hline
        \texttt{Token\_concat/RNN/1:4} &  $71.5 \pm 0.224$ &  $72.5 \pm 0.460$ &  $80.5 \pm 0.192$ &  $79.1 \pm 0.130$ &   $70.5 \pm 0.278$ \\
        \hline
    \end{tabular}}
\end{table*}

\begin{table*}[p]
\begin{center}
\caption{The $30$-nearest neighbors of $24$ different domains according to our trained \texttt{Domain\_only/1:1} representation model.}
\LARGE
\label{tab:nearest_neighbors_1}
\resizebox{1\textwidth}{!}{
\input{./nearest_neighbors_1-1.tex}
}
\end{center}
\end{table*}
\begin{table*}[p]
\begin{center}
\LARGE
\caption{The $30$-nearest neighbors of $24$ different domains according to our trained \texttt{Domain\_only/1:4} representation model.}
\label{tab:nearest_neighbors_4}
\resizebox{1.\textwidth}{!}{
\input{./nearest_neighbors.tex}
}
\end{center}
\end{table*}

\subsection{Representation Visualisation}

In this section we visualise the embedding vectors of $24$ selected domains along with those of the thirty closest URLs of each one of them, which have been learned by \texttt{Domain/1:1} (Fig.~\ref{fig:url_representation/1:1}) and \texttt{Domain/1:4} (Fig.~\ref{fig:url_representation/1:4}) representation models. Looking carefully at Fig.~\ref{fig:url_representation}, we observe that clusters close on the embedding space produced by \texttt{Domain/1:1} are also close on the embedding space of \texttt{Domain/1:4}. For instance, clusters (6) [\href{https://expedia.fr}{expedia.fr}] and (24) [\href{https://tripadvisor.fr}{tripadvisor.fr}] are close in both cases. That is normal, as these two clusters contain URLs about \emph{travelling}. The same also holds for the  URLs embeddings of clusters ($2$) [\href{https://lesechos.fr}{lesechos.fr}], ($4$) [\href{https://leparisien.fr}{leparisien.fr}], and ($23$) [\href{https://vogue.fr}{vogue.fr}] that belong to the \emph{news} category. Moreover, clusters  ($10$) [\href{https://auto-moto.com}{auto-moto.com}] and ($19$) [\href{https://renault-laguna.com}{renault-laguna.com}] are also close as both are related to \emph{automobile}. Another example is that of clusters $16$ [\href{https://anti-crise.fr}{anti-crise.fr}] and $17$ [\href{https://auchan.fr}{auchan.fr}] that are about \emph{promotions}. To be sure that the URLs belong to the same category (and are not just random URLs) with that of their closest domain, Tables~\ref{tab:nearest_neighbors_1} and \ref{tab:nearest_neighbors_4} provide the $30$-nearest URLs for each one of the $24$ domains for the \texttt{Domain/1:1} and \texttt{Domain/1:4}, respectively. Finally, Fig.~\ref{fig:embeddings_vis} illustrates the t-SNE visualization of the embedding matrix ($22101 \times 100$) trained by \texttt{Domain\_only/1:4} representation model after $0$, $50$, $100$, $150$, and $200$ epochs, respectively.

%% file: representation_learning.tikz
%auto-ignore
\begin{tikzpicture}
\node[rec] at (0,0) {};
\node[textnode] at (-0.88, 0.35) {$1$};
\node[rec] at (0.3,-0.3) {};
\node[textnode] at (-0.58, 0.05) {$2$};
\node[textnode] at (-0.4, -0.1) {$.$};
\node[textnode] at (-0.3, -0.2) {$.$};
\node[textnode] at (-0.2, -0.3) {$.$};
\node[rec, text width=4em, text centered] at (0.9,-0.9) {URL \\ Sequences};
\node[textnode] at (0.1, -0.6) {N};
\node[rec, text width=6em, text centered] at (5.,-0.45) {representation \\ model, $f_r$};
\draw[arrow] (2.3,-0.45) -> (3.3,-0.45);
\end{tikzpicture}

%% file: model_learning.tikz
%auto-ignore
\begin{tikzpicture}
\node[rec] at (0,0) {};
\node[textnode] at (-0.88, 0.35) {$1$};
\node[rec] at (0.3,-0.3) {};
\node[textnode] at (-0.58, 0.05) {$2$};
\node[textnode] at (-0.4, -0.1) {$.$};
\node[textnode] at (-0.3, -0.2) {$.$};
\node[textnode] at (-0.2, -0.3) {$.$};
\node[rec, text width=4em, text centered] at (0.9,-0.9) {URL \\    Sequences};
\node[textnode] at (0.1, -0.6) {N};

\node[textnode] at (1.68, 0.35) {$label_1$};
\node[textnode] at (1.88, 0.05) {$label_2$};
\node[textnode] at (1.88, -0.15) {$.$};
\node[textnode] at (1.98, -0.25) {$.$};
\node[textnode] at (2.08, -0.35) {$.$};
\node[textnode] at (2.5, -0.6) {$label_N$};

\draw[arrow] (3.8,-0.65) -> (4.8,-0.65);
\node[rec, text width=6em, text centered] at (4.3,0.0) {representation \\ model, $f_r$};

\node[rec] at (7,0) {};
\node[textnode] at (7-0.88, 0.35) {$1$};
\node[rec] at (7+0.3,-0.3) {};
\node[textnode] at (7-0.58, 0.05) {$2$};
\node[textnode] at (7-0.4, -0.1) {$.$};
\node[textnode] at (7-0.3, -0.2) {$.$};
\node[textnode] at (7-0.2, -0.3) {$.$};
\node[rec, text width=4.7em, text centered] at (7+0.9,-0.9) {URL \\ Embeddings};
\node[textnode] at (7+0.1, -0.6) {N};

\node[textnode] at (7+1.68, 0.35) {$label_1$};
\node[textnode] at (7+1.88, 0.05) {$label_2$};
\node[textnode] at (7+1.88, -0.15) {$.$};
\node[textnode] at (7+1.98, -0.25) {$.$};
\node[textnode] at (7+2.08, -0.35) {$.$};
\node[textnode] at (7+2.5, -0.6) {$label_N$};

\draw[arrow] (10.,-0.65) -> (11.,-0.65);
\node[textnode] at (10.5,-0.35) {$f_m$};

\node[rec, text width=6em, text centered] at (12.3,-0.65) {Classifier, $f_c$};

\end{tikzpicture}

%% file: skipgram.tikz
%auto-ignore
\begin{tikzpicture}[align=center]

%%% Left Column
\node[rectangle, minimum width=5cm, minimum height=1cm, draw=blue!50, fill=blue!10, text width=5cm, text centered] (target_url) at (0,0) {{\bf Target URL} \\ $url_t = [token_{1}^{(t)}, token_{2}^{(t)}, token_{3}^{(t)}]$};

\node[rectangle, minimum width=5cm, minimum height=2.0cm, draw=red!50, text width=5cm, text centered, below=1cm of target_url, fill=red!10] (target_embedding_layer) {{\bf Token Embedding Layer} \\ ({dic\_size} $\times$ {embedding\_dim})};

\draw[arrow] (target_url) --  (target_embedding_layer);

\node[rectangle, dashed, minimum width=5cm, minimum height=3.cm, draw=black, below=1cm of target_embedding_layer, fill=orange!5] (target_token_embeddings) {};
\node[textnode, below=1.cm of target_embedding_layer] (text_tar) {\bf Target Token Embedding};
\node[textnode, minimum width=4.8cm, minimum height=.5cm, draw=orange!80, below=.1cm of text_tar, fill=orange!5] (et1) {Target $token_1$ embedding $\ve_1^{(t)}$};
\node[textnode, minimum width=4.8cm, minimum height=.5cm, draw=orange!80, below=.1cm of et1, fill=orange!5] (et2) {Target $token_2$ embedding $\ve_2^{(t)}$};
\node[textnode, minimum width=4.8cm, minimum height=.5cm, draw=orange!80, below=.1cm of et2, fill=orange!5] (et3) {Target $token_3$ embedding $\ve_3^{(t)}$};

\draw[-] (target_embedding_layer) --  (target_token_embeddings);

\node[rectangle, draw=orange!80, minimum width=5cm, minimum height=1cm, below=1cm of target_token_embeddings, fill=orange!5] (target_url_embedding) {{\bf Target $URL$ embedding layer} \\ \texttt{Domain\_only}, \texttt{Token\_avg}, \texttt{Token\_concat}};

\draw[arrow] (target_token_embeddings) --  (target_url_embedding);

%%% Right Column
\node[rectangle, minimum width=5cm, minimum height=1cm, draw=blue!50, text width=5cm, fill=blue!10, text centered] (context_url) [right=2cm of target_url] {{\bf Context URL} \\ $url_c = [token_{1}^{(c)}, token_{2}^{(c)}, token_{3}^{(c)}]$};

\node[rectangle, minimum width=5cm, minimum height=2.0cm, draw=red!50, text width=5cm, text centered, fill=red!10] (context_embedding_layer) [below=1cm of context_url] {{\bf Token Embedding Layer} \\ ({dic\_size} $\times$ {embedding\_dim})};

\draw[arrow] (context_url) --  (context_embedding_layer);

\node[rectangle, dashed, minimum width=5cm, minimum height=3.cm, draw=orange!80, below=1cm of context_embedding_layer, fill=orange!5] (context_token_embeddings) {};
\node[textnode, below=1.cm of context_embedding_layer] (text_car) {\bf Context Token Embedding};
\node[textnode, minimum width=4.8cm, minimum height=.5cm, draw=orange!80, below=.1cm of text_car, fill=orange!5] (ct1) {Context $token_1$ embedding $\ve_1^{(c)}$};
\node[textnode, minimum width=4.8cm, minimum height=.5cm, draw=orange!80, below=.1cm of ct1, fill=orange!5] (ct2) {Context $token_2$ embedding $\ve_2^{(c)}$};
\node[textnode, minimum width=4.8cm, minimum height=.5cm, draw=orange!80, below=.1cm of ct2, fill=orange!5] (ct3) {Context $token_3$ embedding $\ve_3^{(c)}$};

\draw[-] (context_embedding_layer) --  (context_token_embeddings);

\node[rectangle, draw=orange!80, minimum width=5cm, minimum height=1cm, below=1cm of context_token_embeddings, fill=orange!5] (context_url_embedding) {{\bf Context $URL$ embedding layer} \\ \texttt{Domain\_only}, \texttt{Token\_avg}, \texttt{Token\_concat}};

\draw[arrow] (context_token_embeddings) --  (context_url_embedding);

%%%%%%
\node[rectangle, draw=green!50, below right=1.cm and -1.75 of target_url_embedding, minimum width=5cm, minimum height=1cm, fill=green!10] (merge_layer) {{\bf Similarity Layer} \\ Dot Product};

\draw[arrow] (target_url_embedding.south)  |-  (merge_layer.west) node[midway,sloped,left] {$\vx_t$};
\draw[arrow] (context_url_embedding.south)  |-  (merge_layer.east) node[midway,sloped,right] {$\vx_c$};

\node[rectangle, draw=red!50, below=1cm of merge_layer, minimum width=5cm, minimum height=1cm, fill=red!10] (dense_layer) {{\bf Dense Layer} \\ (Sigmoid)};

\draw[arrow] (merge_layer) --  (dense_layer) node[midway,right] {$\vx_t^\top \vx_c$};

\node[textnode, below=.5cm of dense_layer] (output) {};
\draw[arrow] (dense_layer) --  (output) node[midway,right] {$y \in \{0,1\}$};

\end{tikzpicture}

%% file: nearest_neighbors_1-1.tex
%auto-ignore
\begin{tabular}{C{4cm}|p{30cm}}
\hline
\hline
{\bf Domain} & \begin{minipage}{1.6\columnwidth} \center \bf $30$-nearest neighbors (URLS)\end{minipage} \\
\hline
{ \noindent \bf \color{blue} huffingtonpost.es} & sevilla.abc.es; elconfidencial.com; smoda.elpais.com; verne.elpais.com; elmon.cat; elcorreo.com; expansion.com; infolibre.es; vozpopuli.com; blogs.elconfidencial.com; eldiario.es; okdiario.com; cope.es; m.eldiario.es; elespanol.com; elperiodico.com; levante-emv.com; diariodesevilla.es; eleconomista.es; elcaso.elnacional.cat; periodistadigital.com; farodevigo.es; guiadelocio.com; libertaddigital.com; 20minutos.es; motor.elpais.com; mitele.es; diariodenavarra.es; lasexta.com; diariovasco.com \\
\hline
{ \noindent \bf \color{blue} lesechos.fr} &  business.lesechos.fr; latribune.fr; afrique.latribune.fr; financedemarche.fr; bfmbusiness.bfmtv.com; mieuxvivre-votreargent.fr; photo.capital.fr; capital.fr; challenges.fr; journaldunet.com; lopinion.fr; mtf-b2b-asq.fr; marianne.net; hbrfrance.fr; contrepoints.org; journaldeleconomie.fr; boursier.com; zonebourse.com; investopedia.com; manager-go.com; e-marketing.fr; actufinance.fr; argent.boursier.com; btf-b2b-asq.fr; linkedin.com; mtf-finance-asq.fr; nasdaq.com; btf-finance-asq.fr; lecoindesentrepreneurs.fr; 07-ros-btf-laplacemedia-3.fr \\
\hline
{ \noindent \bf \color{blue} orange.fr} & actu.orange.fr; login.orange.fr; finance.orange.fr; tendances.orange.fr; lemoteur.orange.fr; meteo.orange.fr; messagerie.orange.fr; programme-tv.orange.fr; sports.orange.fr; news.orange.fr; boutique.orange.fr; auto.orange.fr; zapping-tv.orange.fr; webmail.orange.fr; people.orange.fr; occasion.auto.orange.fr; mescontacts.orange.fr; video-streaming.orange.fr; pro.orange.fr; mail02.orange.fr; agenda.orange.fr; tv.orange.fr; cineday.orange.fr; mail01.orange.fr; belote-coinchee.net; 118712.fr; ww.orange.fr; cliquojeux.com; freecell.fr; mundijeux.fr \\
\hline
{ \noindent \bf \color{blue} leparisien.fr} & btf-actu-asq.fr; lefigaro.fr; marianne.net; scoopnest.com; legorafi.fr; cnews.fr; actu17.fr; bfmtv.com; tendanceouest.com; atlasinfo.fr; opex360.com; jeuneafrique.com; lejdd.fr; video.lefigaro.fr; lalibre.be; rtl.be; liberation.fr; fdesouche.com; causeur.fr; 24matins.fr; amp.lefigaro.fr; courrierinternational.com; bladi.net; tunisienumerique.com; lci.fr; courrier-picard.fr; 7sur7.be; air-defense.net; mixcloud.com; pss-archi.eu \\
\hline
{ \noindent \bf \color{blue} reddit.com} & pcgaming.reddit.com; smashbros.reddit.com; old.reddit.com; funny.reddit.com; gaming.reddit.com; europe.reddit.com; france.reddit.com; soccer.reddit.com; askreddit.reddit.com; anime.reddit.com; memes.reddit.com; globaloffensive.reddit.com; starcitizen.reddit.com; freefolk.reddit.com; nintendoswitch.reddit.com; popular.reddit.com; games.reddit.com; aww.reddit.com; leagueoflegends.reddit.com; gameofthrones.reddit.com; politics.reddit.com; redditad.com; dota2.reddit.com; all.reddit.com; totalwar.reddit.com; knowyourmeme.com; pcmasterrace.reddit.com; movies.reddit.com; tinder.reddit.com; nba.reddit.com \\
\hline
{ \noindent \bf \color{blue} expedia.fr} & skyscanner.fr; kayak.fr; momondo.fr; fr.hotels.com; fr.lastminute.com; vol.lastminute.com; liligo.fr; quandpartir.com; esky.fr; govoyage.fr; opodo.fr; lonelyplanet.fr; virail.fr; carnetdescapades.com; voyage.lastminute.com; lastminute.com; bravofly.fr; edreams.fr; easyvols.fr; tripadvisor.fr; locations.lastminute.com; ebookers.fr; voyages.bravofly.fr; opodo.com; reservation.lastminute.com; trainhotel.lastminute.com; flights-results.liligo.fr; voyageforum.com; voyagespirates.fr; govoyages.com \\
\hline
{ \noindent \bf \color{blue} tractorfan.fr} & forum.farm-connexion.com; discountfarmer.com; heure-ouverture.com; technikboerse.com; meteo-sud-aveyron.over-blog.com; ovniclub.com; materiel-agricole.annuairefrancais.fr; enviedechasser.fr; mash70-75.com; unimog-mania.com; pecheapied.net; renault5.forumactif.com; srxteam.forums-actifs.net; palombe.com; spa-du-dauphine.fr; foreca.fr; meteopassion.com; fiatagri.superforum.fr; meteosurfcanarias.com; super-tenere.org; actu-automobile.com; opel-mokka.forumpro.fr; motoconseils.com; fr.agrister.com; esoxiste.com; v-strom.superforum.fr; view.robothumb.com; sudoku-evolution.com; refugebeauregard.forumactif.com; m.meteorama.fr \\
\hline
{ \noindent \bf \color{blue} welt.de} & zeit.de; tagesspiegel.de; sueddeutsche.de; faz.net; sport1.de; badische-zeitung.de; merkur.de; rp-online.de; kicker.de; handelsblatt.com; tz.de; tvmovie.de; abendblatt.de; sportbild.bild.de; nzz.ch; bild.de; seattletimes.com; express.de; morgenpost.de; bz-berlin.de; netzwelt.de; bunte.de; derwesten.de; t-online.de; general-anzeiger-bonn.de; mopo.de; transfermarkt.de; techbook.de; finanzen.net; aargauerzeitung.ch \\
\hline
{ \noindent \bf \color{blue} foreca.fr} & meteo81.fr; meteosurfcanarias.com; sudoku-evolution.com; meteopassion.com; mxcircuit.fr; ledicodutour.com; pecheursunisdelille.com; moncompte-espaceclient.com; impactfm.fr; discountfarmer.com; meteo-normandie.fr; monde-du-velo.com; fr.tutiempo.net; meteojura.fr; ovniclub.com; palombe.com; sur.ly; videos-chasse-peche.com; retroplane.net; pointeduraz.info; carriere.annuairefrancais.fr; meteo-sud-aveyron.over-blog.com; horaires-douverture.fr; banquesenfrance.fr; genealogic.review; fr.meteovista.be; baboun57.over-blog.com; meteonews.ch; pecheapied.net; fournaise.info \\
\hline
{ \noindent \bf \color{blue} auto-moto.com} & feline.cc; largus.fr; news.autojournal.fr; auto-mag.info; test-auto.auto-moto.com; automobile-magazine.fr; caradisiac.com; neowebcar.com; essais.autojournal.fr; promoneuve.fr; autojournal.fr; latribuneauto.com; motorlegend.com; turbo.fr; actu-moteurs.com; worldscoop.forumpro.fr; palais-de-la-voiture.com; voiture.autojournal.fr; autoplus.fr; moniteurautomobile.be; recherche.autoplus.fr; news.sportauto.fr; abcmoteur.fr; fiches-auto.fr; ww2.autoscout24.fr; constructeur.autojournal.fr; zeperfs.com; essais-autos.com; motoservices.com; sportauto.fr \\
\hline
{ \noindent \bf \color{blue} az-online.de} & alittlecraftinyourday.com; theorganisedhousewife.com.au; brittanyherself.com; directoalpaladar.com.mx; mikseri.net; homeplans.com; thesurvivalgardener.com; dmv.org; milliondollarjourney.com; symbols.com; xataka.com.co; kiwilimon.com; mu-43.com; houseofjoyfulnoise.com; leinetal24.de; turniptheoven.com; wetterkanal.kachelmannwetter.com; thinksaveretire.com; lovechicliving.co.uk; wereparents.com; cookrepublic.com; foodinjars.com; lettermenrow.com; raegunramblings.com; thedailytay.com; losreplicantes.com; intmath.com; arthritis-health.com; ourpaleolife.com; juneauempire.com \\
\hline
{ \noindent \bf \color{blue} tempsdecuisson.net} & cuisinenligne.com; mamina.fr; scally.typepad.com; audreycuisine.fr; atelierdeschefs.fr; temps-de-cuisson.info; aux-fourneaux.fr; uneplumedanslacuisine.com; cuisine-facile.com; gateaux-et-delices.com; chefnini.com; humcasentbon.over-blog.com; yummix.fr; fruitsdelamer.com; toutlemondeatabl.canalblog.com; lesjoyauxdesherazade.com; gateaux-chocolat.fr; petitsplatsentreamis.com; pechedegourmand.canalblog.com; certiferme.com; yumelise.fr; quelquesgrammesdegourmandise.com; toques2cuisine.com; ricardocuisine.com; companionetmoi.com; recettessimples.fr; docteurbonnebouffe.com; cuisinealafrancaise.com; lighttome.fr; recetteramadan.com \\
\hline
{ \noindent \bf \color{blue} cnn.com} & us.cnn.com; grimsbytelegraph.co.uk; stadiumtalk.com; uk.reuters.com; euronews.com; expressandstar.com; thedailymash.co.uk; politico.eu; nbcnews.com; thedailybeast.com; trendscatchers.co.uk; lancashiretelegraph.co.uk; blogs.spectator.co.uk; itpro.co.uk; standard.co.uk; kentonline.co.uk; doityourself.com; deliaonline.com; puzzles.independent.co.uk; breakingnews.ie; oxfordmail.co.uk; slashdot.org; sportinglife.com; abcactionnews.com; huffpost.com; indy100.com; anagrammer.com; drivepedia.com; nigella.com; trumpexcel.com \\
\hline
{ \noindent \bf \color{blue} portail-cloture.ooreka.fr} & mur.ooreka.fr; bricolage-facile.net; forum-maconnerie.com; decoration.ooreka.fr; porte.ooreka.fr; abri-de-jardin.ooreka.fr; fr.rec.bricolage.narkive.com; expert-peinture.fr; muramur.ca; amenagementdujardin.net; tondeuse.ooreka.fr; desinsectisation.ooreka.fr; aac-mo.com; fenetre.ooreka.fr; bricoleurpro.com; recuperation-eau-pluie.ooreka.fr; pergola.ooreka.fr; volet.ooreka.fr; papier-peint.ooreka.fr; arrosoirs-secateurs.com; carrelage.ooreka.fr; assainissement.ooreka.fr; pierreetsol.com; poele-cheminee.ooreka.fr; forumbrico.fr; poimobile.fr; parquet.ooreka.fr; forum-plomberie.com; deconome.com; serrure.ooreka.fr \\
\hline
{ \noindent \bf \color{blue} sport.fr} & infomercato.fr; topmercato.com; mercatofootanglais.com; parisfans.fr; footlegende.fr; pariskop.fr; le10sport.com; allpaname.fr; le-onze-parisien.fr; vipsg.fr; planetepsg.com; mercatoparis.fr; paristeam.fr; buzzsport.fr; canal-supporters.com; football.fr; culturepsg.com; footparisien.com; foot-sur7.fr; footradio.com; mercatolive.fr; olympique-et-lyonnais.com; planetelille.com; footballclubdemarseille.fr; blaugranas.fr; 90min.com; sportune.fr; footmarseillais.com; livefoot.fr; foot01.com \\
\hline
{ \noindent \bf \color{blue} anti-crise.fr} & echantillonsclub.com; cfid.fr; gesti-odr.com; plusdebonsplans.com; promoalert.com; forum.anti-crise.fr; madstef.com; franceechantillonsgratuits.com; cataloguemate.fr; mesechantillonsgratuits.fr; argentdubeurre.com; auchan.fr; maximum-echantillons.com; forum.madstef.com; tous-testeurs.com; jeu-concours.biz; vos-promos.fr; tiendeo.fr; echantillonsgratuits.fr; echantinet.com; pubeco.fr; bons-plans-astuces.com; lp.testonsensemble.com; toutdonner.com; conforama.fr; clubpromos.fr; vente-unique.com; ofertolino.fr; promo-conso.net; fr.testclub.com \\
\hline
{ \noindent \bf \color{blue} auchan.fr} & conforama.fr; but.fr; rueducommerce.fr; vente-unique.com; cdiscount.com; touslesprix.com; webmarchand.com; anti-crise.fr; fr.shopping.com; cataloguemate.fr; leguide.com; offrespascher.com; promoalert.com; fr.xmassaver.net; promobutler.be; plusdebonsplans.com; tiendeo.fr; promopascher.com; destockplus.com; meilleurvendeur.com; idealo.fr; pubeco.fr; cdiscountpro.com; argentdubeurre.com; clubpromos.fr; mistergooddeal.com; prixreduits.net; clients.cdiscount.com; horaires.lefigaro.fr; fr.clasf.com \\
\hline
{ \noindent \bf \color{blue} paris-sorbonne.academia.edu} & cnrs.academia.edu; ehess.academia.edu; uclouvain.academia.edu; ephe.academia.edu; univ-paris1.academia.edu; univ-paris8.academia.edu; univ-lorraine.academia.edu; mindtools.com; univ-catholyon.academia.edu; ancient.eu; ffmedievale.forumgratuit.org; oxford.academia.edu; unil.academia.edu; iprofesional.com; theartstory.org; univ-amu.academia.edu; marineinsight.com; mapsofindia.com; trend-online.com; coniugazione.it; diggita.it; docsity.com; biografiasyvidas.com; infoplease.com; fr.actualitix.com; docplayer.es; thelocal.de; lectures49.over-blog.com; diariodocentrodomundo.com.br; cinemagia.ro \\
\hline
{ \noindent \bf \color{blue} renault-laguna.com} & megane3.fr; gps-carminat.com; megane2.superforum.fr; r25-safrane.net; diagnostic-auto.com; renault-zoe.forumpro.fr; techniconnexion.com; gamblewiz.com; forum-super5.fr; lesamisdudiag.com; forum.autocadre.com; minivanchrysler.com; renault-clio-4.forumpro.fr; bmw-one.com; club.caradisiac.com; lesamisdelaprog.com; forum-bmw.fr; alfaromeo-online.com; marcopolo.superforum.fr; v2-honda.com; alfa147-france.net; question-auto.fr; forum308.com; qashqai-passion.info; automobile-conseil.fr; fr.motocrossmag.be; magmotardes.com; auto-evasion.com; btf-automoto-asq.fr; fr.bmwfans.info \\
\hline
{ \noindent \bf \color{blue} excel-plus.fr} & tech-connect.info; jetaide.com; officepourtous.com; lame.buanzo.org; it.ccm.net; lecompagnon.info; patatos.over-blog.com; faclic.com; abracadabrapdf.net; faqword.com; windows.developpez.com; thehackernews.com; jiho.com; cartoucherecharge.fr; questionbureautique.over-blog.com; comment-supprimer.com; cgsecurity.org; silkyroad.developpez.com; technologie.toutcomment.com; java.developpez.com; br.ccm.net; phptester.net; lephpfacile.com; blogosquare.com; monpc-pro.fr; python.developpez.com; astuces.jeanviet.info; poftut.com; tecadmin.net; blog-nouvelles-technologies.fr \\
\hline
{ \noindent \bf \color{blue} jeuxvideo.org} & pvpro.com; fallout.fandom.com; pokecommunity.com; xbox-mag.net; make-fortnite-skins.com; garrycity.fr; en.riotpixels.com; brainly.com; gtalogo.com; pngimg.com; 3daimtrainer.com; creativeuncut.com; twitchoverlay.com; en.magicgameworld.com; frondtech.com; discord.me; 11anim.com; kiranico.com; dllme.com; jeugeek.com; myinstants.com; online-voice-recorder.com; mugenarchive.com; honga.net; strawpoll.com; chompy.app; gamepressure.com; cleverbot.com; rp-manga.forum-canada.com; filedropper.com \\
\hline
{ \noindent \bf \color{blue} farmville2free.com} &  fv2freegifts.org; goldenlifegroup.com; fb1.farm2.zynga.com; juegossocial.com; fv-zprod-tc-0.farmville.com; megazebra-facebook-trails.mega-zebra.com; zy2.farm2.zynga.com; gameskip.com; fv-zprod.farmville.com; actiplay-asn.com; iscool.iscoolapp.com; farmvilledirt.com; secure1.mesmo.tv; megazebra-facebook.mega-zebra.com; prod-web-pool.miniclip.com; banner2.cookappsgames.com; puzzledhearts.com; zynga.com; buggle.cookappsgames.com; apps.facebook.com; deliresetamities.1fr1.net; bubblecoco.cookappsgames.com; apps.fb.miniclip.com; rummikub-apps.com; amomama.fr; gifwi.com; jigsawpuzzlequest.com:3000; pengle.cookappsgames.com; webgl.exoty.com; fr.opossumsauce.com \\
\hline
{ \noindent \bf \color{blue} vogue.fr} & vanityfair.fr; vogue.com; fr.metrotime.be; vivreparis.fr; o.nouvelobs.com; apartmenttherapy.com; vice.com; lefaso.net; timeout.fr; unjourdeplusaparis.com; whosdatedwho.com; pariszigzag.fr; wwd.com; gq.com; taddlr.com; vanityfair.com; elle.com; brain-magazine.fr; admagazine.fr; fashiongonerogue.com; people.com; glamourparis.com; leplus.nouvelobs.com; unilad.co.uk; hellomagazine.com; maliactu.net; noisey.vice.com; france-hotel-guide.com; thisisinsider.com; lanouvelletribune.info \\
\hline
{ \noindent \bf \color{blue} tripadvisor.fr} & monnuage.fr; cityzeum.com; fr.hotels.com; voyageforum.com; rome2rio.com; virail.fr; carnetdescapades.com; petitfute.com; kayak.fr; voyages.michelin.fr; lonelyplanet.fr; expedia.fr; partir.com; quandpartir.com; salutbyebye.com; mackoo.com; actualitix.com; voyages.ideoz.fr; voyage.linternaute.com; week-end-voyage-lisbonne.com; skyscanner.fr; toocamp.com; plages.tv; routard.com; momondo.fr; gotoportugal.eu; evous.fr; esky.fr; l-itineraire.paris; lepetitmoutard.fr \\
\hline
\end{tabular}

%% file: nearest_neighbors.tex
%auto-ignore
\begin{tabular}{C{4cm}|p{30cm}}
\hline
\hline
{\bf Domain} & \begin{minipage}{1.6\columnwidth} \center \bf $30$-nearest neighbors (URLS)\end{minipage} \\
\hline
{ \noindent \bf \color{blue} huffingtonpost.es} & cope.es; m.eldiario.es; okdiario.com; verne.elpais.com; blogs.elconfidencial.com; vozpopuli.com; elespanol.com; smoda.elpais.com; libertaddigital.com; cadenaser.com; sevilla.abc.es; elmon.cat; elperiodico.com; levante-emv.com; kiosko.net; elcorreo.com; motor.elpais.com; cronicaglobal.elespanol.com; elplural.com; ara.cat; rac1.cat; eldiario.es; heraldo.es; elperiodico.cat; eleconomista.es; diariodesevilla.es; guiadelocio.com; lasexta.com; periodistadigital.com; mismarcadores.com \\
\hline
{ \noindent \bf \color{blue} lesechos.fr} & latribune.fr; afrique.latribune.fr; business.lesechos.fr; bfmbusiness.bfmtv.com; financedemarche.fr; challenges.fr; investopedia.com; actufinance.fr; lopinion.fr; contrepoints.org; rfi.fr; etudiant.lefigaro.fr; hbrfrance.fr; capital.fr; e-marketing.fr; marianne.net; journaldunet.com; 05-habillages-theplacetobid.fr; courrierinternational.com; nouvelobs.com; mtf-b2b-asq.fr; lexpansion.lexpress.fr; zonebourse.com; start.lesechos.fr; lentreprise.lexpress.fr; boursier.com; manager-go.com; investing.com; boursedirect.fr; journaldeleconomie.fr \\
\hline
{ \noindent \bf \color{blue} orange.fr} & actu.orange.fr; lemoteur.orange.fr; messagerie.orange.fr; login.orange.fr; finance.orange.fr; sports.orange.fr; meteo.orange.fr; tendances.orange.fr; programme-tv.orange.fr; news.orange.fr; boutique.orange.fr; pro.orange.fr; chaines-tv.orange.fr; ww.orange.fr; agenda.orange.fr; people.orange.fr; zapping-tv.orange.fr; mescontacts.orange.fr; mail01.orange.fr; occasion.auto.orange.fr; video-streaming.orange.fr; musique.orange.fr; tv.orange.fr; mail02.orange.fr; auto.orange.fr; webmail.orange.fr; 118712.fr; cineday.orange.fr; belote-coinchee.net; mahjonggratuit.fr \\
\hline
{ \noindent \bf \color{blue} leparisien.fr} & cnews.fr; atlasinfo.fr; lefigaro.fr; lejdd.fr; jforum.fr; marianne.net; video.lefigaro.fr; tendanceouest.com; bladi.net; observalgerie.com; causeur.fr; scoopnest.com; etudiant.lefigaro.fr; actu17.fr; lalibre.be; fdesouche.com; people.bfmtv.com; rmc.bfmtv.com; bfmtv.com; ici.radio-canada.ca; nouvelobs.com; amp.lefigaro.fr; breizh-info.com; fr.euronews.com; observers.france24.com; tunisienumerique.com; courrierinternational.com; lopinion.fr; sfrpresse.sfr.fr; 94.citoyens.com \\
\hline
{ \noindent \bf \color{blue} reddit.com} & imgur.com; old.reddit.com; askreddit.reddit.com; pcgamer.com; anime.reddit.com; france.reddit.com; gamefaqs.gamespot.com; totalwar.reddit.com; nintendoswitch.reddit.com; gaming.reddit.com; knowyourmeme.com; redditad.com; pcgaming.reddit.com; funny.reddit.com; europe.reddit.com; leagueoflegends.reddit.com; all.reddit.com; freefolk.reddit.com; pcmasterrace.reddit.com; soccer.reddit.com; globaloffensive.reddit.com; gamesradar.com; dankmemes.reddit.com; gfycat.com; gameofthrones.reddit.com; overwatch.reddit.com; popular.reddit.com; smashbros.reddit.com; competitiveoverwatch.reddit.com; aww.reddit.com \\
\hline
{ \noindent \bf \color{blue} expedia.fr} & momondo.fr; skyscanner.fr; kayak.fr; fr.lastminute.com; fr.hotels.com; flights-results.liligo.fr; ebookers.fr; esky.fr; opodo.com; secure.lastminute.com; bravofly.fr; opodo.fr; vol.lastminute.com; vols.idealo.fr; locations.lastminute.com; quandpartir.com; opodo.ch; reservation.lastminute.com; quellecompagnie.com; trainhotel.lastminute.com; lonelyplanet.fr; easyvols.fr; voyages.bravofly.fr; edreams.fr; sejour.lastminute.com; rome2rio.com; voyagespirates.fr; jetcost.com; voyage.lastminute.com; virail.fr \\
\hline
{ \noindent \bf \color{blue} tractorfan.fr} &  discountfarmer.com; forum.farm-connexion.com; angleterre.meteosun.com; songs-tube.net; materieltp.fr; assovttroc.clicforum.fr; opel-mokka.forumpro.fr; spa-du-dauphine.fr; vanvesactualite.blog4ever.com; calcul-frais-de-notaire.fr; sectr.net; cuir-creation.forum-box.com; fc-fief-geste.footeo.com; classements.snt-voile.org; rjm-radio.fr; gps-tomtom.fr; v-strom.superforum.fr; migrateurs.forumgratuit.org; gazoline.forumactif.com; recuperation-metaux.annuairefrancais.fr; voitures-societe.ooreka.fr; fcplouay.footeo.com; equishopping.com; annonce123.com; 36kines.com; bastia.onvasortir.com; renault5.forumactif.com; globaldjmix.com; enviedechasser.fr; squidtv.net \\
\hline
{ \noindent \bf \color{blue} welt.de} & zeit.de; sueddeutsche.de; faz.net; tagesspiegel.de; sport1.de; kicker.de; saarbruecker-zeitung.de; tz.de; bild.de; sportbild.bild.de; hartgeld.com; nzz.ch; abendblatt.de; express.de; n-tv.de; bunte.de; handelsblatt.com; merkur.de; bz-berlin.de; aargauerzeitung.ch; badische-zeitung.de; promiflash.de; focus.de; t-online.de; transfermarkt.de; finanzen.net; sport.de; flashscore.de; rp-online.de; stylebook.de \\
\hline
{ \noindent \bf \color{blue} foreca.fr} & my-meteo.com; fr.meteovista.be; fr.tutiempo.net; meteopassion.com; de.sat24.com; nosvolieres.com; meteo-sud-aveyron.over-blog.com; xn--mto-bmab.fr; palombe.com; calculerdistance.fr; meteo81.fr; meteosurfcanarias.com; meteo-normandie.fr; meteobelgique.be; planete-ardechoise.com; parisbrestparis2007.actifforum.com; indicatifs.htpweb.fr; etatdespistes.com; easycounter.com; hauteurdeneige.com; testadsl.net; m.meteorama.fr; discountfarmer.com; prevision-meteo.ch; refugeanimalierdupaysdelanderneau.over-blog.com; infosski.com; grottes-france.com; meteolanguedoc.com; impactfm.fr; pont-ile-de-re.com \\
\hline
{ \noindent \bf \color{blue} auto-moto.com} & caradisiac.com; largus.fr; news.autojournal.fr; test-auto.auto-moto.com; auto-mag.info; feline.cc; motorlegend.com; essais.autojournal.fr; automobile-magazine.fr; turbo.fr; autoplus.fr; promoneuve.fr; automobile-sportive.com; neowebcar.com; latribuneauto.com; moniteurautomobile.be; palais-de-la-voiture.com; fr.automobiledimension.com; motoservices.com; autotitre.com; fiches-auto.fr; autojournal.fr; blogzineauto.com; voiture.autojournal.fr; essais-autos.com; notice-utilisation-voiture.fr; sportauto.fr; abcmoteur.fr; recherche.autoplus.fr; news.sportauto.fr \\
\hline
{ \noindent \bf \color{blue} az-online.de} & abountifulkitchen.com; thesurvivalgardener.com; leinetal24.de; brittanyherself.com; symbols.com; ourpaleolife.com; msl24.de; milliondollarjourney.com; arthritis-health.com; thehollywoodunlocked.com; preschoolmom.com; lovechicliving.co.uk; paleoglutenfree.com; lettermenrow.com; theeasyhomestead.com; raegunramblings.com; evolving-science.com; mu-43.com; juneauempire.com; mikseri.net; e1.ru; thedailytay.com; alittlecraftinyourday.com; comfortablydomestic.com; chicksonright.com; brepurposed.porch.com; kiwilimon.com; grandforksherald.com; catholicstand.com; greatandhra.com \\
\hline
{ \noindent \bf \color{blue} tempsdecuisson.net} & cuisine-facile.com; temps-de-cuisson.info; yummix.fr; aux-fourneaux.fr; cuisinenligne.com; audreycuisine.fr; mamina.fr; uneplumedanslacuisine.com; cnz.to; ricardocuisine.com; chefnini.com; cuisinealafrancaise.com; toutlemondeatabl.canalblog.com; marciatack.fr; atelierdeschefs.fr; lesepicesrient.fr; yumelise.fr; lacuisinededoria.over-blog.com; cuisinebassetemperature.com; recettessimples.fr; perleensucre.com; la-cuisine-des-jours.over-blog.com; lesjoyauxdesherazade.com; fraichementpresse.ca; gustave.com; gateaux-chocolat.fr; toques2cuisine.com; recettesduchef.fr; petitsplatsentreamis.com; amandinecooking.com \\
\hline
{ \noindent \bf \color{blue} cnn.com} & us.cnn.com; stadiumtalk.com; thedailybeast.com; itpro.co.uk; uk.reuters.com; euronews.com; theargus.co.uk; theatlantic.com; thedailymash.co.uk; trendscatchers.co.uk; grimsbytelegraph.co.uk; lancashiretelegraph.co.uk; digg.com; spectator.co.uk; politico.eu; blogs.spectator.co.uk; newstatesman.com; huffingtonpost.co.uk; expressandstar.com; puzzles.bestforpuzzles.com; chroniclelive.co.uk; derbytelegraph.co.uk; irishexaminer.com; globalnews.ca; sportinglife.com; slashdot.org; rte.ie; farandwide.com; kentonline.co.uk; thenational.scot \\
\hline
{ \noindent \bf \color{blue} portail-cloture.ooreka.fr} & bricolage-facile.net; mur.ooreka.fr; pierreetsol.com; bricolage.jg-laurent.com; abri-de-jardin.ooreka.fr; aac-mo.com; fr.rec.bricolage.narkive.com; decoration.ooreka.fr; piscineinfoservice.com; bricoleurpro.com; amenagementdujardin.net; betonniere.ooreka.fr; assainissement.ooreka.fr; fenetre.ooreka.fr; expert-peinture.fr; terrasse.ooreka.fr; tondeuse.ooreka.fr; debroussailleuse.ooreka.fr; peinture.ooreka.fr; carrelage.ooreka.fr; parquet.ooreka.fr; amenagement-de-jardin.ooreka.fr; toiture.ooreka.fr; poimobile.fr; schema-electrique.net; installation-electrique.ooreka.fr; forum-maconnerie.com; muramur.ca; wc.ooreka.fr; plaque-de-cuisson.ooreka.fr \\
\hline
{ \noindent \bf \color{blue} sport.fr} & infomercato.fr; parisfans.fr; topmercato.com; vipsg.fr; footradio.com; mercatofootanglais.com; le10sport.com; buzzsport.fr; footparisien.com; foot-sur7.fr; planetepsg.com; pariskop.fr; paristeam.fr; footlegende.fr; le-onze-parisien.fr; mercatoparis.fr; 90min.com; canal-supporters.com; allpaname.fr; sportune.fr; livefoot.fr; culturepsg.com; jeunesfooteux.com; losclive.com; mercatolive.fr; footballclubdemarseille.fr; parisunited.fr; football-addict.com; olympique-et-lyonnais.com; football.fr \\
\hline
{ \noindent \bf \color{blue} anti-crise.fr} & cfid.fr; forum.anti-crise.fr; gesti-odr.com; echantillonsclub.com; plusdebonsplans.com; cataloguemate.fr; promoalert.com; argentdubeurre.com; madstef.com; forum.madstef.com; vos-promos.fr; mesechantillonsgratuits.fr; franceechantillonsgratuits.com; tiendeo.fr; tous-testeurs.com; maximum-echantillons.com; ofertolino.fr; auchan.fr; pubeco.fr; echantinet.com; echantillonsgratuits.fr; promo-conso.net; bons-plans-astuces.com; commerces.com; lp.testonsensemble.com; grattweb.fr; promobutler.be; jeu-concours.biz; mafamillezen.com; hitwest.com \\
\hline
{ \noindent \bf \color{blue} auchan.fr} & but.fr; conforama.fr; vente-unique.com; rueducommerce.fr; fr.shopping.com; cdiscount.com; touslesprix.com; promobutler.be; webmarchand.com; mistergooddeal.com; offrespascher.com; argentdubeurre.com; cataloguemate.fr; leguide.com; promoalert.com; vos-promos.fr; fr.xmassaver.net; cdiscountpro.com; clients.cdiscount.com; promopascher.com; meilleurvendeur.com; clubpromos.fr; tiendeo.fr; plusdebonsplans.com; promo-conso.net; iziva.com; destockplus.com; pubeco.fr; meonho.info; fr.clasf.com \\
\hline
{ \noindent \bf \color{blue} paris-sorbonne.academia.edu} & flux-info.fr; elmostrador.cl; makaan.com; univ-montp3.academia.edu; e-lawresources.co.uk; babycenter.com; newocr.com; insight.co.kr; grandes-inventions.com; police-scientifique.com; entrainementfootballpro.fr; muyinteresante.es; ancient.eu; iprofesional.com; slovnik.aktuality.sk; midis101.com; quemas.mamaslatinas.com; adventureinyou.com; wardrawings.be; esky.ro; puzzle-futoshiki.com; univ-paris8.academia.edu; letssingit.com; learn101.org; mindtools.com; medicoresponde.com.br; br.rfi.fr; lazycatkitchen.com; realmenrealstyle.com; eve-adam.over-blog.com \\
\hline
{ \noindent \bf \color{blue} renault-laguna.com} & megane3.fr; gps-carminat.com; megane2.superforum.fr; lesamisdudiag.com; car-actu.com; diagnostic-auto.com; r25-safrane.net; lesamisdelaprog.com; forum.autocadre.com; renault-clio-4.forumpro.fr; renault-zoe.forumpro.fr; v2-honda.com; ccfrauto.com; minivanchrysler.com; techniconnexion.com; forum308.com; lemecano.fr; obd-data.com; forum-super5.fr; club.caradisiac.com; toutsurlamoto.com; cliomanuel.org; forum-kia-sportage.com; tlemcen-electronic.com; focusrstteam.com; 306inside.com; cmonofr.net; 207.fr; automobile-conseil.fr; gamblewiz.com \\
\hline
{ \noindent \bf \color{blue} excel-plus.fr} & tech-connect.info; thehackernews.com; lecompagnon.info; panoptinet.com; slice42.com; aliasdmc.fr; astuces.jeanviet.info; nalaweb.com; patatos.over-blog.com; jiho.com; abavala.com; ohmymac.fr; br.ccm.net; easy-pc.org; wisibility.com; filedesc.com; semageek.com; windows.developpez.com; jetaide.com; galaxynote.fr; fr.stealthsettings.com; chezcyril.over-blog.com; tuto4you.fr; faclic.com; alvinalexander.com; thegeekstuff.com; hirensbootcd.org; faqword.com; openshot.org; zoomonapps.com \\
\hline
{ \noindent \bf \color{blue} jeuxvideo.org} & alsumaria.tv; minecraft-zh.gamepedia.com; infovisual.info; everyonepiano.com; footstream.live; memedroid.com; darkandlight.gamepedia.com; mbti.forumactif.fr; gachagames.net; honga.net; en.magicgameworld.com; garrycity.fr; satelis-passion.forumactif.com; blog-insideout.com; fallout.fandom.com; howtomechatronics.com; cshort.org; fr.trackitonline.ru; openthefile.net; lutain.over-blog.com; alphabetagamer.com; solveyourtech.com; online-voice-recorder.com; png2jpg.com; ffxforever.over-blog.com; arkhamhorrorfr.forumactif.com; en.riotpixels.com; pexiweb.be; petri.com; rasage-traditionnel.com \\
\hline
{ \noindent \bf \color{blue} farmville2free.com} &  goldenlifegroup.com; fv2freegifts.org; juegossocial.com; fv-zprod-tc-0.farmville.com; fb1.farm2.zynga.com; zy2.farm2.zynga.com; gameskip.com; fv-zprod.farmville.com; megazebra-facebook-trails.mega-zebra.com; farmvilledirt.com; megazebra-facebook.mega-zebra.com; iscool.iscoolapp.com; secure1.mesmo.tv; belote-prod-multi.iscoolapp.com; jigsawpuzzlequest.com:3000; fr.puzzle-loop.com; connect.arkadiumhosted.com; pengle.cookappsgames.com; buggle.cookappsgames.com; banner2.cookappsgames.com; apps.fb.miniclip.com; goobox.fr; prod-web-pool.miniclip.com; actiplay-asn.com; apps.facebook.com; snf-web.popreach.com; bubblecoco.cookappsgames.com; rummikub-apps.com; watersplash.cookappsgames.com; zynga.com \\
\hline
{ \noindent \bf \color{blue} vogue.fr} & vanityfair.fr; vogue.com; vivreparis.fr; fr.metrotime.be; brain-magazine.fr; o.nouvelobs.com; parismatch.be; pariszigzag.fr; admagazine.fr; unilad.co.uk; konbini.com; monblogdefille.com; affairesdegars.com; neonmag.fr; vice.com; nordpresse.be; leplus.nouvelobs.com; lataille.fr; people-bokay.com; commeuncamion.com; fr.euronews.com; demotivateur.fr; star24.tv; madmoizelle.com; vl-media.fr; whosdatedwho.com; sciencepost.fr; physiquedereve.fr; ztele.com; twog.fr \\
\hline
{ \noindent \bf \color{blue} tripadvisor.fr} & fr.hotels.com; cityzeum.com; voyages.michelin.fr; lonelyplanet.fr; monnuage.fr; voyageforum.com; rome2rio.com; toocamp.com; virail.fr; partir.com; carnetdescapades.com; quellecompagnie.com; mackoo.com; expedia.fr; momondo.fr; salutbyebye.com; orangesmile.com; skyscanner.fr; voyages.ideoz.fr; cars.liligo.fr; quandpartir.com; kelbillet.com; gotoportugal.eu; officiel-des-vacances.com; busradar.fr; week-end-voyage-lisbonne.com; les-escapades.fr; voyage.linternaute.com; bouger-voyager.com; voyagespirates.fr \\
\hline
\end{tabular}